\PassOptionsToPackage{table}{xcolor}
\documentclass[]{afu}
\usepackage{amsmath,amssymb,amsthm}
\usepackage{fontawesome5}
\definecolor{primary}{HTML}{1B2A4A}
\usepackage{fix-cm}

\usepackage{xcolor}
\usepackage{multirow}
\usepackage{colortbl}
\usepackage{tabularx}
\usepackage{tcolorbox}
\usepackage{enumitem}
\usepackage{pifont}
\usepackage{needspace}

\usepackage{url}
\usepackage{makecell}
\usepackage{booktabs}
\usepackage{placeins}
\usepackage{longtable}
\usepackage{xltabular}
\usepackage{ragged2e}
\usepackage{siunitx}
\usepackage{xspace}
\usepackage{etoolbox}
\usepackage{graphbox}
\usepackage{tikz}
\usepackage{pgfplots}
\definecolor{ecgtableblue}{HTML}{E4F1FB}
\definecolor{ecgtablegray}{HTML}{ECECEC}
\definecolor{ecgtrendblue}{HTML}{287DB1}
\definecolor{ecgtrendred}{HTML}{C94C46}
\newcommand{\ecgmedical}{\textcolor[HTML]{D52B2B}{\raisebox{0.05ex}{\tikz[baseline=-0.55ex,x=1ex,y=1ex]{\fill (0.35,0) rectangle (0.75,1.1);\fill (0,0.35) rectangle (1.1,0.75);}}}}
\newcommand{\ecgup}{\textcolor{ecgtrendblue}{{\fontsize{7.5}{9}\selectfont\ensuremath{\uparrow}}}}
\newcommand{\ecgdown}{\textcolor{ecgtrendred}{{\fontsize{7.5}{9}\selectfont\ensuremath{\downarrow}}}}
\newcommand{\ecgerrdown}{\textcolor{ecgtrendblue}{{\fontsize{7.5}{9}\selectfont\ensuremath{\downarrow}}}}
\newcommand{\ecgpass}{\textcolor{ecgtrendblue}{\ding{51}}}
\newcommand{\ecgdelta}[1]{{\fontsize{8}{9.5}\selectfont #1}}
\newcommand{\ecgtablefont}{\fontsize{9.5}{11.5}\selectfont}

\definecolor{caseevidence}{HTML}{235E91}
\definecolor{casecautioncolor}{HTML}{8A551A}
\newcommand{\casekey}[1]{\textcolor{caseevidence}{\textbf{#1}}}
\newcommand{\casecaution}[1]{\textcolor{casecautioncolor}{\textbf{#1}}}
\newcommand{\casehl}[1]{\tcbox[on line,boxrule=0pt,colback=ecgtableblue,arc=1pt,left=1.2pt,right=1.2pt,top=0.1pt,bottom=0.1pt,boxsep=0pt]{\textcolor{caseevidence}{\textbf{#1}}}}

\usepackage{algorithm}
\usepackage{algpseudocode}

\definecolor{mygray}{gray}{0.9}
\definecolor{myblue}{HTML}{F0FFFF}
\definecolor{myblue_light}{HTML}{EAF6FF}
\definecolor{myblue_dark}{HTML}{2B57A0}
\definecolor{myblue_lightv1}{HTML}{F2F7E0}
\definecolor{mygreen}{RGB}{60, 179, 113}
\definecolor{myyellow_light}{RGB}{255, 255, 224}
\definecolor{lightgreen}{HTML}{D3E5C7}
\definecolor{deepgreen}{RGB}{46,139,87}
\definecolor{mydarkgreen}{rgb}{0,0,0.6}

\definecolor{brandnavy}{HTML}{0B2857}
\definecolor{brandblue}{HTML}{176DCE}
\definecolor{brandcyan}{HTML}{60A4D0}
\definecolor{brandgreen}{HTML}{18B45D}
\definecolor{brandmint}{HTML}{59CE8A}
\definecolor{brandcoral}{HTML}{F65669}
\definecolor{brandgold}{HTML}{FCAB12}
\definecolor{brandslate}{HTML}{576D90}

\usetikzlibrary{positioning, fit, backgrounds, arrows.meta, shapes.geometric}
\pgfplotsset{compat=1.18}

\newtcolorbox[auto counter, number within=section]{prompt}[1][]{
    colbacktitle=black!60,
    coltitle=white,
    fontupper=\normalsize,
    boxsep=5pt,
    left=2pt,
    right=2pt,
    top=2pt,
    bottom=2pt,
    boxrule=1pt,
    #1,
}

\newcommand{\targetmodel}{Qwen3.5-35B-A3B}

\newcommand{\AppendixNeedspace}[1]{\par\begingroup\dimen0=\pagegoal\advance\dimen0 by -\pagetotal\ifdim\dimen0<#1\newpage\fi\endgroup}
\title{ECG-RSI: Recursive Self-Improvement through Targeted Electrocardiogram Waveform Evolution}
\renewcommand{\titlelist}{%
  \makebox[\linewidth][c]{{\Large\bfseries
  \raisebox{-0.5\height}{\includegraphics[height=2.1em]{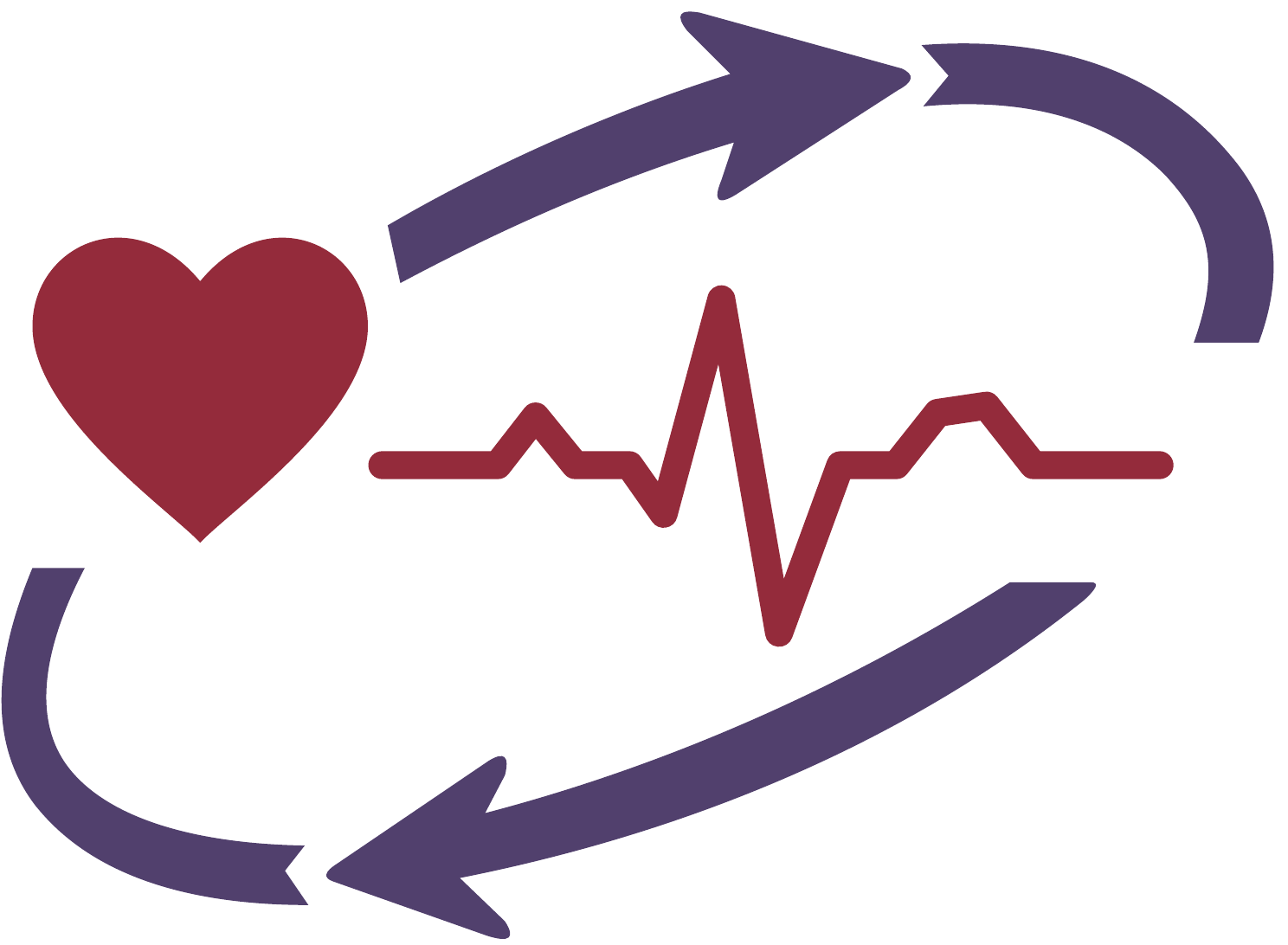}}\hspace{0.7em}%
  \raisebox{-0.5\height}{\shortstack[c]{ECG-RSI: Recursive Self-Improvement through\\[3pt]
Targeted Electrocardiogram Waveform Evolution}}}}%
}
\renewcommand{\authorlist}{%
  {\small\normalfont\centering
  \mbox{\mbox{\href{https://openreview.net/profile?id=\string~Wang_Xiaoliang2}{\textcolor{black}{Xiaoliang Wang}}\textsuperscript{1,2,*}},\hspace{0.25em}\mbox{\href{https://openreview.net/profile?id=\string~Zhuo_Chen3}{\textcolor{black}{Zhuo Chen}}\textsuperscript{2,*}},\hspace{0.25em}\mbox{\href{https://openreview.net/profile?id=\string~Chuanyang_Zheng2}{\textcolor{black}{Chuanyang Zheng}}\textsuperscript{2}},\hspace{0.25em}\mbox{\href{https://openreview.net/profile?id=\string~Jiangwei_Lao1}{\textcolor{black}{Jiangwei Lao}}\textsuperscript{2}},\hspace{0.25em}\mbox{\href{https://openreview.net/profile?id=\string~Quanbin_Wang1}{\textcolor{black}{Quanbin Wang}}\textsuperscript{2}}}\\[2pt]
  \mbox{\mbox{\href{https://openreview.net/profile?id=\string~Honglin_Lin2}{\textcolor{black}{Honglin Lin}}\textsuperscript{3,4}},\hspace{0.25em}\mbox{\href{https://openreview.net/profile?id=\string~Jiajia_Liu1}{\textcolor{black}{Jiajia Liu}}\textsuperscript{2}},\hspace{0.25em}\mbox{\href{https://openreview.net/profile?id=\string~Zhihong_Lu2}{\textcolor{black}{Zhihong Lu}}\textsuperscript{2}},\hspace{0.25em}\mbox{\href{https://openreview.net/profile?id=\string~Hao_Jin4}{\textcolor{black}{Hao Jin}}\textsuperscript{1}},\hspace{0.25em}\mbox{\href{https://openreview.net/profile?id=\string~Jian_Wang25}{\textcolor{black}{Jian Wang}}\textsuperscript{2}},\hspace{0.25em}\mbox{Jinjie Gu\textsuperscript{2}},\hspace{0.25em}\mbox{\href{https://openreview.net/profile?id=\string~Lijun_Wu1}{\textcolor{black}{Lijun Wu}}\textsuperscript{2,3,\ensuremath{\dagger}}}}\par
  }%
}
\renewcommand{\affiliationlist}{%
  {\small\normalfont\centering
  \textsuperscript{1}Zhejiang University\quad
  \textsuperscript{2}Ant Group, AFu Team\\[1pt]
  \textsuperscript{3}Shanghai Artificial Intelligence Laboratory\quad
  \textsuperscript{4}Shanghai Jiao Tong University\par}%
}
\renewcommand{\contributionlist}{%
  {\footnotesize\normalfont\centering\vspace{1mm}
  \textsuperscript{*}Equal contribution\quad
  \textsuperscript{\ensuremath{\dagger}}Corresponding author\par}%
}
\renewcommand{\metadatalist}{%
  {\small\begin{tabularx}{\linewidth}{@{}l@{\enspace}X@{}}
  \textbf{Correspondence:} & Lijun Wu, \email{wlj554215@antgroup.com}\\
  \textbf{Project:} & \url{https://pork-pay.github.io/ECG-RSI/}\\
  \textbf{Code:} & \url{https://github.com/pork-pay/ECG-RSI}
  \end{tabularx}}%
}
\hypersetup{pdftitle={ECG-RSI: Recursive Self-Improvement through Targeted Electrocardiogram Waveform Evolution},
  pdfauthor={Xiaoliang Wang, Zhuo Chen, Chuanyang Zheng, Jiangwei Lao, Quanbin Wang, Honglin Lin, Jiajia Liu, Zhihong Lu, Hao Jin, Jian Wang, Jinjie Gu, Lijun Wu}}

\abstract{
Electrocardiogram (ECG) models need supervision that targets evolving weaknesses and remains faithful to waveform evidence. We introduce \textbf{ECG-RSI}, a recursive self-improvement framework combining verified waveform synthesis with feedback-guided intervention search. Signal remeasurement verifies waveform edits; construction errors guide generator repair, while learner errors motivate interventions compared on a separate selection set. Across seven benchmarks, three rounds raise Qwen3.5-35B-A3B\textquotesingle s mean question-answering accuracy from \textbf{34.57\% to 56.04\%} and mean classification F1 from \textbf{24.27 to 34.03}. Across the retained trajectory, \textbf{the preferred supervision shifts from direct reinforcement to confusion-focused contrasts}. Continued feedback correction yields diminishing external gains. These results motivate \textbf{separating error diagnosis from intervention selection} and checking whether gains on the feedback set transfer to independent benchmarks.

}

\let\paperabstracttext\abstractlist
\renewcommand{\abstractlist}{%
  {\setlength{\parskip}{0pt}\vspace{3mm}
   {\centering\bfseries Abstract\par}
   \vspace{2mm}\paperabstracttext\par}%
}

\begin{document}
\maketitle

\suppressfloats[t]

\begin{figure}[t]
\centering
\includegraphics[width=\linewidth,height=0.72\textheight,keepaspectratio]{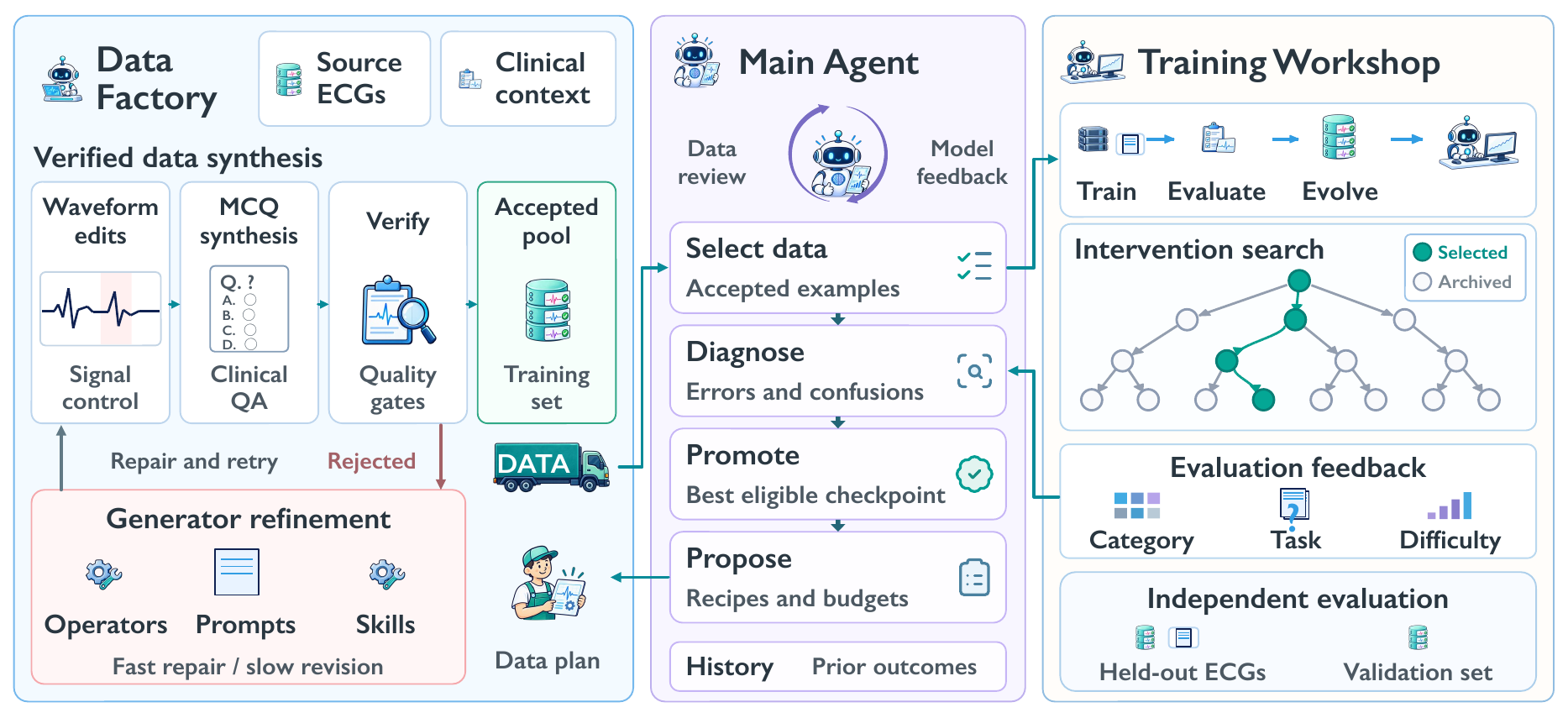}
\caption{\textbf{ECG-RSI}: feedback-driven data and model evolution. The data factory constructs and verifies ECG questions, with rejected candidates driving local revisions of operators, prompts, and skills. The main agent diagnoses feedback, proposes interventions, and selects accepted data for training. The training workshop evaluates candidate checkpoints and supplies category-, task-, and difficulty-level feedback. Independent evaluation is separated from the update loop. The search tree and highlighted path illustrate candidate comparison and model selection.}
\label{fig:method}
\end{figure}

\section{Introduction}
\label{sec:introduction}
\label{sec:intro:shortcut}\label{sec:intro:approach}\label{sec:intro:contributions}

Electrocardiogram (ECG) interpretation requires connecting \textbf{waveform patterns} to \textbf{measurable features} and clinically meaningful conclusions. Multimodal models have begun to learn this connection from ECG images and instruction data~\citep{liu2024pulse,lan2025gem,DBLP:journals/corr/abs-2602-04279}. Effective ECG supervision grounds each answer in the waveform presented to the model. Changes in heart rate, intervals, or morphology can alter the interpretation of related ECG findings.

Existing studies improve ECG supervision through instruction construction and evidence-aware learning~\citep{liu2024pulse,lan2025gem,DBLP:journals/corr/abs-2602-04279}, while conditional signal generators expand waveform diversity~\citep{chen2022megan,lai2025diffusets}. Yet it remains difficult to turn a learner’s specific weakness into a controlled training example. A generated waveform must express the intended ECG feature without introducing conflicting changes, and its question and answer must agree with the resulting signal. Training mixtures must adapt to newly identified weaknesses, persistent confusions, and shifts in task difficulty after each round. Our central insight is to use \textbf{learner feedback} to specify ECG evidence and \textbf{signal measurements} to verify it.

We propose \textbf{ECG-RSI}, a framework that connects targeted ECG data synthesis with feedback-driven recursive self-improvement (Fig.~\ref{fig:method}). Its data factory generates questions from existing signals and uses constrained reverse perturbations to synthesize waveforms with requested feature contrasts. The resulting waveforms are remeasured before their questions and answers are accepted. Construction failures guide revisions to perturbation operators and question-generation workflows, while the learner’s errors guide the categories, tasks, and difficulty levels of subsequent training data. At each round, candidate data recipes are used to train and compare model branches, and the selected branch informs the \mbox{next round} of synthesis and training. 

\Needspace*{4\baselineskip}
Our contributions are threefold:
\begin{itemize}[leftmargin=*,label=\textbullet]
\item We propose \textbf{ECG-RSI}, a framework that separates construction feedback for repairing data generation from learner feedback for proposing and comparing subsequent training interventions.
\item We develop \textbf{target-driven reverse waveform synthesis}, which translates desired ECG feature contrasts into constrained signal perturbations and verifies the resulting waveforms to provide \mbox{measurement-grounded supervision}.
\item We analyze \textbf{how intervention utility changes with the learner}: the retained trajectory shifts from direct reinforcement to confusion-focused contrasts, while continued feedback correction yields diminishing gains on external benchmarks.
\end{itemize}

\section{Related Work}
\label{sec:related_work}

\subsection{ECG Multimodal Learning and Evidence-Grounded Supervision}
\label{sec:related_ecgmllm}

{\emergencystretch=1em
Medical vision-language foundation models provide a broader basis for domain adaptation. MedGemma adapts Gemma 3 models for \textbf{medical image and text understanding} and supports further task-specific fine-tuning~\citep{sellergren2025medgemma}. Complementary work examines the integration of knowledge graphs with multimodal learning~\citep{DBLP:journals/corr/abs-2402-05391}.\par}

ECG-language models have progressed from \textbf{ECG instruction tuning} and multimodal alignment toward more flexible question answering and \textbf{clinically grounded interpretation}. PULSE, a multimodal large language model for ECG image comprehension, is instruction-tuned on ECGInstruct, an ECG image instruction dataset, and evaluated on ECGBench, a benchmark for ECG image interpretation~\citep{liu2024pulse}. GEM links signal and image representations to measurable ECG evidence~\citep{lan2025gem}. ECG-R1 incorporates clinical protocols and diagnostic-evidence supervision into ECG interpretation ~\citep{DBLP:journals/corr/abs-2602-04279}. Subsequent work broadens the types of questions, input formats, and supervision available to ECG models: anyECG-chat supports flexible ECG inputs and multiple tasks ~\citep{supplied2026ref34}, Q-Heart introduces knowledge-informed ECG question answering ~\citep{supplied2026ref35}, ECG-LLM uses clinically structured questions for cardiac reasoning ~\citep{supplied2026ref31}, and ELF studies a simpler encoder-free ECG-language architecture ~\citep{han2026elf}. MEETI aligns signals, images, extracted features, and interpretations within a multimodal ECG dataset ~\citep{supplied2026ref37}, while ECG-Expert-QA provides diagnostically diverse questions for evaluation ~\citep{supplied2026ref32}. ECG input representations also affect performance ~\citep{supplied2026ref38}. Multi-source contrastive learning offers another way to broaden supervision: MS-ECG-FM aligns ECGs with ECG, echocardiography, radiology, and discharge reports, whose diagnostic information is complementary~\citep{lewis2026msecgfmuniversalelectrocardiogramfoundation}.

ECG Mirage probes whether predictions depend on the \textbf{patient-specific ECG} ~\citep{liang2026ecgmirage}. ECG-Reasoning-Benchmark examines stepwise ECG interpretation ~\citep{oh2026ecgbenchmark}, and recent work evaluates ECG reasoning by separating waveform perception from clinical deduction ~\citep{xu2026ecgreasoneval}. Image-based model evaluations likewise expose the difficulty of reliable ECG interpretation from rendered tracings ~\citep{supplied2026ref42}. These studies motivate supervision that remains tied to the waveform presented to the learner. ECG-RSI addresses this need by constructing requested feature contrasts and deriving questions and answers from measurements of the \mbox{resulting signals}. 

\subsection{Conditional ECG Synthesis and Controlled Interventions}
\label{sec:related_ecg_synthesis}

ECG synthesis has advanced through \textbf{disease-conditioned} adversarial generation ~\citep{chen2022megan}, conditional diffusion ~\citep{alcaraz2023sssd,skorik2024sssdnle}, clinical-text conditioning ~\citep{chung2023texttoecg}, and report- and patient-conditioned generation ~\citep{lai2025diffusets}. MIDT-ECG and FlowECG further investigate \textbf{waveform fidelity}, inter-lead coherence, and generation efficiency ~\citep{huang2025midtecg,bondar2025flowecg}. More directly related to controllability, ECGTwin generates condition-guided personalized ECGs while seeking to preserve individual characteristics ~\citep{supplied2026ref33}, and TransDiffECG explores semantically controllable signal generation ~\citep{supplied2026ref40}. Generative counterfactual ECGs have also been used to study how waveform changes affect an AI-ECG model’s predictions ~\citep{supplied2026ref41}.

ECG-RSI constructs training examples for a \emph{\textbf{specified learner weakness}}: it starts with a training-side waveform and a requested feature contrast, applies a constrained intervention, and remeasures the result. It accepts an example only when the intended change is achieved and relevant non-target properties are preserved; its question and answer are then grounded in the \textbf{resulting waveform}. 

\subsection{Feedback-Driven Data Evolution and Recursive Self-Improvement}
\label{sec:related_rsi}\label{sec:related:student-distillation}\label{sec:related_data_rsi}\label{sec:related_positioning}

Earlier self-improvement methods revise responses or retain verbal feedback ~\citep{madaan2023self,DBLP:conf/nips/ShinnCGNY23}, generate instruction data ~\citep{wang2023selfinstruct,xu2024wizardlm}, or iteratively refresh reasoning supervision ~\citep{zelikman2022star}. More recent systems connect generated tasks to \textbf{learner improvement}: CoEvolve guides task synthesis using learner behavior ~\citep{yang2026coevolve}, NeoHorse-1 adapts post-training mixtures using capability evidence ~\citep{neohorse2026}, and Absolute Zero and VisPlay couple task generation with solver improvement ~\citep{zhao2025absolutezero,he2025visplay}. SynCo makes this dependence explicit by jointly training a synthesizer and reasoner, with task validity, answer reliability, and outcome-grounded teachability providing feedback to the synthesizer~\citep{yang2026syncodatasynthesiscotraining}. Multiple reasoner rollouts per task provide this feedback, so estimated teaching value depends on the current policy's observed outcomes. RSIBench-Data places data construction and selection at the center of RSI evaluation ~\citep{DBLP:journals/corr/abs-2607-25886}. Related studies examine whether a self-synthetic pipeline continues to provide learnable information ~\citep{liu2026learnableinfo}, while VISA revises multimodal instruction synthesis using verification and \textbf{target-model failures} ~\citep{supplied2026ref44}. Recursive on-policy distillation co-evolves the privileged teacher with the student~\citep{yin2026recursiveopd}, while RLTL;DR internalizes concise feedback from failed attempts~\citep{kirchhof2026rltldr}. Both make the supervision produced by an earlier attempt useful to subsequent learning.

A complementary line changes the \textbf{mechanism of improvement}: agent code and workflows~\citep{zelikman2024stop,zhang2025dgm}, skills~\citep{wang2026metaskillevolve}, data preparation~\citep{DBLP:journals/corr/abs-2608-29966,supplied2026ref45}, or environments and search strategies~\citep{zheng2026dreamrsi,liu2026evox}. These updates require assessment: revision selection~\citep{supplied2026ref43,supplied2026ref48}, evolving self-evaluation~\citep{liu2026recurse}, hidden evaluations~\citep{srikanth2026aide2}, and pairwise candidate ranking~\citep{fu2026sift} provide different ways to compare candidates. Other studies guide subsequent adaptation~\citep{DBLP:journals/corr/abs-2608-27287,supplied2026ref50} or alternate harness evolution with model learning~\citep{xue2026sciencebuddy}. Structured knowledge is another object of improvement~\citep{supplied2026ref49,DBLP:journals/corr/abs-2608-23397}; in medicine, MedZERO couples task generation with evidence-grounded reasoning and controlled knowledge accumulation~\citep{dang2026medzero}.

Paired evaluations also help identify diminishing returns and earlier artifacts worth retaining~\citep{yin2026stopping}. In a setting with frozen model weights, Agent Plasticity measures how efficiently experience produces future held-out gains relative to learning cost, showing that endpoint performance, acquisition efficiency, and transfer to harder conditions can diverge~\citep{singh2026agentplasticitymeasuringselfimprovement}. ECG-RSI grounds intervention proposals in ECG-specific error diagnostics: construction feedback repairs synthesis, learner feedback proposes interventions, and separate selection and external evaluations assess their utility.

\section{Method}
\label{sec:method}

ECG-RSI couples \textbf{verified synthesis} with \textbf{feedback-driven training} (Fig.~\ref{fig:method}). Construction feedback checks waveform and question validity; learner feedback selects data for model-specific utility~\citep{DBLP:journals/corr/abs-2607-25886}.

\subsection{Framework and Feedback Interfaces}
\label{sec:method:rsi}

\paragraph{System state and data boundaries.}

At round \(t\), the system state comprises the accepted learner \(\theta_t\), retained training pool \(D_t\), and \textbf{construction policy} \(\phi_t\). The policy contains retrieval constraints, waveform operators, prompts, and reusable rules, separate from learner weights~\citep{zelikman2024stop}. Source ECGs and morphology donors are restricted to training data and are \textbf{patient-disjoint} from development and independent evaluation data.

\(V_{\mathrm{fb}}\) diagnoses errors; \(V_{\mathrm{sel}}\) selects models. Evaluation data are held out from generation and selection.

\paragraph{Two feedback loops.}

\textbf{Construction feedback} identifies invalid waveform edits or questions and guides generator repair. \textbf{Learner feedback} guides recipes using gaps in ECG capability, question type, difficulty, and confusion pairs.

For a selected update path, one round can be summarized as

\begin{equation}
\begin{aligned}
e_t &= \operatorname{Diagnose}(\theta_t,V_{\mathrm{fb}}),\\
q_t &= \operatorname{Plan}(e_t,D_t;B_t),\\
(A_t,L_t,\phi_{t+1}) &= \operatorname{ConstructValidate}(q_t,S_{\mathrm{train}},\phi_t),\\
\widetilde{\theta}_{t+1} &= \operatorname{Train}(\theta_t,A_t\cup R_t),\\
\theta_{t+1} &= \operatorname{Promote}(\theta_t,\widetilde{\theta}_{t+1};V_{\mathrm{sel}}).
\end{aligned}
\label{eq:rsi-loop}
\end{equation}

Here \(B_t\) is the construction budget, \(A_t\) the accepted new examples, \(L_t\) the construction log, and \(R_t\) the replay data. Equation~\eqref{eq:rsi-loop} describes one candidate; sibling recipes use the \textbf{same parent} with separate examples and policy versions. Promotion adds the winning examples to \(D_t\); otherwise, the parent and retained pool remain unchanged.

\subsection{Targeted ECG Synthesis with Measurement-Grounded Supervision}
\label{sec:method:synthesis}

\subsubsection{Targeted Waveform Construction}
\label{sec:method:construction}

\textbf{Forward synthesis} derives questions, answers, and explanations from measured training ECGs. \textbf{Reverse synthesis} edits an eligible waveform to express a requested feature contrast, then remeasures the signal before constructing supervision. Both routes check evidence, context, answer uniqueness, and explanation support.

Reverse requests come from learner weaknesses or distractors in training questions. A normal-versus-prolonged PR contrast specifies a measurable target and a corresponding waveform edit.

\paragraph{Target selection and constrained editing.}
\label{sec:method:pool}

A delineator extracts \(u=f(S)\) from ECG \(S\): heart rate (HR), R--R interval (RR), P-onset-to-QRS-onset interval (PR), QRS duration, QRS-onset-to-T-end interval (QT), corrected QT (QTc), and amplitudes. Feature \(k\) on source \(i\) is scored by \(s_{ik}=a_{ik}r_kd_k\): \textbf{definability} \(a_{ik}\), reliability \(r_k\), and diagnostic utility \(d_k\). Eligible scores exceed the registered threshold. Appendix~\ref{app:parameter-extraction} details conversions, weighting, and scale normalization.

\begin{figure}[!htbp]
\centering
\includegraphics[width=\linewidth,height=0.70\textheight,keepaspectratio]{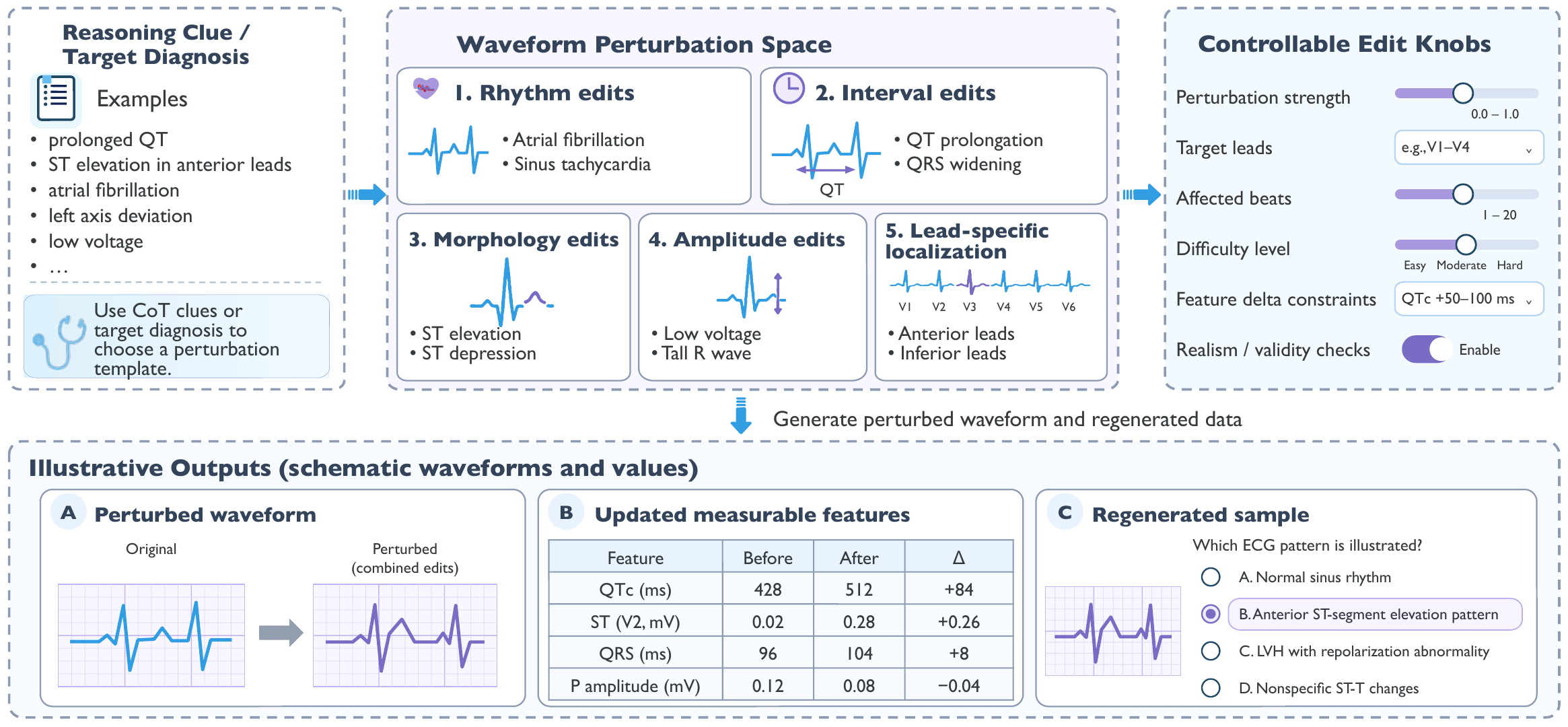}
\caption{Controllable waveform perturbations across rhythm, intervals, morphology, amplitude, and individual leads. The examples illustrate how an intended feature change is translated into a localized signal edit. Edited waveforms are remeasured to check the target and protected features before question generation, linking controllable signal changes to \mbox{evidence-grounded supervision}.}
\label{fig:perturbation-space}
\end{figure}

An eligible request specifies the \textbf{target range}, protected features and tolerances, permitted operators and donors, and attempt budget. The policy applies an operator sequence \(S'=T_\delta(S)\) after checking applicability. PR edits require P and QRS onsets; complex diagnoses require morphology beyond interval changes (Fig.~\ref{fig:perturbation-space}).

\subsubsection{Validation and Generator Repair}
\label{sec:method:anchor}

Re-extract \(u'=f(S')\) after editing. For target feature \(k\) and requested change \(\Delta_k\), record the \textbf{target error}

\[
E_{\mathrm{target}}
=
\left|
\bigl[f_k(S')-f_k(S)\bigr]-\Delta_k
\right|
\]

separately from the drift of each protected feature \(j\),

\[
E_{\mathrm{protect},j}
=
\left|f_j(S')-f_j(S)\right|.
\]

Acceptance requires target and protected-feature errors to satisfy their \textbf{respective tolerances}, with checks of boundary continuity, unedited regions, lead consistency, morphology, leakage, diversity, and cost.

The accepted signal is rendered for question, answer, context, and explanation generation. Review requires a \textbf{unique supported answer} and evidence for each claimed observation, so supervision follows the remeasured signal.

\paragraph{Construction failure and generator repair.}
\label{sec:method:repair}

 Repairs follow failures~\citep{DBLP:journals/corr/abs-2608-29966}: \textbf{signal errors} guide operator revisions; missing morphology guides applicability/retrieval revisions; ambiguity guides question/context revisions; unsupported explanations guide prompt revisions.

Repairs record their trigger, changed rule, validation result, and scope. Old and revised versions are compared on \textbf{matched sources, requests, seeds, and budgets}; adoption requires signal and question checks to pass. Appendices~\ref{app:repair-details} and~\ref{app:waveform-construction-details} describe the rules and recording protocol; Appendix~\ref{app:prompt-templates} provides stage prompts.

\Needspace*{5\baselineskip}
\subsection{Feedback-Driven Recursive Model Improvement}
\label{sec:method:feedback-system}

\subsubsection{Learner Diagnosis and Data Recipes}
\label{sec:method:student-fit}

Errors, invalid outputs, and regressions guide \textbf{recipes} by category \(g\), learner-relative difficulty \(d\), and supervision \(s\):

\begin{equation}
\pi_t^a(g,d,s)=p_t^a(g)\,p_t^a(d\mid g)\,p_t^a(s\mid g,d).
\label{eq:recipe}
\end{equation}

Difficulty is \textbf{learner-relative}~\citep{zhao2025absolutezero}. For training-side question \(i\), \(K_i\) independent answers under fixed decoding settings give \(\widehat p_{it}=K_i^{-1}\sum_{r=1}^{K_i}\mathbf1[\widehat y_{ir}=y_i]\), used to group reliably solved, mixed-success, and unresolved examples. Probe budgets, scoring, output-validity requirements, and thresholds are fixed before branch comparison.

Supervision can retain the teacher explanation, expand intermediate ECG evidence, or add checks targeting an observed learner error. As in iterative reasoning supervision~\citep{zelikman2022star}, explanations supply training targets; paired variants must preserve the \textbf{verified image, question, options, answer, and clinical facts}.

Recipes also fix source/edit constraints, replay, and training budgets. Recipe selection changes the data mixture; operator or prompt repair changes \(\phi_t\). These interventions are tracked separately (Appendix~\ref{app:student-fit-details}).

\subsubsection{Same-Parent Branch Exploration}
\label{sec:method:stage1}

Feedback and training-side probes define three recipe families from the \textbf{same parent} \(\theta_t\)~\citep{zhang2025dgm,zheng2026dreamrsi}:

\begin{itemize}[nosep,leftmargin=*]
\item \textbf{A: Direct reinforcement.} Increase quotas for weak categories and question types while retaining related-category controls: a recipe for atrioventricular (AV) block emphasizes conduction measurements and diagnostic interpretation.
\item \textbf{B: Confusion contrast.} Sample recurrently confused category pairs; construct options and explanations around their distinguishing ECG evidence. Regional ST-elevation myocardial infarction (STEMI) contrasts, for example, emphasize lead-specific localization.
\item \textbf{C: Difficulty and task curriculum.} Reweight difficulty bins and question types using learner probes, e.g., increasing L3, multistep, or evidence-chain questions. Curriculum denotes a branch-specific mixture.
\end{itemize}
Recipes specify \textbf{sampling quotas and construction requests}: retrieve eligible waveforms or apply permitted edits, then run the shared measurement and question-validity gates (Section~\ref{sec:method:anchor}). Category, difficulty, and task constraints can overlap; Tab.~\ref{tab:round-branches} lists each round's actual mixtures.

Each branch retains the same historical training manifest and adds \textbf{10,000 previously unused accepted questions}, without an extra shared baseline set. All branches use three epochs of supervised fine-tuning (SFT) with matched settings; the language model is updated while the vision encoder and multimodal aligner remain frozen. For ECG image \(I\), question \(Q\), optional context \(C\), and explanation--answer target \(y\), the \textbf{training loss} is
\begin{equation}
\mathcal L_{\mathrm{MCQ}}=-\sum_{\ell}\log p_\theta(y_\ell\mid I,Q,C,y_{<\ell}).
\label{eq:mcsq}
\end{equation}
Replay and new-data exposure are recorded in examples and tokens; training settings appear in Appendix~\ref{app:training-config}.

\subsubsection{Model Selection and Retained Lineage}
\label{sec:method:stage2}

\begin{figure}[!htbp]

\centering
\includegraphics[width=0.60\linewidth]{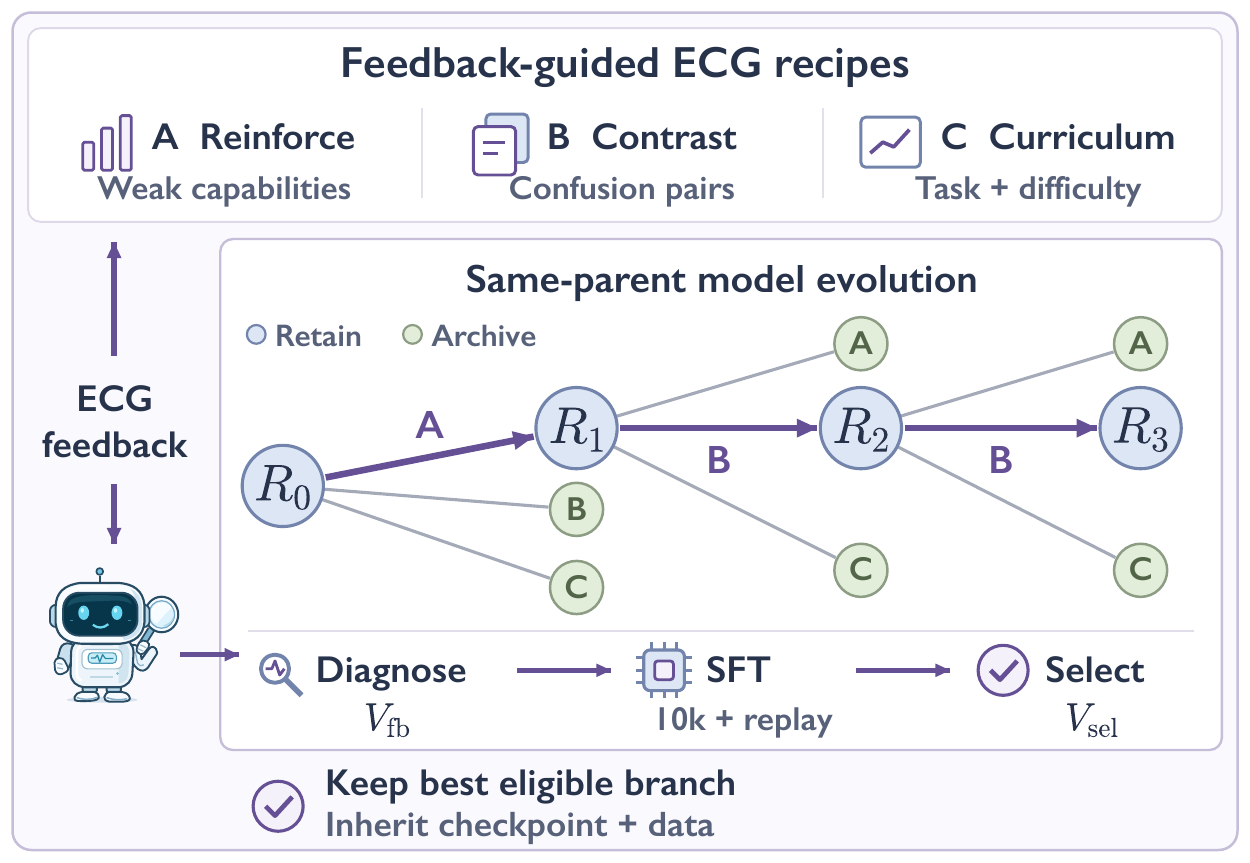}
\caption{\textbf{Schematic of feedback-guided model evolution.} A/B/C change category coverage, confusion contrasts, or difficulty/task mixtures. Each candidate adds 10,000 accepted questions plus replay. Purple shows an illustrative lineage. Tab.~\ref{tab:round-branches} reports measured mixtures, scores, and choices.}
\label{fig:tree-evolution}\label{fig:feedback-selection}
\end{figure}

After matched SFT, evaluate candidates on \(V_{\mathrm{sel}}\). Apply the predesignated \textbf{output-validity and non-target regression tolerances}, then retain the highest-scoring qualifying checkpoint and its data~\citep{supplied2026ref48}. Archive other branches; if none qualifies, retain the parent.

Reevaluate the retained learner on \(V_{\mathrm{fb}}\) to construct new recipes: \textbf{directions are reselected each round}. Fig.~\ref{fig:tree-evolution} illustrates the process; Section~\ref{sec:round-branch-analysis} reports the choices.

\section{Experiments}
\label{sec:experiments}
We evaluate benchmark performance, round-wise gains, and the training value of feedback-guided synthesis. The analysis asks how feedback granularity affects data utility, which candidate directions yield the strongest updates, and whether reductions in \textbf{feedback errors} transfer to external benchmarks.

The seven-benchmark aggregate (S7) is the \textbf{equal-weight mean} of four accuracy and three F1 scores on a 0--100 scale. MMMU-ECG, CSN, and G12EC use single-choice accuracy; ECG-QA uses exact-match accuracy over answer sets. PTB-XL, CPSC2018, and CODE-15\% are multi-label classification tasks. F1 is the harmonic mean of precision and recall; $\Delta$ denotes an absolute score difference. 

\paragraph{Evaluation tasks.}
We evaluate eight ECG tasks comprising 10,840 items. The seven tasks included in S7 contain 10,697 items; the 143-item MedXpertQA-MM ECG subset is reported separately. S7 weights task scores equally and excludes MedX and LUDB. Appendix~\ref{app:evaluation-splits} (Tab.~\ref{tab:evaluation-splits}) lists the datasets, split identifiers, sample counts, and metrics.

\FloatBarrier
\subsection{Main Benchmark Results}

Tab.~\ref{tab:current-main} compares the recorded scores of closed-source, open-source, and ECG-specific models with ECG-RSI on seven benchmarks. ECG-RSI R3 has the \textbf{highest S7 (46.61)} among the compared models.

After three rounds, Qwen3.5-35B-A3B\textquotesingle s mean accuracy across the four question-answering (QA) benchmarks increases from 34.57\% to 56.04\%, a \textbf{21.47-percentage-point gain}; its mean F1 across the three classification benchmarks increases from 24.27 to 34.03, a \textbf{9.77-point gain}. Gains are calculated from unrounded means.

\begin{table}[!htb]
\centering
\begin{minipage}{\linewidth}
\caption{Comparison of ECG-RSI with proprietary, open-source, and ECG-specific models on seven benchmarks. The first four columns report question-answering accuracy and the next three report classification F1; S7 is their equal-weight mean on a 0--100 scale. ECG-RSI R3 achieves the highest aggregate score.}
\label{tab:current-main}
\ecgtablefont
\definecolor{resultbluefill}{HTML}{E4F1FB}
\setlength{\tabcolsep}{1.8pt}
\renewcommand{\arraystretch}{1.15}
\begin{tabularx}{\linewidth}{>{\raggedright\arraybackslash}p{0.22\linewidth}*{8}{>{\centering\arraybackslash}X}}
\toprule
\rowcolor{ecgtablegray}
\textbf{Model / method} & \textbf{\makecell{MMMU\\ECG\\Acc.}} & \textbf{\makecell{CSN\\Acc.}} & \textbf{\makecell{G12EC\\Acc.}} & \textbf{\makecell{ECG-QA\\Acc.}} & \textbf{\makecell{PTB-XL\\F1}} & \textbf{\makecell{CPSC\\2018\\F1}} & \textbf{\makecell{CODE\\15\%\\F1}} & \textbf{S7} \\
\midrule
\rowcolor{ecgtablegray}\multicolumn{9}{c}{\textit{Proprietary models}} \\
\midrule
GPT-5.6-sol & \underline{56.00} & \textbf{58.70} & \textbf{60.20} & \underline{32.10} & \textbf{37.40} & \textbf{24.30} & \textbf{42.80} & \textbf{44.50} \\
Gemini-3.8-Flash & \textbf{59.10} & \underline{57.90} & \underline{59.60} & \textbf{33.40} & \underline{36.80} & \underline{23.70} & \underline{40.90} & \underline{44.49} \\
\midrule
\rowcolor{ecgtablegray}\multicolumn{9}{c}{\textit{Open-source and base models}} \\
\midrule
\ecgmedical\ LINGSHU-7B~\citep{xu2025lingshu} & \underline{47.60} & \underline{45.20} & 46.80 & 26.30 & \underline{33.87} & \underline{18.60} & \underline{36.50} & \underline{36.41} \\
\ecgmedical\ LINGSHU-32B~\citep{xu2025lingshu} & \textbf{51.30} & \textbf{48.90} & \textbf{50.40} & \textbf{28.70} & \textbf{39.45} & \textbf{20.80} & \textbf{38.20} & \textbf{39.68} \\
\ecgmedical\ HuatuoGPT-V-7B~\citep{DBLP:journals/corr/abs-2406-19280} & 44.80 & 41.60 & 43.20 & 25.40 & 31.15 & 17.30 & 33.90 & 33.91 \\
\targetmodel{} & 42.00 & 30.91 & 41.36 & 23.99 & 29.50 & 13.10 & 30.20 & 30.15 \\
\ecgmedical\ ECG-R1~\citep{DBLP:journals/corr/abs-2602-04279} & 38.12 & 44.87 & \underline{47.05} & \underline{28.63} & 22.91 & 17.88 & 34.55 & 33.43 \\
\ecgmedical\ PULSE~\citep{liu2024pulse} & 39.24 & 45.16 & 43.87 & 21.55 & 26.08 & 14.92 & 26.82 & 31.09 \\
\midrule
\rowcolor{ecgtablegray}\multicolumn{9}{c}{\textit{ECG-RSI}} \\
\midrule
\ecgmedical\ ECG-RSI R0 & \underline{42.00} & \underline{30.91} & \underline{41.36} & \underline{23.99} & \underline{29.50} & \underline{13.10} & \underline{30.20} & \underline{30.15} \\
\rowcolor{resultbluefill}
\ecgmedical\ \textbf{ECG-RSI R3} & \textbf{57.50} & \textbf{67.23} & \textbf{66.24} & \textbf{33.18} & \textbf{38.30} & \textbf{22.80} & \textbf{41.00} & \textbf{46.61} \\
\bottomrule
\end{tabularx}
\par\smallskip
{\small\raggedright All scores are on a 0--100 scale. Bold and underlining mark the best and second-best values within each model group; ties share a rank. Blue shading highlights our final model. \ecgmedical\ denotes medical specialization. \targetmodel{} is the initial model (ECG-RSI R0). ECG-R1 and PULSE are ECG-specific comparison models.\par}
\end{minipage}
\end{table}

\subsection{Evolution across Rounds}

Tab.~\ref{tab:current-rounds} and Fig.~\ref{fig:current-main} show the R0--R3 trajectory of series R-A. Each winning branch adds 10,000 new accepted questions to the \textbf{cumulative training set} and replays all prior data. Appendix~\ref{sec:current-data-appendix} provides feedback breakdowns.

\begin{table}[!htbp]
\centering
\caption{Round-wise benchmark performance of series R-A. R0 is the starting model; each accepted update adds 10,000 new questions while retaining historical training data. Bold and underlining mark the best and second-best values within each column; blue/red arrows indicate increases/decreases from the previous round, and blue shading marks R3. Aggregate performance improves through R3, although several individual metrics peak at R2.}
\label{tab:current-rounds}
\ecgtablefont
\let\tabletwoup\ecgup
\let\tabletwodown\ecgdown
\renewcommand{\ecgup}{\rlap{\tabletwoup}}
\renewcommand{\ecgdown}{\rlap{\tabletwodown}}
\setlength{\tabcolsep}{3pt}
\renewcommand{\arraystretch}{1.15}
\begin{tabularx}{\linewidth}{@{}l*{8}{>{\centering\arraybackslash}X}@{}}
\toprule
\rowcolor{ecgtablegray}
R-A round & \makecell{MMMU\\ECG\\Acc.} & \makecell{CSN\\Acc.} & \makecell{G12EC\\Acc.} & \makecell{ECG-QA\\Acc.} & \makecell{PTB-XL\\F1} & \makecell{CPSC\\2018\\F1} & \makecell{CODE\\15\%\\F1} & S7 \\
\midrule
R0 & 42.00 & 30.91 & 41.36 & 23.99 & 29.50 & 13.10 & 30.20 & 30.15 \\
R1 & 48.00\ecgup & 61.68\ecgup & 64.61\ecgup & \underline{33.56}\ecgup & 32.90\ecgup & \underline{23.00}\ecgup & 38.80\ecgup & 43.22\ecgup \\
R2 & \underline{56.00}\ecgup & \underline{64.86}\ecgup & \textbf{66.39}\ecgup & \textbf{34.55}\ecgup & \underline{36.40}\ecgup & \textbf{23.90}\ecgup & \textbf{41.50}\ecgup & \underline{46.23}\ecgup \\
\rowcolor{ecgtableblue}
\textbf{R3} & \textbf{57.50}\ecgup & \textbf{67.23}\ecgup & \underline{66.24}\ecgdown & 33.18\ecgdown & \textbf{38.30}\ecgup & 22.80\ecgdown & \underline{41.00}\ecgdown & \textbf{46.61}\ecgup \\
\midrule
$\Delta$(R3$-$R0) & \ecgdelta{+15.50}\ecgup & \ecgdelta{+36.32}\ecgup & \ecgdelta{+24.88}\ecgup & \ecgdelta{+9.19}\ecgup & \ecgdelta{+8.80}\ecgup & \ecgdelta{+9.70}\ecgup & \ecgdelta{+10.80}\ecgup & \ecgdelta{+16.46}\ecgup \\
\bottomrule
\end{tabularx}
\par\smallskip
{\small\raggedright R0 is the starting model in R-A. $\Delta$(R3$-$R0) reports the total score difference; arrows in R1--R3 indicate changes from the previous round. \par}
\end{table}

Section~\ref{sec:feedback-diagnostics} analyzes \textbf{feedback errors} by ECG capability, question type, and difficulty; Tab.~\ref{tab:current-feedback-breakdown} reports the question-type and difficulty breakdowns, and Tab.~\ref{tab:current-fixed-feedback} reports the ECG-category breakdown.

\subsection{Validation-Set Performance}
MedX denotes the ECG subset of MedXpertQA. LUDB denotes the Lobachevsky University Electrocardiography Database, whose waveform annotations include P-wave, QRS-complex, and T-wave boundaries and peaks. The LUDB accuracy below refers to the ECG evaluation task constructed from that database.

\begin{figure}[!htbp]
\centering
\includegraphics[width=\linewidth]{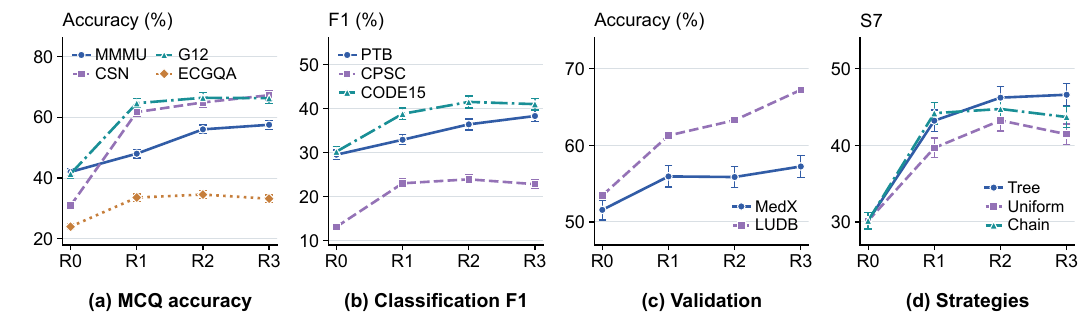}
\caption{Round-wise benchmark, validation, and strategy results. (a) Four multiple-choice question (MCQ) accuracy metrics; (b) three classification F1 metrics; (c) MedX and LUDB validation accuracy; (d) seven-benchmark aggregate score (S7) for tree, uniform-allocation, and chain trajectories. Higher values indicate better performance. Error bars show $\pm1$ standard deviation (SD) where available; LUDB has no SD bars. The panels distinguish broad gains from task-specific fluctuations and the late-round declines of the two strategy controls.}
\label{fig:current-main}
\end{figure}

Tab.~\ref{tab:current-validation} and Fig.~\ref{fig:current-main}(c) show that MedX accuracy rises from \textbf{51.57\% to 57.24\%}, despite a small R2 decline. LUDB accuracy improves at every round, from \textbf{53.45\% to 67.24\%}.

\begin{table}[!htbp]
\centering
\caption{Validation accuracy on MedX and LUDB across four rounds. The change columns report later minus earlier accuracy, in percentage points. Both datasets improve overall; LUDB increases at every round, whereas MedX shows a small \mbox{intermediate decline}.}
\label{tab:current-validation}
\ecgtablefont
\setlength{\tabcolsep}{3pt}
\renewcommand{\arraystretch}{1.15}
\begin{tabularx}{\linewidth}{@{}l*{8}{>{\centering\arraybackslash}X}@{}}
\toprule
 & \multicolumn{4}{c}{Accuracy (\%)} & \multicolumn{4}{c}{Accuracy difference} \\
\cmidrule(lr){2-5}\cmidrule(l){6-9}
\rowcolor{ecgtablegray}
Dataset & R0 & R1 & R2 & \cellcolor{ecgtableblue}R3 & R1$-$R0 & R2$-$R1 & R3$-$R2 & \cellcolor{ecgtableblue}R3$-$R0 \\
\midrule
MedX & 51.57 & 55.94 & 55.86 & \cellcolor{ecgtableblue}\textbf{57.24} & \ecgdelta{+4.37} & \ecgdelta{-0.08} & \ecgdelta{+1.38} & \cellcolor{ecgtableblue}\ecgdelta{\textbf{+5.67}} \\
LUDB & 53.45 & 61.28 & 63.28 & \cellcolor{ecgtableblue}\textbf{67.24} & \ecgdelta{+7.83} & \ecgdelta{+2.00} & \ecgdelta{+3.96} & \cellcolor{ecgtableblue}\ecgdelta{\textbf{+13.79}} \\
\bottomrule
\end{tabularx}
\end{table}

\Needspace*{2\baselineskip}
Fig.~\ref{fig:current-main} summarizes \textbf{round-wise benchmark and validation performance} together with historical strategy trajectories.

\subsection{Feedback Diagnostics across Rounds}
\label{sec:feedback-diagnostics}

Error rates fall from 26.11\% to 12.97\% for L1, from 38.52\% to 24.43\% for L2, and from 58.20\% to 38.17\% for L3. L3 has the largest reduction and \textbf{highest R3 error rate}. Errors keep falling as \textbf{S7 gains diminish} (Section~\ref{sec:discussion}).

Tab.~\ref{tab:current-feedback-breakdown} summarizes error rates by question type and difficulty; Tab.~\ref{tab:current-fixed-feedback} gives the ECG-category breakdown. These \textbf{development-feedback statistics} diagnose data needs. Overall feedback uses \textbf{3,000 questions}, with 1,189, 983, 803, and 640 errors at R0--R3. Appendix~\ref{sec:current-data-appendix} lists the category-level counts.

\par\smallskip\noindent\begin{minipage}{\linewidth}\captionsetup{type=table}\centering
\caption{Feedback error rates (\%) by question type and difficulty across R0--R3. $N$ is the group size; full difficulty-level error counts appear in Tab.~\ref{tab:current-feedback-difficulty-counts}.}\label{tab:current-feedback-breakdown}
\ecgtablefont\setlength{\tabcolsep}{3pt}\renewcommand{\arraystretch}{1.15}
\begin{tabularx}{\linewidth}{@{}l*{6}{>{\centering\arraybackslash}X}@{}}
\toprule
 & \multicolumn{3}{c}{\textbf{Question type}} & \multicolumn{3}{c}{\textbf{Difficulty}} \\
\cmidrule(lr){2-4}\cmidrule(l){5-7}
\rowcolor{ecgtablegray}
\multicolumn{1}{c}{Round} & PM & IA & CD & L1 & L2 & L3 \\
\midrule
$N$ & 1,037 & 1,186 & 777 & 1,218 & 1,093 & 689 \\
R0 & 32.98 & 38.87 & 43.37 & 26.11 & 38.52 & 58.20 \\
R1 & 26.62\ecgerrdown & 32.80\ecgerrdown & 38.74\ecgerrdown & 20.28\ecgerrdown & 33.12\ecgerrdown & 51.81\ecgerrdown \\
R2 & \underline{21.31}\ecgerrdown & \underline{27.57}\ecgerrdown & \underline{33.85}\ecgerrdown & \underline{16.09}\ecgerrdown & \underline{28.27}\ecgerrdown & \underline{44.41}\ecgerrdown \\
\rowcolor{ecgtableblue}
\textbf{R3} & \textbf{17.65}\ecgerrdown & \textbf{23.10}\ecgerrdown & \textbf{29.73}\ecgerrdown & \textbf{12.97}\ecgerrdown & \textbf{24.43}\ecgerrdown & \textbf{38.17}\ecgerrdown \\
\midrule
$\Delta_{0\to3}$ & \textcolor{ecgtrendblue}{15.33}\ecgerrdown & \textcolor{ecgtrendblue}{15.77}\ecgerrdown & \textcolor{ecgtrendblue}{13.64}\ecgerrdown & \textcolor{ecgtrendblue}{13.14}\ecgerrdown & \textcolor{ecgtrendblue}{14.09}\ecgerrdown & \textcolor{ecgtrendblue}{20.03}\ecgerrdown \\
\bottomrule\end{tabularx}
\par\smallskip
{\small\raggedright PM: perception/measurement; IA: interpretation/analysis; CD: clinical decision/treatment. Blue shading marks R3; bold/underlining mark the lowest/second-lowest rates. Blue arrows indicate decreases from the previous round; $\Delta_{0\to3}$ is R0 minus R3 error rate, in percentage points.\par}
\end{minipage}\par

\par\medskip\begin{table}[!htbp]
\centering
\caption{ECG-category feedback error rates (\%). $N$: recorded sample count; $\Delta_{0\to3}$: R0 minus R3 error rate, in percentage points (computed before rounding). Blue marks R3; bold/underline mark the best/second-best rates; $\downarrow$ marks reductions. Full names and counts: Tab.~\ref{tab:current-feedback-categories}.}\label{tab:current-fixed-feedback}
\fontsize{7.5}{9}\selectfont\setlength{\tabcolsep}{0pt}\renewcommand{\arraystretch}{1.2}
\begin{tabularx}{\linewidth}{@{}l*{23}{>{\centering\arraybackslash}X}@{}}
\toprule\rowcolor{ecgtablegray}
Round & \rotatebox{70}{NSR} & \rotatebox{70}{AF} & \rotatebox{70}{AFL} & \rotatebox{70}{ST} & \rotatebox{70}{SB} & \rotatebox{70}{PVC} & \rotatebox{70}{PAC} & \rotatebox{70}{AVB-I} & \rotatebox{70}{Mobitz-I} & \rotatebox{70}{Mobitz-II} & \rotatebox{70}{AVB-III} & \rotatebox{70}{LBBB} & \rotatebox{70}{RBBB} & \rotatebox{70}{Ant-STEMI} & \rotatebox{70}{Inf-STEMI} & \rotatebox{70}{Lat-STEMI} & \rotatebox{70}{Post-STEMI} & \rotatebox{70}{LVH} & \rotatebox{70}{RVH} & \rotatebox{70}{LQTS} & \rotatebox{70}{HyperK} & \rotatebox{70}{WPW} & \rotatebox{70}{Brugada} \\
\midrule
$N$ & 168 & 154 & 147 & 132 & 128 & 143 & 136 & 121 & 109 & 105 & 118 & 139 & 141 & 149 & 145 & 116 & 108 & 152 & 124 & 134 & 117 & 113 & 101 \\
R0 & 25.0 & 39.6 & 44.9 & 28.0 & 26.6 & 40.6 & 39.0 & 38.0 & 42.2 & 48.6 & 44.1 & 36.7 & 35.5 & 44.3 & 43.4 & 46.6 & 49.1 & 37.5 & 41.1 & 41.0 & 42.7 & 41.6 & 45.5 \\
R1 & 19.6\textcolor{ecgtrendblue}{\ensuremath{\scriptstyle\downarrow}} & 32.5\textcolor{ecgtrendblue}{\ensuremath{\scriptstyle\downarrow}} & 38.1\textcolor{ecgtrendblue}{\ensuremath{\scriptstyle\downarrow}} & 22.0\textcolor{ecgtrendblue}{\ensuremath{\scriptstyle\downarrow}} & 20.3\textcolor{ecgtrendblue}{\ensuremath{\scriptstyle\downarrow}} & 33.6\textcolor{ecgtrendblue}{\ensuremath{\scriptstyle\downarrow}} & 32.4\textcolor{ecgtrendblue}{\ensuremath{\scriptstyle\downarrow}} & 31.4\textcolor{ecgtrendblue}{\ensuremath{\scriptstyle\downarrow}} & 34.9\textcolor{ecgtrendblue}{\ensuremath{\scriptstyle\downarrow}} & 41.0\textcolor{ecgtrendblue}{\ensuremath{\scriptstyle\downarrow}} & 36.4\textcolor{ecgtrendblue}{\ensuremath{\scriptstyle\downarrow}} & 30.2\textcolor{ecgtrendblue}{\ensuremath{\scriptstyle\downarrow}} & 29.1\textcolor{ecgtrendblue}{\ensuremath{\scriptstyle\downarrow}} & 37.6\textcolor{ecgtrendblue}{\ensuremath{\scriptstyle\downarrow}} & 36.6\textcolor{ecgtrendblue}{\ensuremath{\scriptstyle\downarrow}} & 38.8\textcolor{ecgtrendblue}{\ensuremath{\scriptstyle\downarrow}} & 40.7\textcolor{ecgtrendblue}{\ensuremath{\scriptstyle\downarrow}} & 30.9\textcolor{ecgtrendblue}{\ensuremath{\scriptstyle\downarrow}} & 33.9\textcolor{ecgtrendblue}{\ensuremath{\scriptstyle\downarrow}} & 34.3\textcolor{ecgtrendblue}{\ensuremath{\scriptstyle\downarrow}} & 35.9\textcolor{ecgtrendblue}{\ensuremath{\scriptstyle\downarrow}} & 34.5\textcolor{ecgtrendblue}{\ensuremath{\scriptstyle\downarrow}} & 37.6\textcolor{ecgtrendblue}{\ensuremath{\scriptstyle\downarrow}} \\
R2 & \underline{14.9}\textcolor{ecgtrendblue}{\ensuremath{\scriptstyle\downarrow}} & \underline{26.6}\textcolor{ecgtrendblue}{\ensuremath{\scriptstyle\downarrow}} & \underline{32.0}\textcolor{ecgtrendblue}{\ensuremath{\scriptstyle\downarrow}} & \underline{16.7}\textcolor{ecgtrendblue}{\ensuremath{\scriptstyle\downarrow}} & \underline{15.6}\textcolor{ecgtrendblue}{\ensuremath{\scriptstyle\downarrow}} & \underline{27.3}\textcolor{ecgtrendblue}{\ensuremath{\scriptstyle\downarrow}} & \underline{26.5}\textcolor{ecgtrendblue}{\ensuremath{\scriptstyle\downarrow}} & \underline{25.6}\textcolor{ecgtrendblue}{\ensuremath{\scriptstyle\downarrow}} & \underline{28.4}\textcolor{ecgtrendblue}{\ensuremath{\scriptstyle\downarrow}} & \underline{34.3}\textcolor{ecgtrendblue}{\ensuremath{\scriptstyle\downarrow}} & \underline{29.7}\textcolor{ecgtrendblue}{\ensuremath{\scriptstyle\downarrow}} & \underline{24.5}\textcolor{ecgtrendblue}{\ensuremath{\scriptstyle\downarrow}} & \underline{23.4}\textcolor{ecgtrendblue}{\ensuremath{\scriptstyle\downarrow}} & \underline{31.5}\textcolor{ecgtrendblue}{\ensuremath{\scriptstyle\downarrow}} & \underline{30.3}\textcolor{ecgtrendblue}{\ensuremath{\scriptstyle\downarrow}} & \underline{31.9}\textcolor{ecgtrendblue}{\ensuremath{\scriptstyle\downarrow}} & \underline{34.3}\textcolor{ecgtrendblue}{\ensuremath{\scriptstyle\downarrow}} & \underline{25.0}\textcolor{ecgtrendblue}{\ensuremath{\scriptstyle\downarrow}} & \underline{27.4}\textcolor{ecgtrendblue}{\ensuremath{\scriptstyle\downarrow}} & \underline{28.4}\textcolor{ecgtrendblue}{\ensuremath{\scriptstyle\downarrow}} & \underline{29.9}\textcolor{ecgtrendblue}{\ensuremath{\scriptstyle\downarrow}} & \underline{28.3}\textcolor{ecgtrendblue}{\ensuremath{\scriptstyle\downarrow}} & \underline{30.7}\textcolor{ecgtrendblue}{\ensuremath{\scriptstyle\downarrow}} \\
\rowcolor{ecgtableblue}
R3 & \textbf{11.3}\textcolor{ecgtrendblue}{\ensuremath{\scriptstyle\downarrow}} & \textbf{21.4}\textcolor{ecgtrendblue}{\ensuremath{\scriptstyle\downarrow}} & \textbf{26.5}\textcolor{ecgtrendblue}{\ensuremath{\scriptstyle\downarrow}} & \textbf{12.9}\textcolor{ecgtrendblue}{\ensuremath{\scriptstyle\downarrow}} & \textbf{11.7}\textcolor{ecgtrendblue}{\ensuremath{\scriptstyle\downarrow}} & \textbf{21.7}\textcolor{ecgtrendblue}{\ensuremath{\scriptstyle\downarrow}} & \textbf{21.3}\textcolor{ecgtrendblue}{\ensuremath{\scriptstyle\downarrow}} & \textbf{19.8}\textcolor{ecgtrendblue}{\ensuremath{\scriptstyle\downarrow}} & \textbf{22.0}\textcolor{ecgtrendblue}{\ensuremath{\scriptstyle\downarrow}} & \textbf{27.6}\textcolor{ecgtrendblue}{\ensuremath{\scriptstyle\downarrow}} & \textbf{23.7}\textcolor{ecgtrendblue}{\ensuremath{\scriptstyle\downarrow}} & \textbf{19.4}\textcolor{ecgtrendblue}{\ensuremath{\scriptstyle\downarrow}} & \textbf{18.4}\textcolor{ecgtrendblue}{\ensuremath{\scriptstyle\downarrow}} & \textbf{26.2}\textcolor{ecgtrendblue}{\ensuremath{\scriptstyle\downarrow}} & \textbf{24.8}\textcolor{ecgtrendblue}{\ensuremath{\scriptstyle\downarrow}} & \textbf{25.9}\textcolor{ecgtrendblue}{\ensuremath{\scriptstyle\downarrow}} & \textbf{27.8}\textcolor{ecgtrendblue}{\ensuremath{\scriptstyle\downarrow}} & \textbf{19.7}\textcolor{ecgtrendblue}{\ensuremath{\scriptstyle\downarrow}} & \textbf{21.8}\textcolor{ecgtrendblue}{\ensuremath{\scriptstyle\downarrow}} & \textbf{22.4}\textcolor{ecgtrendblue}{\ensuremath{\scriptstyle\downarrow}} & \textbf{23.9}\textcolor{ecgtrendblue}{\ensuremath{\scriptstyle\downarrow}} & \textbf{22.1}\textcolor{ecgtrendblue}{\ensuremath{\scriptstyle\downarrow}} & \textbf{23.8}\textcolor{ecgtrendblue}{\ensuremath{\scriptstyle\downarrow}} \\
\midrule
$\Delta_{0\to3}\!\downarrow$ & \textcolor{ecgtrendblue}{13.7} & \textcolor{ecgtrendblue}{18.2} & \textcolor{ecgtrendblue}{18.4} & \textcolor{ecgtrendblue}{15.2} & \textcolor{ecgtrendblue}{14.8} & \textcolor{ecgtrendblue}{18.9} & \textcolor{ecgtrendblue}{17.6} & \textcolor{ecgtrendblue}{18.2} & \textcolor{ecgtrendblue}{20.2} & \textcolor{ecgtrendblue}{21.0} & \textcolor{ecgtrendblue}{20.3} & \textcolor{ecgtrendblue}{17.3} & \textcolor{ecgtrendblue}{17.0} & \textcolor{ecgtrendblue}{18.1} & \textcolor{ecgtrendblue}{18.6} & \textcolor{ecgtrendblue}{20.7} & \textcolor{ecgtrendblue}{21.3} & \textcolor{ecgtrendblue}{17.8} & \textcolor{ecgtrendblue}{19.4} & \textcolor{ecgtrendblue}{18.7} & \textcolor{ecgtrendblue}{18.8} & \textcolor{ecgtrendblue}{19.5} & \textcolor{ecgtrendblue}{21.8} \\
\bottomrule\end{tabularx}

\end{table}\par\medskip

\section{Ablation Studies}
\label{sec:current-ablations}
We first assess \textbf{waveform validity} and the contribution of synthesis components. We then examine three linked questions: whether finer feedback improves \textbf{training utility}, whether the most prominent weakness determines the best training direction, and whether continued feedback improvement translates into external gains.

\subsection{Effectiveness of the Synthesis Method}
\paragraph{Quantitative waveform-feature validation.}
We report \textbf{waveform-feature error statistics}, including mean absolute error (MAE) and root mean squared error (RMSE), from a single synthesis run of 10,000 samples in Tab.~\ref{tab:current-waveform-metrics}.

\begin{table}[!htbp]
\centering
\caption{Waveform-feature errors from a single synthesis run of 10,000 samples and comparison with the adopted MAE criteria. Blue marks MAE values meeting the adopted criteria.}
\label{tab:current-waveform-metrics}
\ecgtablefont
\setlength{\tabcolsep}{2.5pt}
\renewcommand{\arraystretch}{1.15}
\begin{tabularx}{\linewidth}{@{}ll*{4}{>{\centering\arraybackslash}X}cc@{}}
\toprule
\rowcolor{ecgtablegray}
Feature & Unit & Bias & MAE & RMSE & $s_e^2$ & \makecell{MAE\\limit} & Status \\
\midrule
HR & bpm & 0.63 & \cellcolor{ecgtableblue}\textbf{2.14} & 2.22 & 4.53 & $\leq 3$ & \textcolor{ecgtrendblue}{Met} \\
RR & ms & 3.11 & \cellcolor{ecgtableblue}\textbf{15.28} & 15.36 & 226.28 & $<20$ & \textcolor{ecgtrendblue}{Met} \\
PR & ms & 0.57 & \cellcolor{ecgtableblue}\textbf{7.71} & 7.78 & 60.21 & $\leq 25$ & \textcolor{ecgtrendblue}{Met} \\
QRS & ms & 4.12 & \cellcolor{ecgtableblue}\textbf{4.58} & 4.61 & 4.28 & $\leq 30$ & \textcolor{ecgtrendblue}{Met} \\
QT & ms & 0.94 & \cellcolor{ecgtableblue}\textbf{9.71} & 9.85 & 96.15 & $\leq 30$ & \textcolor{ecgtrendblue}{Met} \\
QTc & ms & 0.86 & \cellcolor{ecgtableblue}\textbf{10.22} & 10.34 & 106.19 & $\leq 40$ & \textcolor{ecgtrendblue}{Met} \\
\bottomrule
\end{tabularx}
\par\smallskip
{\small\raggedright Error is remeasured minus target. Bias, MAE, RMSE, and limits use the listed unit; $s_e^2$ uses squared units and denominator $n-1$. Limits are the evaluation criteria adopted for this experiment. Variance values are approximations reconstructed from the reported two-decimal Bias and RMSE, with $n=10{,}000$ valid observations per feature: $s_e^2=\frac{n}{n-1}(\mathrm{RMSE}^2-\mathrm{Bias}^2)$.\par}
\end{table}

All six features meet the adopted MAE criteria (Status = Met), supporting numerical fidelity of the synthesized waveforms.

For each feature, let $e_i=\hat x_i-x_i$ and compute
\begin{equation}
\overline e=\frac{1}{n}\sum_i e_i,\quad
\mathrm{MAE}=\frac{1}{n}\sum_i|e_i|,\quad
\mathrm{RMSE}=\sqrt{\frac{1}{n}\sum_i e_i^2},\quad
s_e^2=\frac{1}{n-1}\sum_i(e_i-\overline e)^2.
\label{eq:current-waveform-error}
\end{equation}

\Needspace*{6\baselineskip}
\subsection{Blinded Review by Human Physicians}
\label{app:clinical-review}

We randomly sample 240 examples stratified by six edit types (HR, RR, PR, QRS, QT, and QTc), with 40 examples per type. \textbf{Two human physicians} score the examples \textbf{independently}. We report 95\% confidence intervals (CIs) for the review metrics.

\par
\begin{table}[!htbp]
\centering

\centering
\caption{Physician review of 240 examples (95\% confidence intervals). Stages 1--3 assess morphology, editing fidelity, and answers/explanations.}
\label{tab:clinical-review-overall}
\ecgtablefont
\setlength{\tabcolsep}{3.5pt}
\renewcommand{\arraystretch}{1.15}
\begin{tabularx}{\linewidth}{@{}>{\raggedright\arraybackslash}Xcccl@{}}
\toprule
\rowcolor{ecgtablegray}
Metric & Value & 95\% CI & $n$ & Source \\
\midrule
Target attainment & \textbf{0.82} & [0.76, 0.87] & 240 & Stage 2 \\
Non-target preservation & \textbf{0.79} & [0.73, 0.84] & 240 & Stage 2 \\
Morphology consistency & \textbf{0.86} & [0.81, 0.90] & 240 & Stage 1 \\
Lead consistency & \textbf{0.83} & [0.78, 0.88] & 240 & Stage 1 \\
\rowcolor{ecgtableblue}
Answer accuracy & \textbf{0.88} & [0.83, 0.92] & 240 & Stage 3 \\
\rowcolor{ecgtableblue}
Explanation support & \textbf{0.76} & [0.70, 0.81] & 240 & Stage 3 \\
Expected-parameter references & \textbf{0.09} & [0.06, 0.13] & 240 & Stage 3 \\
Inter-rater $\kappa$ & \textbf{0.78} & [0.71, 0.85] & 240 & Two raters \\
\bottomrule
\end{tabularx}
\par\smallskip
{\small\raggedright CIs use Wilson intervals or bootstrap; $\kappa$ is Cohen's $\kappa$. Lower expected-parameter reference rates are better; higher values are better for the other metrics.\par}
\end{table}

In stage 1, target labels, generation parameters, and teacher responses are hidden while waveform morphology and lead consistency are assessed. In stage 2, the physicians inspect the before/after waveforms and editing targets to assess target attainment and non-target preservation. Stage 3 assesses answers, explanations, and expected-parameter references.

\par\smallskip
\noindent
\textbf{Answer correctness and evidence support.} Mean physician ratings are 0.88 for answer correctness and 0.76 for explanation support ($\kappa=0.78$). The stages assess morphology, editing fidelity, and answer/explanation quality.

\par
\begin{table}[!htbp]
\centering

\centering
\caption{Clinical review by edit type: 40 examples per type, 240 overall. Values are aggregated physician ratings; higher is better.}
\label{tab:clinical-review-stratified}
\ecgtablefont
\setlength{\tabcolsep}{2.5pt}
\renewcommand{\arraystretch}{1.15}
\begin{tabularx}{\linewidth}{@{}lc*{5}{>{\centering\arraybackslash}X}@{}}
\toprule
\rowcolor{ecgtablegray}
Edit & $n$ & Target & Preserve & Morph. & Answer & Support \\
\midrule
HR & 40 & 0.85 & 0.80 & 0.88 & 0.90 & 0.78 \\
RR & 40 & 0.78 & 0.75 & 0.82 & 0.85 & 0.72 \\
PR & 40 & 0.80 & 0.78 & 0.85 & 0.88 & 0.75 \\
QRS & 40 & 0.83 & 0.81 & 0.87 & 0.89 & 0.77 \\
QT & 40 & 0.82 & 0.79 & 0.86 & 0.87 & 0.76 \\
QTc & 40 & 0.84 & 0.80 & 0.88 & 0.89 & 0.78 \\
\midrule
\rowcolor{ecgtableblue}
\textbf{Total} & \textbf{240} & \textbf{0.82} & \textbf{0.79} & \textbf{0.86} & \textbf{0.88} & \textbf{0.76} \\
\bottomrule
\end{tabularx}
\end{table}
\noindent\textbf{Variation across edit types.} Target attainment ranges from 0.78 for RR to 0.85 for HR. Explanation support varies from 0.72 to 0.78 and remains below answer accuracy for every edit type. The stratified results show that the gap is shared across editing targets.

\noindent The columns distinguish target attainment, preservation of non-target features, morphology consistency, answer accuracy, and explanation support.

\subsection{Multi-Model Question-Quality Scores}
\label{app:question-quality-scores}
\begin{figure}[!htbp]
\centering
\includegraphics[width=\linewidth]{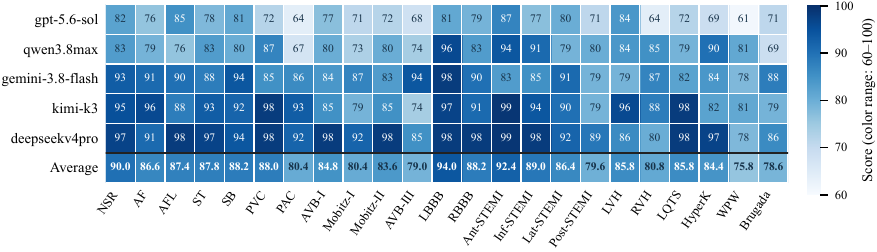}
\caption{Recorded scores from five models for synthetic ECG questions across 23 categories (0--100 scale). The bottom row is the per-category arithmetic mean \mbox{across models}; the unweighted mean over all 115 model--category cells is 85.09, reported as the full-method Question score in Tab.~\ref{tab:current-synthesis-ablation}. The shared color scale spans 60--100, from light blue (low) to dark blue (high).}
\label{fig:five-model-score-demo}\label{fig:multimodel-score-heatmap}
\end{figure}

Fig.~\ref{fig:five-model-score-demo} shows \textbf{five-model scores} across 23 ECG categories. R1 uses 10,000 of 16,193 candidate questions.

\Needspace{4\baselineskip}
\subsection{Data Construction and Feedback Selection}
\label{sec:data-construction-ablation}

Four data configurations each add 10,000 SFT questions from the same checkpoint (S7 = 43.22).

\Needspace{4\baselineskip}
\begin{table}[!htbp]

\centering
\caption{Data-construction and selection ablation from a shared checkpoint with S7 = 43.22.}
\label{tab:data-construction-ablation}
\ecgtablefont
\setlength{\tabcolsep}{4pt}
\renewcommand{\arraystretch}{1.22}
\begin{tabularx}{\linewidth}{@{}l>{\raggedright\arraybackslash}Xcc@{}}
\toprule
\rowcolor{ecgtablegray}
Group & Added data & S7 & $\Delta$S7 \\
\midrule
A & Ordinary training-side questions & 44.15 & \ecgdelta{+0.93} \\
B & Forward-only synthesis & 44.27 & \ecgdelta{+1.05} \\
C & Forward + reverse, fixed selection proportions & 45.56 & \ecgdelta{+2.34} \\
\rowcolor{ecgtableblue}
D & Forward + reverse, feedback-guided selection & \textbf{46.23} & \ecgdelta{\textbf{+3.01}} \\
\bottomrule
\end{tabularx}
\end{table}
\noindent
Tab.~\ref{tab:data-construction-ablation} separates ordinary questions, forward synthesis, added reverse synthesis, and \textbf{feedback-guided selection}. $\Delta$S7 is relative to the shared checkpoint. Ordinary questions and forward-only synthesis yield similar scores (44.15 and 44.27). Adding reverse synthesis raises S7 to \textbf{45.56}; feedback-guided selection further raises it to \textbf{46.23}, \mbox{the largest gain}.
\par\medskip

\subsection{Synthesis-Component Ablation}
\par\medskip\noindent Tab.~\ref{tab:current-synthesis-ablation} compares synthesis updates at a fixed target of 10,000 accepted questions. The full method requires 16,193 candidates to reach this target, versus 36,723 with operator updates frozen and 22,398 with skill updates frozen (2.27 and 1.38 times as many, respectively). Freezing updates reduces question-quality scores and downstream R1 S7.

\begin{table}[!htbp]
\centering
\centering
\caption{R1 synthesis efficiency and training outcomes at a fixed target of 10,000 accepted questions. Candidate counts include rejected questions, and the pass rate measures the fraction accepted. Question scores assess generated-data quality, whereas S7 measures downstream model performance; $\Delta$S7 is relative to the full method. Freezing operator or skill updates increases candidate requirements and reduces both quality scores and downstream performance.}
\label{tab:current-synthesis-ablation}
\ecgtablefont
\setlength{\tabcolsep}{2pt}
\renewcommand{\arraystretch}{1.15}
\begin{tabularx}{\linewidth}{@{}>{\raggedright\arraybackslash}p{0.28\linewidth}*{5}{>{\centering\arraybackslash}X}@{}}
\toprule
\rowcolor{ecgtablegray}
Method & \makecell{Candidates\\required} & \makecell{Pass\\(\%)} & \makecell{Question\\score} & S7 & $\Delta$S7 \\
\midrule
\rowcolor{ecgtableblue}
Full method & \textbf{16,193} & \textbf{61.76} & \textbf{85.09} & \textbf{43.22} & 0.00 \\
Freeze operator updates & 36,723 & 27.23 & 71.90 & 39.29 & \ecgdelta{-3.93} \\
Freeze skill updates & 22,398 & 44.65 & 79.20 & 40.15 & \ecgdelta{-3.07} \\
\bottomrule
\end{tabularx}
\par\smallskip
{\small\raggedright Each method trains on 10,000 accepted questions. The full-method Question score (85.09) is the unweighted arithmetic mean of the 115 model--category scores (five models $\times$ 23 categories) in Fig.~\ref{fig:five-model-score-demo}, using the same question set and scoring rubric. Candidates include rejections; $\Delta$S7 compares with the full method.\par}
\end{table}
\par\medskip

\subsection{Feedback Granularity and Training Utility}
\label{sec:feedback-granularity}
To isolate \textbf{feedback resolution}, Tab.~\ref{tab:feedback-granularity} compares four selection configurations from the \textbf{same parent} (S7 = 43.22), each adding 10,000 accepted questions. Only the feedback dimensions used for selection vary.

\Needspace{4\baselineskip}
\begin{table}[!htbp]

\centering
\caption{Feedback granularity under a common parent and a fixed budget of 10,000 added questions per configuration. $\Delta$S7 is relative to the parent score of 43.22. The final column isolates the gain over fixed selection.}
\label{tab:feedback-granularity}
\ecgtablefont
\setlength{\tabcolsep}{4pt}
\renewcommand{\arraystretch}{1.05}
\begin{tabularx}{\linewidth}{@{}>{\raggedright\arraybackslash}Xccc@{}}
\toprule
\rowcolor{ecgtablegray}
Feedback used for selection & S7 & $\Delta$S7 & \makecell{$\Delta$S7 vs.\\fixed} \\
\midrule
None: fixed mixture & 45.56 & \ecgdelta{+2.34} & 0.00 \\
ECG category & 45.60 & \ecgdelta{+2.38} & \ecgdelta{+0.04} \\
Category + question type & 45.98 & \ecgdelta{+2.76} & \ecgdelta{+0.42} \\
\rowcolor{ecgtableblue}
Category + type + difficulty + confusion pairs & \textbf{46.23} & \ecgdelta{\textbf{+3.01}} & \ecgdelta{\textbf{+0.67}} \\
\bottomrule
\end{tabularx}
\end{table}

The fixed-mixture and four-dimensional configurations share their endpoints with Tab.~\ref{tab:data-construction-ablation}. Category-only feedback yields S7 = 45.60, close to 45.56 with fixed selection. Adding question type raises S7 to 45.98; including difficulty and confusion pairs raises it further to \textbf{46.23}. With the same added-data volume, resolving these finer capability gaps produces more useful training examples.

\subsection{Round-Wise Branch Selection and Data Composition}
\label{sec:round-branch-analysis}
Feedback-guided branch evolution compares candidate interventions at each round and retains the best qualifying update on a separate selection set. Tab.~\ref{tab:round-branches} reports all three same-parent candidates at each transition. Each candidate adds 10,000 accepted questions with matched historical replay and training settings. Selection uses $S_{\mathrm{sel}}$ on $V_{\mathrm{sel}}$, distinct from external S7.

\begin{table}[!htbp]
\centering
\caption{Actual candidate compositions and branch choices across three rounds. A, B, and C denote direct reinforcement, confusion contrast, and difficulty/task curriculum. Exactly one branch is retained per round. Percentages describe overlapping attributes of each branch's 10,000 examples. Pair percentages cover the listed pair families jointly.}
\label{tab:round-branches}
\ecgtablefont
\setlength{\tabcolsep}{3pt}
\renewcommand{\arraystretch}{1.15}
\begin{tabularx}{\linewidth}{@{}ll>{\raggedright\arraybackslash}Xcl@{}}
\toprule
\rowcolor{ecgtablegray}
Round & Branch & New-data composition & $S_{\mathrm{sel}}$ & Decision \\
\midrule
\rowcolor{ecgtableblue}
R0$\to$R1 & A & AV block 45\%; L3 50\%; measurement questions 60\% & \textbf{42.66} & \ecgpass\ Retain \\
 & B & Mobitz I/II and LBBB/RBBB contrasts 55\%; L3 40\% & 40.51 & Archive \\
 & C & L2+L3 60\%; clinical-reasoning questions 50\% & 38.27 & Archive \\
\midrule
R1$\to$R2 & A & Four regional STEMI categories 50\%; L3 55\%; image-localization questions 45\% & 44.12 & Archive \\
\rowcolor{ecgtableblue}
 & B & Anterior/lateral and inferior/posterior STEMI contrasts 60\%; L3 45\% & \textbf{45.08} & \ecgpass\ Retain \\
 & C & L3 70\%; multistep-reasoning questions 55\% & 43.55 & Archive \\
\midrule
R2$\to$R3 & A & AF/AFL/PAC/PVC categories 45\%; L3 50\%; rhythm-reading questions 55\% & 45.90 & Archive \\
\rowcolor{ecgtableblue}
 & B & AF/AFL and PAC/PVC contrasts 55\%; L3 45\% & \textbf{46.03} & \ecgpass\ Retain \\
 & C & L3 65\%; evidence-chain explanation questions 50\% & 44.71 & Archive \\
\bottomrule
\end{tabularx}
\par\smallskip
{\footnotesize\raggedright AV: atrioventricular; LBBB/RBBB: left/right bundle branch block; STEMI: ST-segment elevation myocardial infarction; AF/AFL: atrial fibrillation/flutter; PAC/PVC: premature atrial/ventricular contraction.\par}
\end{table}

\paragraph{From weak-category coverage to discriminative evidence.}
\textbf{Direct reinforcement is preferred in the first comparison}, despite a 52\% Mobitz I/II confusion error in the parent. Branch A reaches $S_{\mathrm{sel}}=42.66$, compared with 40.51 for the contrast-focused branch. Its emphasis on AV-block coverage and measurement is followed by reductions in L3 and Mobitz II errors (Tab.~\ref{tab:retained-data-outcomes}). A prominent confusion therefore does not, by itself, identify the best candidate intervention in this comparison.

\textbf{The later comparisons favor confusion-focused supervision}. Regional STEMI contrasts score 45.08 in R1$\to$R2, versus 44.12 for direct reinforcement and 43.55 for curriculum adjustment. Rhythm contrasts score 46.03 in R2$\to$R3, versus 45.90 and 44.71 for the alternatives. L3 errors remain substantial throughout these comparisons, but the curriculum-focused candidate is not selected in either comparison.

\begin{table}[!htbp]
\centering
\caption{Data retained along the selected lineage and subsequent feedback outcomes. Counts are derived from the winning branch percentages and its 10,000 accepted questions. Category/pair counts overlap with L3 counts. Blue shading marks the final update; bold counts identify the targeted data, and blue values mark the updated error rates.}
\label{tab:retained-data-outcomes}
\ecgtablefont
\setlength{\tabcolsep}{3pt}
\renewcommand{\arraystretch}{1.15}
\begin{tabularx}{\linewidth}{@{}l>{\raggedright\arraybackslash}X>{\raggedright\arraybackslash}X@{}}
\toprule
\rowcolor{ecgtablegray}
Update & Retained data & Error rate (\%): before $\to$ after \\
\midrule
R0$\to$R1 (A) & \textbf{AV block: 4,500} (45\%); L3: 5,000 (50\%); measurement: 6,000 (60\%) & L3: 58.20$\to$\textcolor{ecgtrendblue}{51.81}; Mobitz II: 48.57$\to$\textcolor{ecgtrendblue}{40.95} \\
R1$\to$R2 (B) & \textbf{Regional STEMI contrasts: 6,000} (60\%); L3: 4,500 (45\%) & L3: 51.81$\to$\textcolor{ecgtrendblue}{44.41}; posterior STEMI: 40.74$\to$\textcolor{ecgtrendblue}{34.26} \\
\rowcolor{ecgtableblue}
\textbf{R2$\to$R3 (B)} & \textbf{AF/AFL and PAC/PVC contrasts: 5,500} (55\%); L3: 4,500 (45\%) & L3: 44.41$\to$\textcolor{ecgtrendblue}{38.17}; AF: 26.62$\to$\textcolor{ecgtrendblue}{21.43}; AFL: 31.97$\to$\textcolor{ecgtrendblue}{26.53} \\
\bottomrule
\end{tabularx}
\end{table}

Tab.~\ref{tab:retained-data-outcomes} connects the selected recipes to subsequent feedback outcomes. Both later recipes retain 4,500 L3 questions while changing the targeted diagnostic contrasts, illustrating that difficulty and content are overlapping recipe choices.

\subsection{Comparison of RSI Evolution Strategies}
Tab.~\ref{tab:current-strategy} and~\ref{tab:current-strategy-summary} compare tree, uniform-allocation, and chain trajectories in R-A. Tree finishes at \textbf{S7 = 46.61}, ahead of uniform (41.45), chain (43.70), and chain\textquotesingle s best round (44.75).

\begin{table}[!htbp]
\centering
\caption{S7 trajectories in R-A. The last three columns report total gain, the final-round change, and the final gap to tree. Blue shading marks tree; all differences use the S7 scale.}
\label{tab:current-strategy}
\ecgtablefont
\setlength{\tabcolsep}{2pt}
\renewcommand{\arraystretch}{1.15}
\begin{tabularx}{\linewidth}{@{}l*{7}{>{\centering\arraybackslash}X}@{}}
\toprule
\rowcolor{ecgtablegray}
Strategy & R0 & R1 & R2 & R3 & \makecell{R3$-$R0} & \makecell{R3$-$R2} & \makecell{R3 vs.\\tree} \\
\midrule
\rowcolor{ecgtableblue}
Tree & 30.15 & 43.22 & \textbf{46.23} & \textbf{46.61} & \ecgdelta{\textbf{+16.46}} & \ecgdelta{+0.38} & 0.00 \\
Uniform & 30.15 & 39.67 & 43.25 & 41.45 & \ecgdelta{+11.30} & \ecgdelta{-1.80} & \ecgdelta{-5.16} \\
Chain & 30.15 & \textbf{44.20} & 44.75 & 43.70 & \ecgdelta{+13.55} & \ecgdelta{-1.05} & \ecgdelta{-2.91} \\
\bottomrule
\end{tabularx}
\par\smallskip
\end{table}
All three strategies improve over R0. In the final round, tree rises from 46.23 to \textbf{46.61}, while uniform falls from 43.25 to 41.45 and chain from 44.75 to 43.70. Chain leads R1; \textbf{tree finishes highest}.

\begin{table}[!htbp]
\centering
\caption{Synthesis-update ablations and historical strategy trajectories (S7, higher is better). Left: R1 mean $\pm$ standard deviation; $\Delta$ is relative to the full method. Right: round-wise S7; $\Delta_{0\to3}$ is the total improvement. Blue/red arrows denote increases/decreases from the reference or previous round. Bold/underlining mark the best/second-best scores in each comparison. Blue shading highlights the full method and R3.}
\label{tab:current-strategy-summary}
\label{app:strategy-plot}
\begin{minipage}[t]{0.49\linewidth}\vspace{0pt}
\centering\ecgtablefont
\textbf{Synthesis updates (R1)}\par\smallskip
\setlength{\tabcolsep}{3pt}\renewcommand{\arraystretch}{1.15}
\begin{tabularx}{\linewidth}{@{}>{\raggedright\arraybackslash}Xcc@{}}
\toprule\rowcolor{ecgtablegray}
Method & S7 (mean $\pm$ SD) & $\Delta$ \\
\midrule\rowcolor{ecgtableblue}
Full method & $\mathbf{43.22}\pm1.40$ & 0.00 \\
Freeze operators & $39.29\pm1.30$ & $\ecgdelta{-3.93}$\ecgdown \\
Freeze skills & $\underline{40.15}\pm1.30$ & $\ecgdelta{-3.07}$\ecgdown \\
\bottomrule
\end{tabularx}
\par\smallskip
{\small\raggedright R1 S7 falls from 43.22 to 39.29 with frozen operators and to 40.15 with frozen skills. See Tab.~\ref{tab:current-synthesis-ablation} for synthesis efficiency.\par}
\end{minipage}\hfill
\begin{minipage}[t]{0.48\linewidth}\vspace{0pt}
\centering\ecgtablefont
\textbf{Historical strategies}\par\smallskip
\setlength{\tabcolsep}{3pt}\renewcommand{\arraystretch}{1.15}
\begin{tabularx}{\linewidth}{@{}l*{3}{>{\centering\arraybackslash}X}@{}}
\toprule\rowcolor{ecgtablegray}
Round & Tree & Uniform & Chain \\
\midrule
R0 & 30.15 & 30.15 & 30.15 \\
R1 & \underline{43.22}\ecgup & 39.67\ecgup & \textbf{44.20}\ecgup \\
R2 & \textbf{46.23}\ecgup & 43.25\ecgup & \underline{44.75}\ecgup \\
\rowcolor{ecgtableblue}
\textbf{R3} & \textbf{46.61}\ecgup & 41.45\ecgdown & \underline{43.70}\ecgdown \\
\midrule
$\Delta_{0\to3}$ & \ecgdelta{+16.46}\ecgup & \ecgdelta{+11.30}\ecgup & \ecgdelta{+13.55}\ecgup \\
\bottomrule
\end{tabularx}
\end{minipage}
\end{table}

\subsection{Results Analysis}
\label{sec:discussion}
\paragraph{Fine-grained feedback improves training utility.}
With a shared parent and 10,000 added questions, S7 reaches \textbf{46.23} with full feedback, versus 45.56 with fixed selection and 45.60 with category-only feedback (Tab.~\ref{tab:feedback-granularity}).

\paragraph{Error diagnosis proposes candidates; comparison selects the update.}
The retained path in Tab.~\ref{tab:round-branches} changes from direct reinforcement to confusion-focused supervision. This supports treating feedback as a source of candidate interventions, rather than a deterministic prescription for the next data mixture. \textbf{The training utility of a proposal is assessed by comparing candidates from the same parent}.

\paragraph{Useful training evidence changes with the learner.}
The selected data move from AV-block measurements to regional STEMI contrasts and then rhythm contrasts. These changes concern both \textbf{what evidence is targeted and how it is presented}. Historical data are replayed, while fresh feedback redirects the next additions; the reported path therefore reflects repeated intervention selection rather than repetition of a fixed recipe.

\paragraph{Local correction has diminishing external returns.}
In the category-level summary, feedback error falls from \textbf{39.63\% to 21.33\%} across R0--R3, but successive external S7 gains shrink to \textbf{13.07, 3.01, and 0.38} (Tabs.~\ref{tab:current-feedback-categories} and~\ref{tab:current-rounds}). ECG-QA, CPSC2018, and CODE-15\% regress in R3 despite fewer feedback errors. In this run, continued correction on the feedback set is therefore insufficient to characterize external progress. \textbf{Transfer value and regressions must be examined alongside error frequency}.

\section{Conclusion}
\label{sec:conclusion}
ECG-RSI combines \textbf{verified bidirectional synthesis} with empirical comparison of training interventions. The retained recipe shifts from direct reinforcement to confusion-focused contrasts as the learner changes. This supports \textbf{separating weakness diagnosis from intervention selection} within the tested setting: an error profile proposes directions, while candidate comparison determines which update to retain.

Across three rounds, S7 rises from \textbf{30.15 to 46.61}, and MedXpert ECG accuracy from 51.57\% to 57.24\%. Operator and skill updates improve synthesis efficiency and downstream performance. However, continued feedback-error reductions yield diminishing S7 gains, so \textbf{transfer and regression checks} remain necessary for interpreting progress beyond the feedback set.

\paragraph{Declaration of AI Assistance.}
AI tools assisted with \textbf{language polishing}, formatting, and figure preparation. 

\paragraph{Reproducibility Statement.}
The main text reports benchmark trajectories, feedback diagnostics, controlled ablations, blinded clinical review, and model-based question-quality scores. Supplementary materials provide the construction and training protocols, category-level records, prompts, and \textbf{complete case responses}. The present release is limited to the manuscript; the implementation and training-data files are not released.

\FloatBarrier
\bibliographystyle{plainnat}
\setcitestyle{numbers}
\bibliography{ref}

\begin{thebibliography}{65}
\providecommand{\natexlab}[1]{#1}
\providecommand{\url}[1]{\texttt{#1}}
\expandafter\ifx\csname urlstyle\endcsname\relax
  \providecommand{\doi}[1]{doi: #1}\else
  \providecommand{\doi}{doi: \begingroup \urlstyle{rm}\Url}\fi

\bibitem[Alcaraz and Strodthoff(2023)]{alcaraz2023sssd}
Juan Miguel~Lopez Alcaraz and Nils Strodthoff.
\newblock {Diffusion-based Conditional {ECG} Generation with Structured State
  Space Models}.
\newblock \emph{CoRR}, abs/2301.08227, 2023.

\bibitem[Bondar et~al.(2025)Bondar, Semenov, Babenko, and
  Holovniak]{bondar2025flowecg}
Vitalii Bondar, Serhii Semenov, Vira Babenko, and Dmytro Holovniak.
\newblock {FlowECG: Using Flow Matching to Create a More Efficient {ECG} Signal
  Generator}.
\newblock \emph{CoRR}, abs/2509.10491, 2025.

\bibitem[Chen et~al.(2022)Chen, Liao, Wei, Ying, Chen, and Wu]{chen2022megan}
Jintai Chen, Kuanlun Liao, Kun Wei, Haochao Ying, Danny~Z. Chen, and Jian Wu.
\newblock {{ME-GAN:} Learning Panoptic Electrocardio Representations for
  Multi-view {ECG} Synthesis Conditioned on Heart Diseases}.
\newblock \emph{CoRR}, abs/2207.10670, 2022.

\bibitem[Chen et~al.(2024{\natexlab{a}})Chen, Ouyang, Gao, Chen, Chen, Wang,
  Zhang, Cai, Ji, Yu, Wan, and Wang]{DBLP:journals/corr/abs-2406-19280}
Junying Chen, Ruyi Ouyang, Anningzhe Gao, Shunian Chen, Guiming~Hardy Chen,
  Xidong Wang, Ruifei Zhang, Zhenyang Cai, Ke~Ji, Guangjun Yu, Xiang Wan, and
  Benyou Wang.
\newblock {HuatuoGPT-Vision, Towards Injecting Medical Visual Knowledge into
  Multimodal LLMs at Scale}.
\newblock \emph{CoRR}, abs/2406.19280, 2024{\natexlab{a}}.

\bibitem[Chen et~al.(2026)Chen, Xu, Wang, Liu, Yao, and He]{supplied2026ref50}
Xiaoyin Chen, Canwen Xu, Yite Wang, Boyi Liu, Zhewei Yao, and Yuxiong He.
\newblock {Learning to Self-Evolve}.
\newblock \emph{CoRR}, abs/2603.18620, 2026.

\bibitem[Chen et~al.(2024{\natexlab{b}})Chen, Zhang, Fang, Geng, Guo, Chen, Li,
  Zhang, Chen, Zhu, Li, Liu, Pan, Zhang, and
  Chen]{DBLP:journals/corr/abs-2402-05391}
Zhuo Chen, Yichi Zhang, Yin Fang, Yuxia Geng, Lingbing Guo, Xiang Chen, Qian
  Li, Wen Zhang, Jiaoyan Chen, Yushan Zhu, Jiaqi Li, Xiaoze Liu, Jeff~Z. Pan,
  Ningyu Zhang, and Huajun Chen.
\newblock {Knowledge Graphs Meet Multi-Modal Learning: {A} Comprehensive
  Survey}.
\newblock \emph{CoRR}, abs/2402.05391, 2024{\natexlab{b}}.

\bibitem[Chong et~al.(2026)Chong, Zhang, Fan, and Du]{supplied2026ref49}
Meiduo Chong, Shaolei Zhang, Ju~Fan, and Xiaoyong Du.
\newblock {EvoOntology: {A} Self-Evolving Ontology Layer for Data Agents}.
\newblock \emph{CoRR}, abs/2609.15779, 2026.

\bibitem[Chung et~al.(2023)Chung, Kim, Kwon, Jeon, Lee, and
  Choi]{chung2023texttoecg}
Hyunseung Chung, Jiho Kim, Joon{-}Myoung Kwon, Ki{-}Hyun Jeon, Min~Sung Lee,
  and Edward Choi.
\newblock {Text-to-ECG: 12-Lead Electrocardiogram Synthesis conditioned on
  Clinical Text Reports}.
\newblock \emph{CoRR}, abs/2303.09395, 2023.

\bibitem[Dang et~al.(2026)Dang, Ruan, Yang, Wang, Hu, Li, and
  Heng]{dang2026medzero}
Xilin Dang, Weilin Ruan, Xue Yang, Jinghao Wang, Xiaowei Hu, Jinpeng Li, and
  Pheng-Ann Heng.
\newblock {MedZERO: Self-Evolving Agents for Open-Ended Medical Reasoning
  Through Controlled Knowledge Accumulation}.
\newblock \emph{arXiv preprint arXiv:2610.08327}, 2026.
\newblock URL \url{https://arxiv.org/abs/2610.08327}.

\bibitem[Fu et~al.(2026)Fu, Kulanthaivelu, and Yamada]{fu2026sift}
Xinghong Fu, Aravinth Kulanthaivelu, and Yutaro Yamada.
\newblock {Self Improvement via Fast Tree-search}.
\newblock \emph{CoRR}, abs/2609.19526, 2026.

\bibitem[{Google Research} and {Google DeepMind}(2025)]{sellergren2025medgemma}
{Google Research} and {Google DeepMind}.
\newblock {MedGemma Technical Report}.
\newblock \emph{CoRR}, abs/2507.05201, 2025.

\bibitem[Han et~al.(2025)Han, Duan, Cen, Yao, Song, Mhaskar, Leong, Rosenberg,
  Liu, and Zhao]{supplied2026ref38}
William Han, Chaojing Duan, Zhepeng Cen, Yihang Yao, Xiaoyu Song, Atharva
  Mhaskar, Dylan Leong, Michael~A. Rosenberg, Emerson Liu, and Ding Zhao.
\newblock {Signal, Image, or Symbolic: Exploring the Best Input Representation
  for Electrocardiogram-Language Models Through a Unified Framework}.
\newblock \emph{CoRR}, abs/2505.18847, 2025.

\bibitem[Han et~al.(2026)Han, Chen, Duan, Song, Yao, Yang, Rosenberg, Liu, and
  Zhao]{han2026elf}
William Han, Tony Chen, Chaojing Duan, Xiaoyu Song, Yihang Yao, Yuzhe Yang,
  Michael~A. Rosenberg, Emerson Liu, and Ding Zhao.
\newblock {Encoder-Free ECG-Language Models}.
\newblock \emph{CoRR}, abs/2601.18798, 2026.

\bibitem[He et~al.(2025)He, Huang, Li, Huang, and Yang]{he2025visplay}
Yicheng He, Chengsong Huang, Zongxia Li, Jiaxin Huang, and Yonghui Yang.
\newblock {VisPlay: Self-Evolving Vision-Language Models from Images}.
\newblock \emph{CoRR}, abs/2511.15661, 2025.

\bibitem[Hu et~al.(2026)Hu, Deng, Xing, Mu, Zou, Yang, Chen, Wang, Hao, Zhang,
  Chen, Wang, Zhang, Zeng, and Chen]{DBLP:journals/corr/abs-2608-27287}
Jinxin Hu, Hao Deng, Haibo Xing, Lingyu Mu, Muyu Zou, Weiqin Yang, Sirui Chen,
  Bohao Wang, Zhezheng Hao, Hao Zhang, Zulong Chen, Shizhun Wang, Yu~Zhang,
  Xiaoyi Zeng, and Jiawei Chen.
\newblock {Astar: Learning to Propose Evolution Directions for Self-Evolving
  Industrial {AI} Systems}.
\newblock \emph{CoRR}, abs/2608.27287, 2026.

\bibitem[Huang et~al.(2025)Huang, Sahoo, Kumari, Kumar, Cai, Cao, Kang, Xia,
  Chatterjee, Hausman, Jay, Rosenthal, Srinivasan, Hasan, Fedorov, and
  Vesal]{huang2025midtecg}
Zhuoyi Huang, Nutan Sahoo, Anamika Kumari, Girish Kumar, Kexuan Cai, Shixing
  Cao, Yue Kang, Tian Xia, Somya Chatterjee, Nick Hausman, Aidan Jay, Eric~S.
  Rosenthal, Soundar Srinivasan, Sadid Hasan, Alex Fedorov, and Sulaiman Vesal.
\newblock {High-Fidelity Synthetic {ECG} Generation via Mel-Spectrogram
  Informed Diffusion Training}.
\newblock \emph{CoRR}, abs/2510.05492, 2025.

\bibitem[Jang et~al.(2025)Jang, Jo, Kang, Son, Lee, myoung Kwon, and
  Lee]{supplied2026ref41}
Jong-Hwan Jang, Yong-Yeon Jo, Sora Kang, Jeong~Min Son, Hak~Seung Lee, Joon
  myoung Kwon, and Min~Sung Lee.
\newblock {A novel XAI framework for explainable AI-ECG using generative
  counterfactual XAI (GCX)}.
\newblock \emph{Scientific Reports}, 15:\penalty0 23608, 2025.
\newblock \doi{10.1038/s41598-025-08080-5}.
\newblock URL \url{https://doi.org/10.1038/s41598-025-08080-5}.

\bibitem[Jin et~al.(2026)Jin, Wang, Wu, Fang, Lan, Wang, Zhang, Liu, Zhang, Wu,
  Li, and Hong]{DBLP:journals/corr/abs-2602-04279}
Jiarui Jin, Haoyu Wang, Xingliang Wu, Xiaocheng Fang, Xiang Lan, Zihan Wang,
  Deyun Zhang, Bo~Liu, Yingying Zhang, Xian Wu, Hongyan Li, and Shenda Hong.
\newblock {{ECG-R1:} Protocol-Guided and Modality-Agnostic {MLLM} for Reliable
  {ECG} Interpretation}.
\newblock \emph{CoRR}, abs/2602.04279, 2026.

\bibitem[Kirchhof et~al.(2026)Kirchhof, Gualdoni, Szot, Gatmiry, Lotfi,
  Kazerouni, Attia, Chowdhury, and Toshev]{kirchhof2026rltldr}
Michael Kirchhof, Eleonora Gualdoni, Andrew Szot, Khashayar Gatmiry, Aryo
  Lotfi, Abbas Kazerouni, Omar Attia, Sanjoy Chowdhury, and Alexander Toshev.
\newblock {RLTL;DR: Self-improvement by Internalizing Self-generated Feedback}.
\newblock \emph{CoRR}, abs/2609.37633, 2026.

\bibitem[Lai et~al.(2025{\natexlab{a}})Lai, Chen, Zhang, Wang, Geng, Li, and
  Hong]{lai2025diffusets}
Yongfan Lai, Jiabo Chen, Deyun Zhang, Yue Wang, Shijia Geng, Hongyan Li, and
  Shenda Hong.
\newblock {DiffuSETS: 12-lead {ECG} Generation Conditioned on Clinical Text
  Reports and Patient-Specific Information}.
\newblock \emph{CoRR}, abs/2501.05932, 2025{\natexlab{a}}.

\bibitem[Lai et~al.(2025{\natexlab{b}})Lai, Liu, Guan, Zhao, Li, and
  Hong]{supplied2026ref33}
Yongfan Lai, Bo~Liu, Xinyan Guan, Qinghao Zhao, Hongyan Li, and Shenda Hong.
\newblock {ECGTwin: Personalized {ECG} Generation Using Controllable Diffusion
  Model}.
\newblock \emph{CoRR}, abs/2508.02720, 2025{\natexlab{b}}.

\bibitem[Lan et~al.(2025)Lan, Wu, He, Zhao, Hong, and Feng]{lan2025gem}
Xiang Lan, Feng Wu, Kai He, Qinghao Zhao, Shenda Hong, and Mengling Feng.
\newblock {{GEM:} Empowering {MLLM} for Grounded {ECG} Understanding with Time
  Series and Images}.
\newblock \emph{CoRR}, abs/2503.06073, 2025.

\bibitem[{LASA Team} et~al.(2025){LASA Team}, Xu, Chan, Li, Aljunied, Yuan,
  Wang, Xiao, Chen, Liu, Li, Sun, Shen, Wang, Tan, Zhao, Xu, Zhang, and
  Rong]{xu2025lingshu}
{LASA Team}, Weiwen Xu, Hou~Pong Chan, Long Li, Mahani Aljunied, Ruifeng Yuan,
  Jianyu Wang, Chenghao Xiao, Guizhen Chen, Chaoqun Liu, Zhaodonghui Li,
  Yu~Sun, Junao Shen, Chaojun Wang, Jie Tan, Deli Zhao, Tingyang Xu, Hao Zhang,
  and Yu~Rong.
\newblock {Lingshu: {A} Generalist Foundation Model for Unified Multimodal
  Medical Understanding and Reasoning}.
\newblock \emph{CoRR}, abs/2506.07044, 2025.

\bibitem[Lewis et~al.(2026)Lewis, Chiu, Verrier, Raghuram, Marvel,
  Abbaspourazad, Mishra, Sapiro, Miller, and
  Futoma]{lewis2026msecgfmuniversalelectrocardiogramfoundation}
Robert~A. Lewis, I-Min Chiu, Kyle Verrier, Karthik~Jayaraman Raghuram,
  Francoise Marvel, Salar Abbaspourazad, Anshuman Mishra, Guillermo Sapiro,
  Andrew~C. Miller, and Joseph Futoma.
\newblock {MS-ECG-FM: Towards a More Universal Electrocardiogram Foundation
  Model for Health Monitoring using Multi-source Contrastive Learning}, 2026.
\newblock URL \url{https://arxiv.org/abs/2610.07662}.

\bibitem[Li et~al.(2025)Li, Li, Mao, Liu, Sun, and Huang]{supplied2026ref34}
Haitao Li, Ziyu Li, Yiheng Mao, Ziyi Liu, Zhoujian Sun, and Zhengxing Huang.
\newblock {anyECG-chat: {A} Generalist {ECG-MLLM} for Flexible {ECG} Input and
  Multi-Task Understanding}.
\newblock \emph{CoRR}, abs/2506.00942, 2025.

\bibitem[Liang et~al.(2026)Liang, Zhu, and Zhu]{liang2026ecgmirage}
Jinning Liang, Mingcheng Zhu, and Tingting Zhu.
\newblock {{ECG} Mirage: Revealing and Mitigating the Underutilisation of ECGs
  in Vision-Language Models for Clinical Prediction}.
\newblock \emph{CoRR}, abs/2609.21755, 2026.

\bibitem[Lin et~al.(2026)Lin, Lin, Yuan, Wu, Cao, Liang, Cao, Zhu, Zhang, and
  Wu]{lin2026mmvistareason}
Juekai Lin, Honglin Lin, Yuqian Yuan, Xiaolong Wu, Jie Cao, Liang Liang, Yunqi
  Cao, Yun Zhu, Wenqiao Zhang, and Lijun Wu.
\newblock {MMVistaReason}: Toward open-data and post-training recipes for
  multimodal reasoning.
\newblock \emph{CoRR}, abs/2610.01352, 2026.
\newblock \doi{10.48550/ARXIV.2610.01352}.
\newblock URL \url{https://arxiv.org/abs/2610.01352}.

\bibitem[Lin et~al.(2025)Lin, Ma, Dong, Sun, Cong, Wang, Luo, and
  Wang]{supplied2026ref40}
Yuxin Lin, Jing Ma, Suyu Dong, Chaoyu Sun, Wanting Cong, Kuanquan Wang,
  Gongning Luo, and Wei Wang.
\newblock {TransDiffECG: Semantically controllable {ECG} synthesis via
  transformer-based diffusion modeling}.
\newblock \emph{J. Biomed. Informatics}, 172:\penalty0 104948, 2025.

\bibitem[Liu et~al.(2026{\natexlab{a}})Liu, Zhuang, Weng, and
  Ye]{liu2026recurse}
Kaiyuan Liu, Ziyuan Zhuang, Rongxiang Weng, and Jieping Ye.
\newblock {RecurSE: Bounded Recursive Self-Evaluation for {LLM} Rubric Judges}.
\newblock \emph{CoRR}, abs/2608.24231, 2026{\natexlab{a}}.

\bibitem[Liu et~al.(2024)Liu, Bai, Yue, and Zhang]{liu2024pulse}
Ruoqi Liu, Yuelin Bai, Xiang Yue, and Ping Zhang.
\newblock {Teach Multimodal LLMs to Comprehend Electrocardiographic Images}.
\newblock \emph{CoRR}, abs/2410.19008, 2024.

\bibitem[Liu et~al.(2026{\natexlab{b}})Liu, Agarwal, Maheswaran, Cemri, Li,
  Mang, Naren, Boneh, Cheng, Pan, Du, Keutzer, Dimakis, Sen, Zaharia, and
  Stoica]{liu2026evox}
Shu Liu, Shubham Agarwal, Monishwaran Maheswaran, Mert Cemri, Zhifei Li,
  Qiuyang Mang, Ashwin Naren, Ethan Boneh, Audrey Cheng, Melissa~Z. Pan,
  Alexander Du, Kurt Keutzer, Alexandros~G. Dimakis, Koushik Sen, Matei
  Zaharia, and Ion Stoica.
\newblock {EvoX: Meta-Evolution for Automated Discovery}.
\newblock \emph{CoRR}, abs/2602.23413, 2026{\natexlab{b}}.

\bibitem[Liu et~al.(2026{\natexlab{c}})Liu, Qi, Du, and
  He]{liu2026learnableinfo}
Wei Liu, Siya Qi, Yali Du, and Yulan He.
\newblock {Self-Play Only Evolves When Self-Synthetic Pipeline Ensures
  Learnable Information Gain}.
\newblock \emph{CoRR}, abs/2603.02218, 2026{\natexlab{c}}.

\bibitem[Madaan et~al.(2023)Madaan, Tandon, Gupta, Hallinan, Gao, Wiegreffe,
  Alon, Dziri, Prabhumoye, Yang, Gupta, Majumder, Hermann, Welleck,
  Yazdanbakhsh, and Clark]{madaan2023self}
Aman Madaan, Niket Tandon, Prakhar Gupta, Skyler Hallinan, Luyu Gao, Sarah
  Wiegreffe, Uri Alon, Nouha Dziri, Shrimai Prabhumoye, Yiming Yang, Shashank
  Gupta, Bodhisattwa~Prasad Majumder, Katherine Hermann, Sean Welleck, Amir
  Yazdanbakhsh, and Peter Clark.
\newblock {Self-Refine: Iterative Refinement with Self-Feedback}.
\newblock In \emph{NeurIPS}, 2023.

\bibitem[Meng et~al.(2026)Meng, Du, Chen, Zhao, Lu, Hu, and
  Shieh]{DBLP:journals/corr/abs-2607-25886}
Fanqing Meng, Lingxiao Du, Qiguang Chen, Ziqi Zhao, Haocheng Lu, Mengkang Hu,
  and Michael~Qizhe Shieh.
\newblock {RSIBench-Data: Benchmarking Data-Centric Research for Recursive
  Self-Improvement}.
\newblock \emph{CoRR}, abs/2607.25886, 2026.

\bibitem[{NeoHorse Team} et~al.(2026){NeoHorse Team}, Cao, Dai, Guo, Han, Hu,
  Jiang, Kuang, Li, Li, Pei, Tian, Wang, Wang, Wang, Wu, Xu, Zhang, Zhou,
  Cheng, Fan, He, Jiao, Li, Li, Liu, Liu, Liu, Pan, Ren, Song, Wang, Yu, Zhang,
  Zhang, Zheng, and Zong]{neohorse2026}
{NeoHorse Team}, Guoliang Cao, Guohao Dai, Tianyu Guo, Kai Han, Hailin Hu,
  Zihan Jiang, Xiang Kuang, Boxun Li, Yulong Li, Zehua Pei, Yuchuan Tian,
  Jiamin Wang, Yu~Wang, Yunhe Wang, Yihong Wu, Haiyang Xu, Shuo Zhang, Hang
  Zhou, Siyang Cheng, Jiayu Fan, Wei He, Qingrui Jiao, Hongguang Li, Zhiyuan
  Li, Runke Liu, Xi~Liu, Xinchen Liu, Sinno~Jialin Pan, Yi~Ren, Liuyang Song,
  Chenyu Wang, Bei Yu, Quanlu Zhang, Xiangyu Zhang, Mengyu Zheng, and Yingjie
  Zong.
\newblock {NeoHorse-1: Towards Recursive Self-Improvement via Agentic
  Post-Training with Routing Harness}.
\newblock \emph{CoRR}, abs/2609.08183, 2026.

\bibitem[Oh et~al.(2026)Oh, Chung, Lee, Kim, Yoon, Lee, Lee, Yeo, and
  Choi]{oh2026ecgbenchmark}
Jungwoo Oh, Hyunseung Chung, Junhee Lee, Min{-}Gyu Kim, Hangyul Yoon, Ki~Seong
  Lee, Youngchae Lee, Muhan Yeo, and Edward Choi.
\newblock {ECG-Reasoning-Benchmark: {A} Benchmark for Evaluating Clinical
  Reasoning Capabilities in {ECG} Interpretation}.
\newblock \emph{CoRR}, abs/2603.14326, 2026.

\bibitem[Pham et~al.(2025)Pham, Tang, Saeed, and Ma]{supplied2026ref35}
Hung~Manh Pham, Jialu Tang, Aaqib Saeed, and Dong Ma.
\newblock {{Q-HEART:} {ECG} Question Answering via Knowledge-Informed
  Multimodal LLMs}.
\newblock In \emph{{ECAI}}, volume 413 of \emph{Frontiers in Artificial
  Intelligence and Applications}, pages 4545--4552. {IOS} Press, 2025.

\bibitem[Selivanov et~al.(2026)Selivanov, Jungmann, Kehrer, Laugwitz, Martens,
  and Rueckert]{supplied2026ref31}
Alexander Selivanov, Friederike Jungmann, Jan Kehrer, Karl{-}Ludwig Laugwitz,
  Eimo Martens, and Daniel Rueckert.
\newblock {{ECG-LLM:} Foundation Model for ECG-Based Cardiac Reasoning}.
\newblock \emph{CoRR}, abs/2607.16323, 2026.

\bibitem[Shinn et~al.(2023)Shinn, Cassano, Gopinath, Narasimhan, and
  Yao]{DBLP:conf/nips/ShinnCGNY23}
Noah Shinn, Federico Cassano, Ashwin Gopinath, Karthik Narasimhan, and Shunyu
  Yao.
\newblock {Reflexion: language agents with verbal reinforcement learning}.
\newblock In \emph{NeurIPS}, 2023.

\bibitem[Singh et~al.(2026)Singh, Bakhtin, Shao, Synnaeve, Kulikov, Fergus,
  Arora, Keutzer, Weston, Mahajan, and
  Goyal]{singh2026agentplasticitymeasuringselfimprovement}
Harman Singh, Anton Bakhtin, Rulin Shao, Gabriel Synnaeve, Ilia Kulikov, Rob
  Fergus, Sanjeev Arora, Kurt Keutzer, Jason Weston, Anuj Mahajan, and Anirudh
  Goyal.
\newblock {Agent Plasticity: Measuring Self-Improvement Through Experience},
  2026.
\newblock URL \url{https://arxiv.org/abs/2610.08902}.

\bibitem[Skorik and Avetisyan(2024)]{skorik2024sssdnle}
Sergey Skorik and Aram Avetisyan.
\newblock {SSSD-ECG-nle: New Label Embeddings with Structured State-Space
  Models for {ECG} generation}.
\newblock \emph{CoRR}, abs/2407.11108, 2024.

\bibitem[Soubh et~al.(2026)Soubh, Rasenack, Haarmann, Wiedmann, Zabel, Schmidt,
  Suliman, and Bergau]{supplied2026ref42}
Nibras Soubh, Eva Rasenack, Helge Haarmann, Felix Wiedmann, Markus Zabel,
  Constanze Schmidt, Rayan Suliman, and Leonard Bergau.
\newblock {Performance of Vision-Enabled Large Language Models in Image-Based
  Electrocardiogram Interpretation: Exploratory Evaluation}.
\newblock \emph{Journal of Medical Internet Research}, 28:\penalty0 e86692,
  2026.
\newblock \doi{10.2196/86692}.
\newblock URL \url{https://doi.org/10.2196/86692}.

\bibitem[Srikanth et~al.(2026)Srikanth, Zhao, Xu, Wu, and
  Jiang]{srikanth2026aide2}
Dhruv Srikanth, Bingchen Zhao, Dixing Xu, Yuxiang Wu, and Zhengyao Jiang.
\newblock {Recursive self-improvement of {AI} research agents}.
\newblock \emph{CoRR}, abs/2609.26457, 2026.

\bibitem[Wang et~al.(2025)Wang, Kang, and Han]{supplied2026ref32}
Xu~Wang, Jiaju Kang, and Puyu Han.
\newblock {ECG-Expert-QA: {A} Benchmark for Evaluating Medical Large Language
  Models in Heart Disease Diagnosis}.
\newblock \emph{CoRR}, abs/2502.17475, 2025.

\bibitem[Wang et~al.(2023)Wang, Kordi, Mishra, Liu, Smith, Khashabi, and
  Hajishirzi]{wang2023selfinstruct}
Yizhong Wang, Yeganeh Kordi, Swaroop Mishra, Alisa Liu, Noah~A. Smith, Daniel
  Khashabi, and Hannaneh Hajishirzi.
\newblock {Self-Instruct: Aligning Language Models with Self-Generated
  Instructions}.
\newblock In \emph{{ACL} {(1)}}, pages 13484--13508. Association for
  Computational Linguistics, 2023.

\bibitem[Wang et~al.(2026)Wang, Yan, Bi, Yan, Tresp, and
  Ma]{wang2026metaskillevolve}
Zefeng Wang, Minxi Yan, Jinhe Bi, Sikuan Yan, Volker Tresp, and Yunpu Ma.
\newblock {MetaSkill-Evolve: Recursive Self-Improvement of {LLM} Agents via
  Two-Timescale Meta-Skill Evolution}.
\newblock \emph{CoRR}, abs/2607.05297, 2026.

\bibitem[Wu et~al.(2026)Wu, Xie, Sun, Tong, and
  Zhao]{DBLP:journals/corr/abs-2608-23397}
Ruoyu Wu, Shenfu Xie, Yinqian Sun, Haibo Tong, and Feifei Zhao.
\newblock {MediSkill-Evo: Process-Constrained Self-Evolution for
  Evidence-Grounded Clinical Interaction}.
\newblock \emph{CoRR}, abs/2608.23397, 2026.

\bibitem[Xia et~al.(2026)Xia, Han, Wang, Chen, Zhuang, Lee, Huang, Yu, CuiZhu,
  Ming, Yao, Gokturk, Pfister, and Lee]{supplied2026ref48}
Peng Xia, Rujun Han, Zifeng Wang, Yanfei Chen, Yufan Zhuang, Yoonho Lee,
  Chengsong Huang, Han Yu, Zhongying CuiZhu, Yifei Ming, Huaxiu Yao, Burak
  Gokturk, Tomas Pfister, and Chen{-}Yu Lee.
\newblock {{RRSI:} Regularized Recursive Self-Improvement of Agent Harnesses}.
\newblock \emph{CoRR}, abs/2609.24972, 2026.

\bibitem[Xu et~al.(2024)Xu, Sun, Zheng, Geng, Zhao, Feng, Tao, Lin, and
  Jiang]{xu2024wizardlm}
Can Xu, Qingfeng Sun, Kai Zheng, Xiubo Geng, Pu~Zhao, Jiazhan Feng, Chongyang
  Tao, Qingwei Lin, and Daxin Jiang.
\newblock {WizardLM: Empowering Large Pre-Trained Language Models to Follow
  Complex Instructions}.
\newblock In \emph{{ICLR}}. OpenReview.net, 2024.

\bibitem[Xu et~al.(2026)Xu, Haresamudram, Liu, Langer, Pradeepkumar, Mao,
  Ferns, Verma, Sun, Schmiedmayer, Liu, McDuff, Fox, and
  Rehg]{xu2026ecgreasoneval}
Maxwell~A. Xu, Harish Haresamudram, Catherine~W. Liu, Patrick Langer,
  Jathurshan Pradeepkumar, Wanting Mao, Sunita~J. Ferns, Aradhana Verma, Jimeng
  Sun, Paul Schmiedmayer, Xin Liu, Daniel McDuff, Emily~B. Fox, and James~M.
  Rehg.
\newblock {How Well Do Multimodal Models Reason on {ECG} Signals?}
\newblock \emph{CoRR}, abs/2603.00312, 2026.

\bibitem[Xue et~al.(2026)Xue, Zhong, Nan, Li, Yu, Ding, Gao, Zhan, Zhang,
  Cheng, Yin, Wu, and Yang]{xue2026sciencebuddy}
Shuhan Xue, Jianyuan Zhong, Ziyuan Nan, Wenbin Li, Zhaochen Yu, Jinchao Ding,
  Qiang Gao, Pengyu Zhan, Yuntong Zhang, Tian Cheng, Zhenfei Yin, Yingcheng Wu,
  and Ling Yang.
\newblock {ScienceBuddy: Recursive-in-Recursive Self-Improvement for
  Interactive Scientific Agents}.
\newblock \emph{CoRR}, abs/2609.17523, 2026.

\bibitem[Yan et~al.(2026)Yan, Gao, Wang, Mao, and Wang]{supplied2026ref45}
Siyu Yan, Yizhen Gao, Yilin Wang, Dongxing Mao, and Alex~Jinpeng Wang.
\newblock {DataEvolver: Self-Evolving Multi-Agent Data Construction for
  Text-Rich Image Generation}.
\newblock \emph{CoRR}, abs/2606.31537, 2026.

\bibitem[Yang et~al.(2026{\natexlab{a}})Yang, Wu, Lin, Xu, Jiang, Xiong, and
  Guo]{DBLP:journals/corr/abs-2608-29966}
Cehao Yang, Xiaojun Wu, Xueyuan Lin, Chengjin Xu, Xuhui Jiang, Hui Xiong, and
  Jian Guo.
\newblock {DataFoundry: Evolving Data Preparators via Recursive
  Self-Improvement}.
\newblock \emph{CoRR}, abs/2608.29966, 2026{\natexlab{a}}.

\bibitem[Yang et~al.(2026{\natexlab{b}})Yang, Ma, Huang, Hu, Wang, and
  Chu]{yang2026coevolve}
Shidong Yang, Ziyu Ma, Tongwen Huang, Yiming Hu, Yong Wang, and Xiangxiang Chu.
\newblock {CoEvolve: Training {LLM} Agents via Agent-Data Mutual Evolution}.
\newblock \emph{CoRR}, abs/2604.15840, 2026{\natexlab{b}}.

\bibitem[Yang et~al.(2026{\natexlab{c}})Yang, Li, Qin, Wang, Wang, Li, Yang,
  Li, Thomason, Ma, and Zhao]{yang2026syncodatasynthesiscotraining}
Wei Yang, Shawn Li, Yuehan Qin, Yawei Wang, Mingxi Wang, Shixuan Li, Tiankai
  Yang, Jiate Li, Jesse Thomason, Xuezhe Ma, and Yue Zhao.
\newblock {SynCo: Data Synthesis Co-Training for Self-Evolving LLMs via
  Multi-Agent Reinforcement Learning}, 2026{\natexlab{c}}.
\newblock URL \url{https://arxiv.org/abs/2610.11345}.

\bibitem[Yin et~al.(2026{\natexlab{a}})Yin, Liu, and Qi]{yin2026stopping}
Enoch Yin, Bin Liu, and Zhengling Qi.
\newblock {When Is Enough Enough in Self-Evolving LLM Systems?}
\newblock \emph{arXiv preprint arXiv:2610.04756}, 2026{\natexlab{a}}.
\newblock URL \url{https://arxiv.org/abs/2610.04756}.

\bibitem[Yin et~al.(2026{\natexlab{b}})Yin, Zhao, Asadi, Liu, Lu, Mei, Cui,
  Simon, Shi, and Firooz]{yin2026recursiveopd}
Shangjian Yin, Zehao Zhao, Kavosh Asadi, Rui Liu, Yuchen Lu, Shike Mei, Hang
  Cui, Luke Simon, Zhouxing Shi, and Hamed Firooz.
\newblock {Recursive Self-Improvement via On-Policy Distillation for
  Reasoning}.
\newblock \emph{CoRR}, abs/2609.30652, 2026{\natexlab{b}}.

\bibitem[Zelikman et~al.(2022)Zelikman, Wu, Mu, and Goodman]{zelikman2022star}
Eric Zelikman, Yuhuai Wu, Jesse Mu, and Noah~D. Goodman.
\newblock {STaR: Bootstrapping Reasoning With Reasoning}.
\newblock In \emph{NeurIPS}, 2022.

\bibitem[Zelikman et~al.(2024)Zelikman, Lorch, Mackey, and
  Kalai]{zelikman2024stop}
Eric Zelikman, Eliana Lorch, Lester Mackey, and Adam~Tauman Kalai.
\newblock {Self-Taught Optimizer (STOP): Recursively Self-Improving Code
  Generation}.
\newblock In \emph{Conference on Language Modeling (COLM)}, 2024.
\newblock URL \url{https://arxiv.org/abs/2310.02304}.

\bibitem[Zeng et~al.(2026)Zeng, Tan, Cen, Wen, Hu, Bian, Chen, and
  Chen]{supplied2026ref44}
Min Zeng, Guanxin Tan, Libin Cen, Yafei Wen, Rui Hu, Liuyang Bian, Xiaolong
  Chen, and Xiaoxin Chen.
\newblock {{VISA:} Agentic Self-Evolving Data Synthesis for Multimodal
  Instruction Following}.
\newblock \emph{CoRR}, abs/2608.26013, 2026.

\bibitem[Zhang et~al.(2025{\natexlab{a}})Zhang, Lan, Geng, Zhao, Fan, Feng, and
  Hong]{supplied2026ref37}
Deyun Zhang, Xiang Lan, Shijia Geng, Qinghao Zhao, Sumei Fan, Mengling Feng,
  and Shenda Hong.
\newblock {{MEETI:} {A} Multimodal {ECG} Dataset from {MIMIC-IV-ECG} with
  Signals, Images, Features and Interpretations}.
\newblock \emph{CoRR}, abs/2507.15255, 2025{\natexlab{a}}.

\bibitem[Zhang et~al.(2025{\natexlab{b}})Zhang, Hu, Lu, Lange, and
  Clune]{zhang2025dgm}
Jenny Zhang, Shengran Hu, Cong Lu, Robert~T. Lange, and Jeff Clune.
\newblock {Darwin Godel Machine: Open-Ended Evolution of Self-Improving
  Agents}.
\newblock \emph{CoRR}, abs/2505.22954, 2025{\natexlab{b}}.

\bibitem[Zhang et~al.(2026)Zhang, Zhao, Yang, Foerster, Clune, Jiang, Devlin,
  and Shavrina]{supplied2026ref43}
Jenny Zhang, Bingchen Zhao, Wannan Yang, Jakob~N. Foerster, Jeff Clune, Minqi
  Jiang, Sam Devlin, and Tatiana Shavrina.
\newblock {Hyperagents}.
\newblock \emph{CoRR}, abs/2603.19461, 2026.

\bibitem[Zhao et~al.(2025)Zhao, Wu, Yue, Wu, Xu, Yue, Lin, Wang, Wu, Zheng, and
  Huang]{zhao2025absolutezero}
Andrew Zhao, Yiran Wu, Yang Yue, Tong Wu, Quentin Xu, Yang Yue, Matthieu Lin,
  Shenzhi Wang, Qingyun Wu, Zilong Zheng, and Gao Huang.
\newblock {Absolute Zero: Reinforced Self-play Reasoning with Zero Data}.
\newblock \emph{CoRR}, abs/2505.03335, 2025.

\bibitem[Zheng et~al.(2026)Zheng, Wu, Zhang, He, Zhang, Coleman, Wei, Bai, Liu,
  Liu, Wang, Zhuan, Kang, Xiang, Huang, Cheng, and Guo]{zheng2026dreamrsi}
Tong Zheng, Xidong Wu, Zheng Zhang, Zhankui He, Chaoyi Zhang, Benjamin Coleman,
  Ruoqiao Wei, Di~Bai, Haolin Liu, Rui Liu, Xue Wang, Yue Zhuan, Wang{-}Cheng
  Kang, Renkai Xiang, Heng Huang, Xinwu Cheng, and Yunsong Guo.
\newblock {Dream-RSI: Recursive Self-Improvement through Evolving Worlds}.
\newblock \emph{CoRR}, abs/2609.14858, 2026.

\end{thebibliography}

\Needspace{18\baselineskip}
\beginappendix
\definecolor{appendixpurple}{HTML}{70528F}
\definecolor{appendixblue}{HTML}{235E91}
\newcommand{\appkey}[1]{\textcolor{appendixblue}{\textbf{#1}}}
\newcommand{\apptheme}[1]{\textcolor{appendixpurple}{\textit{#1}}}
\newcommand{\appfield}[1]{\textcolor{appendixpurple}{\textbf{#1}}}
\tcbset{lines before break=4}

\Needspace{4\baselineskip}
\section*{Appendix Contents and Reading Guide}
The entries below link to all supplementary sections and their starting pages.
\par\vspace{3pt}
\begingroup
\fontsize{9.5}{11}\selectfont
\setlength{\tabcolsep}{5pt}
\setlength{\aboverulesep}{3.5pt}
\setlength{\belowrulesep}{3.5pt}
\renewcommand{\arraystretch}{0.94}
\newcommand{\guideentry}[3]{\ref{#1} & \textbf{#2}\newline #3 & \pageref{#1}\\\addlinespace[0pt]}
\newcommand{\guideentrylast}[3]{\ref{#1} & \textbf{#2}\newline #3 & \pageref{#1}\\*}
\begin{xltabular}{\linewidth}{@{}p{0.085\linewidth} X r@{}}
\toprule
Section & Contents & Page\\
\midrule
\endfirsthead
\toprule
Section & Contents (continued) & Page\\
\midrule
\endhead
\endfoot
\endlastfoot
\guideentry{app:method-details}{Supplementary Method Details}{Data recipes, construction repair, recipe search, and implementation settings.}
\guideentry{app:student-fit-details}{Student-Conditioned Data Recipes}{Learner-conditioned recipe design and controlled difficulty probes.}
\guideentry{app:repair-details}{Construction Feedback and Versioned Repair}{Construction failure diagnosis and version-controlled generator updates.}
\guideentry{app:waveform-construction-details}{Waveform Construction and Validation Details}{Waveform editing operators and signal-level validation rules.}
\guideentry{sec:method:branch-search}{Bounded Recipe Search within the Learning Loop}{Candidate recipe branching, evaluation, and model promotion.}
\guideentry{app:training-config}{Training Configuration}{Optimizer, batching, distributed training, and frozen components.}
\guideentry{app:parameter-extraction}{Parameter Extraction and Selection Details}{ECG measurement extraction, reliability, and parameter selection.}
\midrule
\guideentry{sec:current-data-appendix}{Supplementary Experiments and Data}{Feedback counts, branch records, and waveform measurements.}
\guideentry{app:guide-difficulty-level-feedback-counts}{Difficulty-Level Feedback Counts}{Round-wise feedback counts for each difficulty level.}
\guideentry{app:guide-category-level-feedback-counts}{Category-Level Feedback Counts}{Round-wise feedback counts across ECG task categories.}
\guideentry{app:guide-supplementary-branch-evaluation-records}{Supplementary Branch-Evaluation Records}{Candidate branch evaluation and selection records.}
\guideentry{app:ecg-multiseed}{Multi-Seed Results on ECG Tasks}{Multi-seed evaluation results and variability across ECG tasks.}
\guideentry{app:evaluation-splits}{Evaluation Splits}{Eight enabled splits, sample counts, and metrics (Tab.~\ref{tab:evaluation-splits}).}
\midrule
\guideentry{app:taxonomy-23}{ECG Task Taxonomy}{Medical grouping of the 23 ECG task categories.}
\midrule
\guideentry{app:prompt-templates}{Stage-wise Prompt Templates}{Instructions for the construction, review, repair, and learning stages.}
\guideentry{app:guide-constrained-waveform-construction}{Constrained waveform construction}{Waveform editing constraints and structured construction outputs.}
\guideentry{app:guide-evidence-grounded-question-and-context-generation}{Evidence-grounded question and context generation}{Question and context generation grounded in waveform evidence.}
\guideentry{app:guide-teacher-answer-and-evidence-explanation}{Teacher answer and evidence explanation}{Teacher answers and supporting evidence explanations.}
\guideentry{app:guide-independent-quality-and-acceptance-review}{Independent quality and acceptance review}{Independent checks for question quality and acceptance.}
\guideentry{app:guide-construction-failure-and-versioned-repair}{Construction failure and versioned repair}{Failure analysis and versioned generator repair.}
\guideentry{app:guide-learner-feedback-diagnosis-and-recipe-proposal}{Learner-feedback diagnosis and recipe proposal}{Learner error diagnosis and next-round recipe proposals.}
\guideentry{app:guide-candidate-comparison-and-model-promotion}{Candidate comparison and model promotion}{Candidate comparison criteria and model promotion decisions.}
\midrule
\guideentry{app:waveform-case-studies}{Prompts and Case Studies}{Complete worked examples with prompts and model responses.}
\guideentry{app:case-pr}{Case 1: PR prolongation}{PR-interval editing and the corresponding question-answer example.}
\guideentrylast{app:case-rate}{Case 2: Heart-rate slowing}{Heart-rate editing and the corresponding question-answer example.}
\bottomrule
\end{xltabular}
\addtocounter{table}{-1} 
\endgroup
\par\medskip

\par\bigskip

\section{Supplementary Method Details}
\label{app:method-details}
\subsection{Student-Conditioned Data Recipes}
\label{app:student-fit-details}
Equation~\eqref{eq:recipe} defines the joint recipe over category, difficulty, and supervision representation.

Here $g$ is a task/capability category, $d$ is learner-relative difficulty, and $s$ is a supervision representation. Each recipe also registers \appkey{retrieval/edit constraints, replay, and budgets}. The operator and prompt library $\phi$ determines which requests can be constructed. Selecting among fixed templates changes the recipe; editing a template or operator changes the library version. These interventions \textbf{must be evaluated separately}.

\paragraph{Difficulty estimation.}
With the current learner and decoding settings fixed, answer each training-side candidate independently $K_i$ times and estimate difficulty from its fraction of correct responses:
\begin{equation}
\widehat p_{it}=\frac{1}{K_i}\sum_{r=1}^{K_i}\mathbf1[\widehat y_{ir}=y_i].
\label{eq:learner-difficulty}
\end{equation}
Group candidates into \appkey{reliably solved, mixed-success, and unresolved} questions. 

\paragraph{Supervision representations.}
A verified question can have \appkey{(i)} the \appkey{original teacher rationale}, \appkey{(ii)} an \appkey{evidence-expanded rationale} making required intermediate observations explicit, and \appkey{(iii)} an \appkey{error-targeted rationale} adding checks motivated by the student's observed failure pattern. ``Chain of thought (CoT)'' denotes teacher-generated explanations used for supervision. Store token count under the student's tokenizer, required evidence-step coverage, and the checks added. Paired expansions \textbf{preserve the image, question, options, answer, and clinical facts}. 

\paragraph{Difficulty belongs to the current learner.}
\textbf{Register the rollout budget}, concept-level scoring, position-counterbalancing policy, minimum valid-output rate, and binning thresholds before branch comparison. Approximate bins are reliably solved, mixed success, and unresolved; $0/K_i$ denotes unresolved performance under this probe.  \textbf{Parsing/truncation remain separate flags}: report end-to-end correctness and correctness conditional on \mbox{valid output}. 

\paragraph{ECG-specific supervision repair.}
For a \appkey{QRS-width error}, a candidate expanded target \textbf{identifies QRS onset/offset}, \textbf{converts grid distance to milliseconds}, and \textbf{checks the threshold}, using independently verified fiducials. If measurements are already correct but the label is wrong, add the relevant \appkey{cross-lead differential evidence}. Error type and waveform evidence guide interventions, supported by contrastive probes, scored outputs, and signals. 

Fig.~\ref{fig:feedback-selection} maps feedback to data-selection and generation requests for branch validation.

\subsection{Construction Feedback and Versioned Repair}
\label{app:repair-details}
Acceptance also checks leakage and diversity.

\paragraph{Repair by failure type.}
Acceptance tracks target attainment, non-target drift, continuity, lead consistency, morphology, context, answer uniqueness, and cost. \appkey{Target errors or boundary artifacts} guide operator revisions; \appkey{missing morphology} guides applicability/retrieval revisions; \appkey{ambiguity} guides question/context revisions; \appkey{unsupported explanations} guide prompt revisions.

\paragraph{Reusable rules and version records.}
Each skill specifies applicability, sources, operator order, parameter ranges, failure handling, and question rules. Updates record feedback, changes, proposer, both versions, validation results, and next-round scope. \textbf{Compare the original and revised versions} on the same source signals, requests, random seeds, and attempt budgets, and adopt revisions after \textbf{independent operator and question regression checks}.

\subsection{Waveform Construction and Validation Details}
\label{app:waveform-construction-details}

\paragraph{Bidirectional synthesis and reverse perturbation.}
 The \appkey{forward route} measures and renders a real training waveform, then constructs a supported question, context, and answer. The \appkey{reverse route} selects an eligible training waveform for a target contrast, applies constrained operators, and remeasures it before rendering and question construction. 

A \appkey{PR-interval contrast} requires \textbf{measurable P and QRS onsets}. The edit must reach the target range while \textbf{preserving heart rate and QRS morphology} within fixed tolerances. Complex disease patterns require verified morphology and clinical evidence. Both routes share question, context, and explanation checks. Bidirectional synthesis creates examples; dual feedback repairs construction and chooses subsequent requests.

 Sources: \appkey{(i)} retrieved training ECGs, \appkey{(ii)} reused/reweighted accepted questions, \appkey{(iii)} \emph{training-distractor} targets, \appkey{(iv)} controlled edits, and \appkey{(v)} audited training-side morphology donors. Check \textbf{feasibility, applicability, and evidence}.

For a source $S$ and permitted donor set, construct $S'=T_\delta(S)$ and re-extract $u'=f(S')$. Store \appkey{target error}
$|[f_k(S')-f_k(S)]-\Delta_k^{\mathrm{request}}|$ and \appkey{protected drift}
$|f_j(S')-f_j(S)|$ separately. Check continuity, unedited regions, lead alignment, and class morphology. Sources/donors are \textbf{patient-disjoint} from feedback, selection, and evaluation.

\FloatBarrier

\subsection{Bounded Recipe Search within the Learning Loop}
\label{sec:method:branch-search}\label{sec:tree_evolution}
\appkey{Construction feedback} validates examples and guides generator repair. \appkey{Learner feedback} supplies the demand profile and delayed utility for recipe selection. Each round follows five registered steps:

\FloatBarrier

\begin{enumerate}[nosep,leftmargin=1.4em,label=\textcolor{appendixblue}{\textbf{\arabic*.}}]
\item \textbf{Score the accepted learner} on $V_{\mathrm{fb}}$ and probe training candidates to update category, difficulty, and error profiles.
\item \textbf{Propose a bounded recipe set} $\mathcal A_t$ from these profiles and prior outcomes; save the proposer and exact recipe.
\item \textbf{Construct and validate} each recipe's training data under fixed gates. From the \emph{same parent checkpoint}, run matched small-budget SFT branches with fixed replay and optimizer settings.
\item \textbf{Compare candidates} under the registered selection protocol, including forgetting and output validity. 
\item \textbf{Promote the accepted candidate} and its recipe/policy versions, then remeasure for the \mbox{next round}. Full-budget retraining of a pilot winner restarts from the registered parent and requires validation before promotion.
\end{enumerate}
For branch $a$, the simplest candidate utility is
\begin{equation}
\Delta U_t(a)=S_{\mathrm{sel}}\!\left(\operatorname{SFT}(\theta_t,D_t^a\cup R_t;b)\right)-S_{\mathrm{sel}}(\theta_t),
\label{eq:branch-utility}
\end{equation}
Here $\Delta U_t(a)$ is the selection-score improvement of branch $a$ over its parent, not a weighted sum over branches. $S_{\mathrm{sel}}$ evaluates a learner on $V_{\mathrm{sel}}$ and is distinct from the external seven-benchmark score S7. Here $b$ is the shared pilot budget; $S_{\mathrm{sel}}$ is the preregistered capability-macro score over fixed eligible tasks. 

Each round trains candidate branches from the same parent, compares them on a fixed development set, and retains the \textbf{best qualifying branch}. The \mbox{next round} starts from that model, while unselected branches are archived. The search tree records candidates; the retained path records the selected \mbox{model updates}.

Fig.~\ref{fig:tree-evolution} illustrates candidate exploration and checkpoint inheritance, with a colored example lineage. Section~\ref{sec:method:stage1} defines the three recipe families; Tab.~\ref{tab:round-branches} reports measured compositions, scores, and selections.

\FloatBarrier

\par\Needspace*{90pt}
\subsection{Training Configuration}
\label{app:training-config}
Each round uses \textbf{three training epochs} and a maximum sequence length of 28,672 tokens. Packing, padding-free batching, and sequence parallelism are enabled. The micro-batch size is per GPU.

\par\smallskip\noindent\begin{minipage}{\linewidth}
\captionsetup{type=table}\centering
\caption{Training hyperparameters and distributed settings.}
\label{tab:training-config}
\ecgtablefont
\setlength{\tabcolsep}{4pt}
\renewcommand{\arraystretch}{1.1}
\begin{tabularx}{\linewidth}{@{}>{\raggedright\arraybackslash}X>{\centering\arraybackslash}p{.17\linewidth}>{\raggedright\arraybackslash}X>{\centering\arraybackslash}p{.17\linewidth}@{}}
\toprule
\rowcolor{ecgtablegray}
\multicolumn{1}{c}{Parameter} & Setting & \multicolumn{1}{c}{Parameter} & Setting \\
\midrule
\rowcolor{ecgtablegray}\multicolumn{2}{c}{\apptheme{Sequence and topology}} & \multicolumn{2}{c}{\apptheme{Optimization}} \\
Epochs / maximum tokens & 3 / 28,672 & Optimizer & \appkey{Distributed Adam} \\
Micro-batch / global batch & 1 / 32 & Peak / minimum LR & $10^{-5}$ / $10^{-7}$ \\
Gradient accumulation & 1 & Schedule / warmup & Cosine / 5\% \\
GPUs (nodes $\times$ GPUs) & $128\ (16\times8)$ & Weight decay / grad. clip & 0.1 (constant) / 1.0 \\
DP / TP / PP / CP & 32 / 4 / 1 / 1 & Adam $\beta_1/\beta_2/\epsilon$ & $0.9/0.95/10^{-8}$ \\
EP / ETP & 8 / 1 & Compute / master precision & bfloat16 / float32 \\
Packing / padding-free & Enabled / enabled & CPU offload fraction & 1.0 \\
\midrule
\rowcolor{ecgtablegray}\multicolumn{4}{c}{\apptheme{Modules and experts}} \\
Attention backend & FlashAttention & Grouped GEMM & Enabled \\
Recompute modules & \texttt{core\_attn}, \texttt{moe}, \texttt{mlp} & Permutation fusion & Disabled \\
Shared-expert overlap & Enabled & Auxiliary loss / router dtype & $10^{-6}$ / float32 \\
\bottomrule
\end{tabularx}
\end{minipage}\par\smallskip
DP, TP, PP, CP, EP, and ETP denote data, tensor, pipeline, context, expert, and expert-tensor \appkey{parallelism}.

Distributed Adam uses cosine learning-rate (LR) decay with 5\% warmup. Computation uses bfloat16; master parameters, gradients, and Adam states use float32. Optimizer CPU offload is enabled with fraction 1.0.

\textbf{The language model is trainable}; \textbf{the vision encoder and multimodal aligner are frozen}. Selected modules are recomputed, with FlashAttention as the attention backend. GEMM denotes general matrix multiplication.
\par\medskip

Open-data curation and capacity-aware post-training for broader multimodal reasoning are also studied in MMVistaReason~\citep{lin2026mmvistareason}.

\subsection{Parameter Extraction and Selection Details}
\label{app:parameter-extraction}

A delineator supplies candidate measurements
\begin{equation}
u=f(S),\quad u\in\mathbb R^M .\label{eq:map}
\end{equation}
At 25\,mm/s, a 1\,mm horizontal square is 40\,ms and a 5\,mm square is 200\,ms; at 10\,mm/mV, 1\,mm vertically is 0.1\,mV (a 5\,mm square is 0.5\,mV). PR is measured from \appkey{P onset to QRS onset}, and QT from \appkey{QRS onset to T offset}. 

The parameter selector uses \appkey{reliability and diagnostic utility},
\begin{equation}
r_k=\frac{1/\hat\sigma_k}{\sum_j1/\hat\sigma_j},\label{eq:reliability}
\end{equation}
\begin{equation}
s_{ik}=a_{ik}r_kd_k,\label{eq:selection_score}
\end{equation}
\begin{equation}
\mathcal C_i=\{k:s_{ik}>\tau_{\mathrm{param}}\}.\label{eq:selection}
\end{equation}
Here $a_{ik}$ is definability, $\hat\sigma_k$ an extraction-error scale, and $d_k$ a probe-based utility.

\noindent\appfield{Scale specification.} Comparing inverse error scales across features requires a \textbf{common dimensionless normalization}, with reference scales for measurements in bpm, ms, and mV.

\section{Supplementary Experiments and Data}
\label{sec:current-data-appendix}

This appendix summarizes round-wise feedback statistics, branch-evaluation results, and waveform data. Overall feedback statistics use a \appkey{fixed set of 3,000 questions}, with 1,189, 983, 803, and 640 errors at R0--R3, respectively.

\par\Needspace{4\baselineskip}
\subsection{Difficulty-Level Feedback Counts}
\label{app:guide-difficulty-level-feedback-counts}

\appfield{Residual difficulty.} L3 error falls from 58.20\% to 38.17\%, the largest reduction across difficulty levels. It also retains the highest final error rate; L1 falls to 12.97\%. Counts and group sizes distinguish changes in error rate from differences in group size.

For this table and Tab.~\ref{tab:current-feedback-categories}, blue marks R3. Bold and underlined values identify the best and second-best rates; $\downarrow$ marks a decrease from the previous round.

\begin{table}[!htbp]
\centering
\caption{Difficulty-level errors: count (\%). Group sizes appear in the header.}
\label{tab:current-feedback-difficulty-counts}
\ecgtablefont
\setlength{\tabcolsep}{3pt}
\renewcommand{\arraystretch}{1.15}
\begin{tabularx}{\linewidth}{@{}l*{3}{>{\centering\arraybackslash}X}@{}}
\toprule
\rowcolor{ecgtablegray}
Round & \makecell{L1\\$N=1{,}218$} & \makecell{L2\\$N=1{,}093$} & \makecell{L3\\$N=689$} \\
\midrule
R0 & 318 (26.11) & 421 (38.52) & 401 (58.20) \\
R1 & 247 (20.28\ecgerrdown) & 362 (33.12\ecgerrdown) & 357 (51.81\ecgerrdown) \\
R2 & 196 (\underline{16.09}\ecgerrdown) & 309 (\underline{28.27}\ecgerrdown) & 306 (\underline{44.41}\ecgerrdown) \\
\rowcolor{ecgtableblue}
R3 & 158 (\textbf{12.97}\ecgerrdown) & 267 (\textbf{24.43}\ecgerrdown) & 263 (\textbf{38.17}\ecgerrdown) \\
\bottomrule
\end{tabularx}
\end{table}

\subsection{Category-Level Feedback Counts}
\label{app:guide-category-level-feedback-counts}
Tab.~\ref{tab:current-feedback-breakdown} and~\ref{tab:current-fixed-feedback} in Section~\ref{sec:feedback-diagnostics} summarize the feedback trajectories. The recorded category-level breakdown follows.

\par\medskip
{

\ecgtablefont
\setlength{\tabcolsep}{3.5pt}
\renewcommand{\arraystretch}{1.15}
\begin{xltabular}{\linewidth}{@{}>{\raggedright\arraybackslash}X*{9}{r}@{}}\caption{Recorded error counts and rates (\%) for 23 ECG categories across R0--R3. $N$: group size. Each question is assigned to one primary category; category counts sum to 3,000. Error rates are error counts divided by the corresponding $N$.}
\label{tab:current-feedback-categories}\\
\toprule
 & & \multicolumn{4}{c}{Error count} & \multicolumn{4}{c}{Error rate (\%)} \\
\cmidrule(lr){3-6}\cmidrule(lr){7-10}
\rowcolor{ecgtablegray}
Group & $N$ & R0 & R1 & R2 & R3 & R0 & R1 & R2 & R3 \\
\midrule
\endfirsthead
\toprule
 & & \multicolumn{4}{c}{Error count} & \multicolumn{4}{c}{Error rate (\%)} \\
\cmidrule(lr){3-6}\cmidrule(lr){7-10}
\rowcolor{ecgtablegray}
Group & $N$ & R0 & R1 & R2 & R3 & R0 & R1 & R2 & R3 \\
\midrule
\endhead

Normal Sinus Rhythm & 168 & 42 & 33 & 25 & 19 & 25.00 & 19.64\ecgerrdown & \underline{14.88}\ecgerrdown & \cellcolor{ecgtableblue}\textbf{11.31}\ecgerrdown \\
Atrial Fibrillation & 154 & 61 & 50 & 41 & 33 & 39.61 & 32.47\ecgerrdown & \underline{26.62}\ecgerrdown & \cellcolor{ecgtableblue}\textbf{21.43}\ecgerrdown \\
Atrial Flutter & 147 & 66 & 56 & 47 & 39 & 44.90 & 38.10\ecgerrdown & \underline{31.97}\ecgerrdown & \cellcolor{ecgtableblue}\textbf{26.53}\ecgerrdown \\
Sinus Tachycardia & 132 & 37 & 29 & 22 & 17 & 28.03 & 21.97\ecgerrdown & \underline{16.67}\ecgerrdown & \cellcolor{ecgtableblue}\textbf{12.88}\ecgerrdown \\
Sinus Bradycardia & 128 & 34 & 26 & 20 & 15 & 26.56 & 20.31\ecgerrdown & \underline{15.63}\ecgerrdown & \cellcolor{ecgtableblue}\textbf{11.72}\ecgerrdown \\
Premature Ventricular Contraction & 143 & 58 & 48 & 39 & 31 & 40.56 & 33.57\ecgerrdown & \underline{27.27}\ecgerrdown & \cellcolor{ecgtableblue}\textbf{21.68}\ecgerrdown \\
Premature Atrial Contraction & 136 & 53 & 44 & 36 & 29 & 38.97 & 32.35\ecgerrdown & \underline{26.47}\ecgerrdown & \cellcolor{ecgtableblue}\textbf{21.32}\ecgerrdown \\
First-Degree AV Block & 121 & 46 & 38 & 31 & 24 & 38.02 & 31.40\ecgerrdown & \underline{25.62}\ecgerrdown & \cellcolor{ecgtableblue}\textbf{19.83}\ecgerrdown \\
Mobitz I AV Block & 109 & 46 & 38 & 31 & 24 & 42.20 & 34.86\ecgerrdown & \underline{28.44}\ecgerrdown & \cellcolor{ecgtableblue}\textbf{22.02}\ecgerrdown \\
Mobitz II AV Block & 105 & 51 & 43 & 36 & 29 & 48.57 & 40.95\ecgerrdown & \underline{34.29}\ecgerrdown & \cellcolor{ecgtableblue}\textbf{27.62}\ecgerrdown \\
Complete AV Block & 118 & 52 & 43 & 35 & 28 & 44.07 & 36.44\ecgerrdown & \underline{29.66}\ecgerrdown & \cellcolor{ecgtableblue}\textbf{23.73}\ecgerrdown \\
Left Bundle Branch Block & 139 & 51 & 42 & 34 & 27 & 36.69 & 30.22\ecgerrdown & \underline{24.46}\ecgerrdown & \cellcolor{ecgtableblue}\textbf{19.42}\ecgerrdown \\
Right Bundle Branch Block & 141 & 50 & 41 & 33 & 26 & 35.46 & 29.08\ecgerrdown & \underline{23.40}\ecgerrdown & \cellcolor{ecgtableblue}\textbf{18.44}\ecgerrdown \\
Anterior STEMI & 149 & 66 & 56 & 47 & 39 & 44.30 & 37.58\ecgerrdown & \underline{31.54}\ecgerrdown & \cellcolor{ecgtableblue}\textbf{26.17}\ecgerrdown \\
Inferior STEMI & 145 & 63 & 53 & 44 & 36 & 43.45 & 36.55\ecgerrdown & \underline{30.34}\ecgerrdown & \cellcolor{ecgtableblue}\textbf{24.83}\ecgerrdown \\
Lateral STEMI & 116 & 54 & 45 & 37 & 30 & 46.55 & 38.79\ecgerrdown & \underline{31.90}\ecgerrdown & \cellcolor{ecgtableblue}\textbf{25.86}\ecgerrdown \\
Posterior STEMI & 108 & 53 & 44 & 37 & 30 & 49.07 & 40.74\ecgerrdown & \underline{34.26}\ecgerrdown & \cellcolor{ecgtableblue}\textbf{27.78}\ecgerrdown \\
Left Ventricular Hypertrophy & 152 & 57 & 47 & 38 & 30 & 37.50 & 30.92\ecgerrdown & \underline{25.00}\ecgerrdown & \cellcolor{ecgtableblue}\textbf{19.74}\ecgerrdown \\
Right Ventricular Hypertrophy & 124 & 51 & 42 & 34 & 27 & 41.13 & 33.87\ecgerrdown & \underline{27.42}\ecgerrdown & \cellcolor{ecgtableblue}\textbf{21.77}\ecgerrdown \\
Long QT Syndrome & 134 & 55 & 46 & 38 & 30 & 41.04 & 34.33\ecgerrdown & \underline{28.36}\ecgerrdown & \cellcolor{ecgtableblue}\textbf{22.39}\ecgerrdown \\
Hyperkalemia & 117 & 50 & 42 & 35 & 28 & 42.74 & 35.90\ecgerrdown & \underline{29.91}\ecgerrdown & \cellcolor{ecgtableblue}\textbf{23.93}\ecgerrdown \\
Wolff-Parkinson-White Syndrome & 113 & 47 & 39 & 32 & 25 & 41.59 & 34.51\ecgerrdown & \underline{28.32}\ecgerrdown & \cellcolor{ecgtableblue}\textbf{22.12}\ecgerrdown \\
Brugada Syndrome & 101 & 46 & 38 & 31 & 24 & 45.54 & 37.62\ecgerrdown & \underline{30.69}\ecgerrdown & \cellcolor{ecgtableblue}\textbf{23.76}\ecgerrdown \\
\midrule
Overall reference & 3,000 & 1,189 & 983 & 803 & 640 & 39.63 & 32.77\ecgerrdown & \underline{26.77}\ecgerrdown & \cellcolor{ecgtableblue}\textbf{21.33}\ecgerrdown \\
\bottomrule
\end{xltabular}
\par\smallskip

}
\par\medskip

\par\Needspace{4\baselineskip}
\subsection{Supplementary Branch-Evaluation Records}
\label{app:guide-supplementary-branch-evaluation-records}
The parent and three candidates are compared on the same R0$\to$R1 development protocol. The table reports overall accuracy, three macro scores, and selection score $S$; $\Delta S$ is relative to the parent.

\textbf{All candidates pass the score and decline gates}. \textbf{A is promoted} with the \appkey{highest selection score (42.66)}; B and C remain eligible. Worst-group changes are accuracy differences from the parent, in percentage points.

Tab.~\ref{tab:round-branches} gives all three rounds of candidate selection and the full A$\to$B$\to$B retained path.

\par\smallskip
\noindent\begin{minipage}{\linewidth}
\centering
\begin{minipage}{.7\linewidth}
\captionsetup{type=table}
\caption{R0$\to$R1 development scores and branch selection. Blue marks the promoted branch.}
\label{tab:current-branch-register}
\ecgtablefont
\setlength{\tabcolsep}{2pt}
\renewcommand{\arraystretch}{1.15}
\begin{tabularx}{\linewidth}{@{}>{\raggedright\arraybackslash}Xc>{\columncolor{ecgtableblue}}ccc@{}}
\toprule
\rowcolor{ecgtablegray}
\multicolumn{1}{@{}c}{Metric} & \multicolumn{1}{c}{Parent} & \multicolumn{1}{c}{A} & \multicolumn{1}{c}{B} & \multicolumn{1}{c@{}}{C} \\
\midrule
Overall (\%) & 30.15 & 43.22 & 41.09 & 38.78 \\
Five-group macro (\%) & 29.80 & 42.10 & 40.25 & 38.10 \\
Task-type macro (\%) & 30.05 & 42.85 & 40.80 & 38.45 \\
Difficulty macro (\%) & 29.60 & 41.90 & 39.95 & 37.70 \\
$S$ & 29.93 & \textbf{42.66} & 40.51 & 38.27 \\
$\Delta S$ & -- & \ecgdelta{+12.73} & \ecgdelta{+10.58} & \ecgdelta{+8.34} \\
\midrule
Worst-group change & -- & \ecgdelta{-1.20} & \ecgdelta{-1.80} & \ecgdelta{-2.10} \\
Score / decline gate & -- & \ecgpass\ Pass & \ecgpass\ Pass & \ecgpass\ Pass \\
\bottomrule
\end{tabularx}
\end{minipage}
\end{minipage}
\par\medskip

\par\medskip

\par\Needspace{4\baselineskip}
\subsection{Multi-Seed Results on ECG Tasks}
\label{app:ecg-multiseed}

Scores are reported for \appkey{three independent training runs}; SD is the sample standard deviation. The base row uses the same fixed initialization.

Mean ECG-task scores rise from 33.40 at R1 to 37.82 at R3, with SD between 0.22 and 0.29 across trained rounds.
\par\medskip

\begin{table}[!htbp]

\centering

\centering
\caption{ECG-task scores across three training seeds.}
\label{tab:ecg-multiseed}
\ecgtablefont
\setlength{\tabcolsep}{2pt}
\renewcommand{\arraystretch}{1.15}
\begin{tabularx}{\linewidth}{@{}l*{3}{>{\centering\arraybackslash}X}>{\centering\arraybackslash}p{0.30\linewidth}@{}}
\toprule
\rowcolor{ecgtablegray}
Model & Seed 1 & Seed 2 & Seed 3 & Mean $\pm$ SD \\
\midrule
Base & 30.15 & 30.15 & 30.15 & $30.15 \pm 0.00$ \\
ECG-RSI R1 & 33.42 & 33.18 & 33.61 & $33.40 \pm 0.22$ \\
ECG-RSI R2 & 35.27 & 35.04 & 35.48 & $35.26 \pm 0.22$ \\
\rowcolor{ecgtableblue}
\textbf{ECG-RSI R3} & 37.86 & 37.52 & 38.09 & $\boldsymbol{37.82 \pm 0.29}$ \\
\bottomrule
\end{tabularx}
\end{table}

\subsection{Evaluation Splits}
\label{app:evaluation-splits}

MMMU-ECG is the MMMU-inspired ECG multiple-choice task in ECGBench. CSN denotes Chapman--Shaoxing--Ningbo; G12EC denotes the Georgia 12-Lead ECG Challenge Database. ECG-QA (ECG-QA: A Comprehensive Question Answering Dataset Combined With Electrocardiogram) uses questions based on PTB-XL. CPSC2018 denotes the China Physiological Signal Challenge 2018, and CODE-15\% is the public subset of Clinical Outcomes in Digital Electrocardiography. MedXpert-ECG contains ECG questions selected from MedXpertQA-MM. The configurations \texttt{ptb-test-report} (500 items) and \texttt{arena} (50 items) are not enabled and are excluded from the total. LUDB is evaluated separately from these eight splits.

Figure labels MMMU, G12, ECGQA, PTB, CPSC, and CODE15 correspond to MMMU-ECG, G12EC, ECG-QA, PTB-XL, CPSC2018, and CODE-15\%, respectively.

\begin{table}[!htbp]
\centering
\caption{Enabled evaluation splits. The first seven contribute equally to S7; MedXpert-ECG is reported separately. Counts describe evaluation items.}
\label{tab:evaluation-splits}
\ecgtablefont
\setlength{\tabcolsep}{3pt}
\begin{tabularx}{\linewidth}{@{}lXrl@{}}
\toprule
Benchmark & Split & $N$ & Task / reported metric \\
\midrule
MMMU-ECG & \texttt{mmmu-ecg} & 200 & Single-choice accuracy \\
CSN & \texttt{csn-test-no-cot} & 1,611 & Single-choice accuracy \\
G12EC & \texttt{g12-test-no-cot} & 2,026 & Single-choice accuracy \\
ECG-QA & \texttt{ecgqa-test} & 1,317 & Answer-set exact-match accuracy \\
PTB-XL & \texttt{ptb-test} & 2,082 & Multi-label F1 \\
CPSC2018 & \texttt{cpsc-test} & 2,061 & Multi-label F1 \\
CODE-15\% & \texttt{code15-test} & 1,400 & Multi-label F1 \\
MedXpert-ECG & \texttt{medxpert-ecg} & 143 & Five-option accuracy (A--E) \\
\midrule
Total & & 10,840 & \\
\bottomrule
\end{tabularx}
\end{table}

\par\medskip
\Needspace{5\baselineskip}
\section{ECG Task Taxonomy}
\label{app:taxonomy-23}
\begin{figure}[!htbp]
\centering
\includegraphics[width=0.68\linewidth]{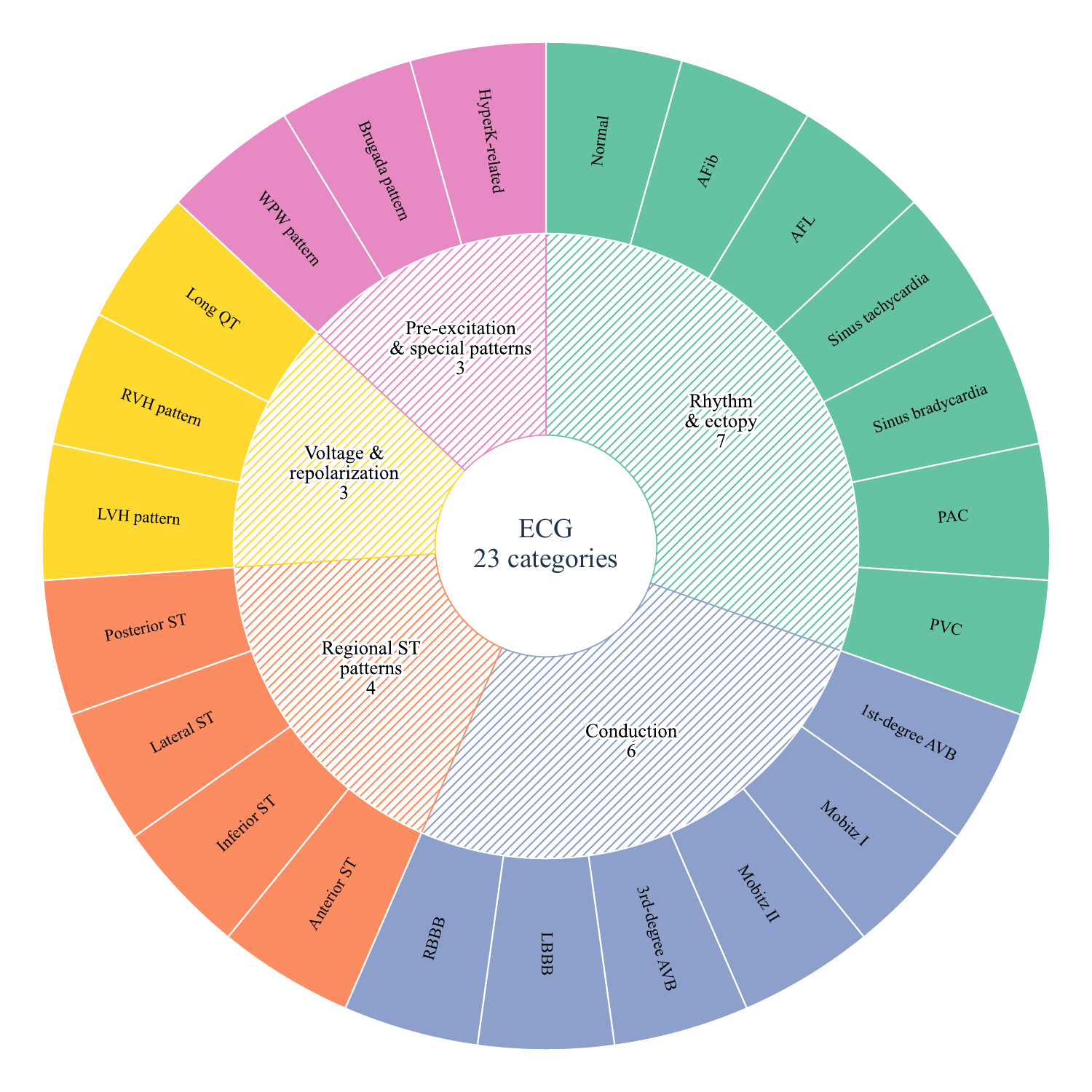}
\caption{Medical taxonomy organizing the 23 ECG task categories. Each outer sector denotes one category; inner sectors group related categories, with areas and counts indicating the number of categories in each group. The diagram summarizes task coverage and the grouping used for category-level feedback.}
\label{fig:taxonomy-23}

\end{figure}

Fig.~\ref{fig:taxonomy-23} groups the 23 ECG tasks by medical category; sector areas encode category counts.

\Needspace*{70pt}
\section{Stage-wise Prompt Templates}
\label{app:prompt-templates}
\definecolor{stagepromptpurple}{HTML}{70528F}
This appendix documents reusable prompt templates for the method interfaces. Bracketed fields are filled with request-specific inputs and tool results. Section~\ref{app:guide-teacher-answer-and-evidence-explanation} reproduces the teacher-answer instruction; the other templates are organized by function. The following cases retain their original questions and saved responses.

\AppendixNeedspace{110pt}
\subsection{Constrained waveform construction}
\label{app:guide-constrained-waveform-construction}
\begin{prompt}[breakable,enhanced,title={Prompt P1},fontupper=\small,colback=stagepromptpurple!4,colframe=stagepromptpurple,colbacktitle=stagepromptpurple!20,coltitle=stagepromptpurple,boxrule=0.5pt]

\appfield{System role.} You are an ECG waveform-construction agent. Convert a target feature request into an auditable edit plan.
\par\smallskip

\appkey{User inputs.} [request ID], [training-source waveform and sampling rate], [lead names and calibration], [measured landmarks and uncertainty], [target ranges], [protected features and tolerances], [allowed operators and donors], [retry budget].
\par\smallskip

\appfield{Instructions.} \textbf{Check source eligibility and landmark reliability}. For the \appkey{forward route}, extract supported features without changing the signal. For the \appkey{reverse route}, specify operator order, parameters, and edit mask. Invoke the supplied editing and measurement tools, then \textbf{compare observed targets and protected drift} with the fixed tolerances. Check boundary continuity, lead alignment, and required morphology. If tools or required landmarks are unavailable, return blocked; if verification fails, repair within budget \mbox{or reject}.
\par\smallskip

\appkey{Output fields.} Status; source identities; operator plan; requested and observed measurements with units; protected drift; validity checks; failure codes; edited-signal reference. \textbf{Report executed results separately from proposed operations}.
\par\smallskip

\end{prompt}

\AppendixNeedspace{110pt}
\subsection{Evidence-grounded question and context generation}
\label{app:guide-evidence-grounded-question-and-context-generation}
\begin{prompt}[breakable,enhanced,title={Prompt P2},fontupper=\small,colback=stagepromptpurple!4,colframe=stagepromptpurple,colbacktitle=stagepromptpurple!20,coltitle=stagepromptpurple,boxrule=0.5pt]

\appfield{System role.} You construct ECG multiple-choice questions from verified signal evidence. Keep internal labels and generator intent out of the student-visible question.
\par\smallskip

\appkey{User inputs.} [verified ECG image], [measurement evidence], [supported label level], [target category], [question type and difficulty], [option count], [context policy], [permitted source context].
\par\smallskip

\appfield{Instructions.} Write one question with exactly the requested number of options and \appkey{one evidence-supported answer}. Use plausible distractors that remain incorrect for this signal. Ask only at the supported level: measurement, ECG pattern, or clinically supported diagnosis. Default to image-only or traceable demographics. \textbf{Mark any synthetic teaching context explicitly} and check it against the achieved waveform. 
\par\smallskip

\appkey{Output fields.} Student-visible question, options, and allowed context; evaluator-only answer, evidence mapping, context provenance, category, and difficulty; ambiguity or insufficient-evidence flag.
\par\smallskip

\end{prompt}

\AppendixNeedspace{110pt}
\subsection{Teacher answer and evidence explanation}
\label{app:guide-teacher-answer-and-evidence-explanation}
\begin{prompt}[breakable,enhanced,title={Prompt P3},fontupper=\small,colback=stagepromptpurple!4,colframe=stagepromptpurple,colbacktitle=stagepromptpurple!20,coltitle=stagepromptpurple,boxrule=0.5pt]

Please reason step by step, and put your final answer within \texttt{\textbackslash boxed\{\}}.

\end{prompt}

\AppendixNeedspace{110pt}
\subsection{Independent quality and acceptance review}
\label{app:guide-independent-quality-and-acceptance-review}
\begin{prompt}[breakable,enhanced,title={Prompt P4},fontupper=\small,colback=stagepromptpurple!4,colframe=stagepromptpurple,colbacktitle=stagepromptpurple!20,coltitle=stagepromptpurple,boxrule=0.5pt]

\appfield{System role.} You audit a candidate ECG training example. 
\par\smallskip

\appkey{User inputs.} [source and edited evidence], [target/protected-feature checks], [question and options], [context provenance], [proposed answer and teacher response], [fixed rubric and aggregation rule].
\par\smallskip

\appfield{Instructions.} \textbf{Independently solve the question}, then audit target attainment, protected drift, morphology, context consistency, answer uniqueness, leakage, and explanation support. Return separate decisions for each gate and cite the relevant observation. Score only with the supplied rubric. The aggregate-score threshold is \appkey{80/100}, but \textbf{every mandatory gate must also pass}.
\par\smallskip

\appkey{Output fields.} Independent answer; gate-wise pass/fail decisions and evidence; rubric scores; aggregate if defined; accept/review/reject status; failure codes and repair target.
\par\smallskip

\end{prompt}

\AppendixNeedspace{110pt}
\subsection{Construction failure and versioned repair}
\label{app:guide-construction-failure-and-versioned-repair}
\begin{prompt}[breakable,enhanced,title={Prompt P5},fontupper=\small,colback=stagepromptpurple!4,colframe=stagepromptpurple,colbacktitle=stagepromptpurple!20,coltitle=stagepromptpurple,boxrule=0.5pt]

\appfield{System role.} You propose a minimal repair to a failed data-construction component while keeping the acceptance \mbox{criteria fixed}.
\par\smallskip

\appkey{User inputs.} [failed attempt and evidence], [failure codes], [current operator/prompt/skill version], [allowed repair scope], [fixed regression suite], [attempt budget].
\par\smallskip

\appfield{Instructions.} Map numerical or boundary failures to operator repair; missing morphology to applicability or donor retrieval; ambiguity/leakage to question or context repair; unsupported explanation to response-prompt repair. \textbf{Specify the exact change} and why it addresses the observed failure. \textbf{Compare old and new versions} on the same sources, requests, seeds, and budget. Quarantine \mbox{or reject} when the budget \mbox{is exhausted}.
\par\smallskip

\appkey{Output fields.} Triggering failure; component; exact revision; old/new version; proposer; validation results; acceptance rationale; next-round scope.
\par\smallskip

\end{prompt}

\AppendixNeedspace{110pt}
\subsection{Learner-feedback diagnosis and recipe proposal}
\label{app:guide-learner-feedback-diagnosis-and-recipe-proposal}
\begin{prompt}[breakable,enhanced,title={Prompt P6},fontupper=\small,colback=stagepromptpurple!4,colframe=stagepromptpurple,colbacktitle=stagepromptpurple!20,coltitle=stagepromptpurple,boxrule=0.5pt]

\appfield{System role.} You propose training-data interventions from development feedback. 
\par\smallskip

\appkey{User inputs.} [accepted parent and feedback version], [category/type/difficulty counts and errors], [confusions and regressions], [training-pool coverage], [construction yield/cost], [history], [token and attempt budgets].
\par\smallskip

\appfield{Instructions.} \textbf{Prioritize deficits with adequate support}; report sparse or uncertain groups explicitly. Propose distinct candidates for \appkey{direct weakness coverage}, \appkey{confusion contrasts}, and \appkey{difficulty or evidence-granularity progression}. For each, give normalized category, difficulty, and supervision-style allocations; requested feature ranges; protected attributes; allowed routes/operators; source filters; and cost limits. A/B/C retain the same previous cumulative training set unchanged and each add 10,000 first-use accepted questions, without an extra shared baseline set; \textbf{only the winning additions enter the next cumulative set}. 
\par\smallskip

\appkey{Output fields.} candidate recipes and hypotheses; normalized allocations and budgets; construction requests; \mbox{provenance restrictions}.
\par\smallskip

\end{prompt}

\AppendixNeedspace{235pt}
\subsection{Candidate comparison and model promotion}
\label{app:guide-candidate-comparison-and-model-promotion}
\begin{prompt}[breakable,enhanced,title={Prompt P7},fontupper=\small,colback=stagepromptpurple!4,colframe=stagepromptpurple,colbacktitle=stagepromptpurple!20,coltitle=stagepromptpurple,boxrule=0.5pt]

\appfield{System role.} You apply a preregistered development-only promotion rule to comparable candidate learners. 
\par\smallskip

\appkey{User inputs.} [parent and candidate checkpoint IDs], [new/replay manifests], [supervised-token budgets], [fixed selection-set results], [primary score], [regression tolerances], [tie/stopping rules].
\par\smallskip

\appfield{Instructions.} \textbf{Check that candidates share the required parent}, training budget, and evaluation protocol. Apply the supplied score, regression, and valid-output gates exactly; do not invent thresholds. \textbf{Promote only an eligible candidate}, otherwise retain the parent. Count only previously unused accepted items from the promoted branch toward cumulative additions, preserving replay provenance. \textbf{Archive other branches and log the decision}. 
\par\smallskip

\appkey{Output fields.} Comparability checks; gate results; chosen checkpoint or retain-parent decision; supporting scores; admitted item manifest; replay manifest; stopping status.
\par\smallskip

\end{prompt}

\FloatBarrier
\section{Prompts and Case Studies}
\label{app:waveform-case-studies}
\noindent\casekey{Blue bold text} marks reasoning cues and reported values; \casecaution{amber text} marks assumptions or grounding concerns.

\definecolor{casestudyblue}{HTML}{326DA8}

\begin{figure}[H]
\centering
\includegraphics[width=.96\linewidth]{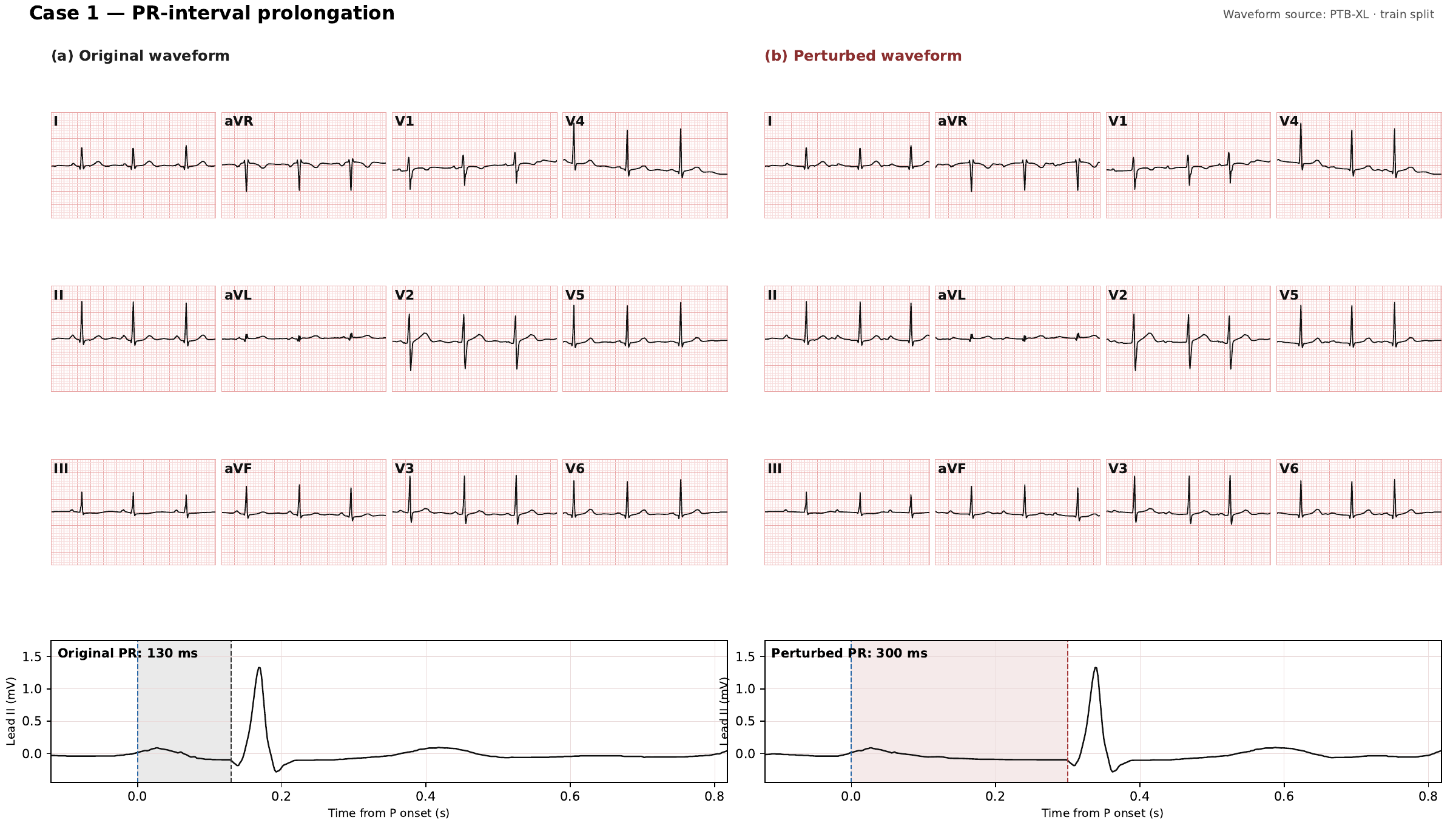}
\caption{Targeted PR prolongation, shown through original and edited ECGs with lead II details. The edit lengthens the PQ segment and compensates in the TP segment to preserve the RR interval. \casekey{The target PR is 300 ms} and the remeasured PR is 298 ms; these automatic measurements support the subsequent parameter-comparison question.}
\label{fig:case-pr}
\end{figure}
\subsection{Case 1: PR prolongation}
\label{app:case-pr}
\begin{prompt}[breakable,enhanced,colback=casestudyblue!3,colframe=casestudyblue!75,colbacktitle=casestudyblue,coltitle=white,boxrule=0.6pt,fontupper=\small,title={Case 1: PR prolongation},title after break={Case 1: PR prolongation (continued)}]
\raggedright\setlength{\parindent}{0pt}\setlength{\parskip}{2pt}
\appfield{Waveform construction}\par
Copy the P wave and QRS--T segment, lengthen the PQ segment, and shorten the TP segment to preserve the RR interval. \casekey{The target PR is 300 ms}; remeasurement yields \casehl{PR 298 ms} and \casehl{HR 74.6 bpm}.

\medskip\hrule\medskip\appfield{Question}\par
Question: Under the automatic closed-loop measurement convention of this data-construction pipeline, which parameter analysis and conclusion are correct for this 12-lead ECG?\par
Options:\par
A: \casehl{HR 74.6 bpm}, PR 130.0 ms, QRS 96.0 ms, QTc 393.7 ms -> Conclusion: no first-degree \mbox{atrioventricular block}\par
B: HR 48.0 bpm, \casehl{PR 298.0 ms}, QRS 96.0 ms, QTc 393.7 ms -> Conclusion: sinus bradycardia with first-degree \mbox{atrioventricular block}\par
C: \casehl{HR 74.6 bpm}, \casehl{PR 298.0 ms}, QRS 96.0 ms, QTc 393.7 ms -> Conclusion: first-degree \mbox{atrioventricular block}\par
D: \casehl{HR 74.6 bpm}, \casehl{PR 298.0 ms}, QRS 130.0 ms, QTc 393.7 ms -> Conclusion: first-degree \mbox{atrioventricular block} with intraventricular conduction delay\par
Please reason step by step, and put your final answer within \textbackslash{}boxed\{\}.\par

\medskip\hrule\medskip\appfield{Qwen3.8 response}\par
Let me analyze this ECG question.\par
The question is asking (in Chinese): "According to the automatic closed-loop measurement criteria of this data construction pipeline, looking at this 12-lead ECG, which of the following parameter analyses and conclusions is correct?"\par
Options:\par
A: \casehl{HR 74.6 bpm}, PR 130.0 ms, QRS 96.0 ms, QTc 393.7 ms \ensuremath{\rightarrow} Conclusion: No first-degree AV \mbox{block observed}\par
B: HR 48.0 bpm, \casehl{PR 298.0 ms}, QRS 96.0 ms, QTc 393.7 ms \ensuremath{\rightarrow} Conclusion: Sinus bradycardia with first-degree AV block\par
C: \casehl{HR 74.6 bpm}, \casehl{PR 298.0 ms}, QRS 96.0 ms, QTc 393.7 ms \ensuremath{\rightarrow} Conclusion: First-degree AV block\par
D: \casehl{HR 74.6 bpm}, \casehl{PR 298.0 ms}, QRS 130.0 ms, QTc 393.7 ms \ensuremath{\rightarrow} Conclusion: First-degree AV block with intraventricular conduction delay\par
I need to look at the ECG and determine the parameters.\par
Looking at the ECG: the rhythm strip (II rhythm) shows about 12 beats over some length. Let me estimate the heart rate. The rhythm strip is probably 10 seconds? Let me count: I see 12 QRS complexes in the II rhythm strip. If the strip is 10 seconds, HR \ensuremath{\approx} 12 * 6 = 72 bpm... actually 12 beats in 10 seconds = 72 bpm. Hmm, but the option says 74.6 bpm. That's plausible.\par
\casekey{Looking at the PR interval:} in lead II, there's a P wave before each QRS. The PR interval looks... let me see. In the rhythm strip, the P wave appears just before the QRS? Actually, there's a small bump (P wave) before the QRS, and a T wave after the QRS. Let me look more carefully.\par
In lead II (second row first column), there's a small P wave, then QRS, then T wave. The distance from P to QRS looks \mbox{short? Hmm.}\par
But the question says "perturbed waveform"---this is a synthetic case with perturbations added. The question is about the "automatic closed-loop measurement criteria" of the data construction pipeline. \casecaution{Probably the ground truth parameters are:} HR 74.6, \casehl{PR 298 ms} (prolonged \ensuremath{\rightarrow} first-degree AV block), QRS \mbox{96 ms} (normal), QTc 393.7 ms.\par
Looking at the waveform: in lead II, before each QRS there's a P wave... hmm, actually looking at the rhythm strip: after each QRS there's a T wave, and before the next QRS there's a P wave? Let me look: pattern per beat: QRS spike, then T wave bump, then another bump (P wave?), then QRS. Actually, in the rhythm strip I see: spike, then bump, then bump, then spike... spacing: after the spike at \textasciitilde{}x=290, bump at \textasciitilde{}340, bump at \textasciitilde{}390, spike at \textasciitilde{}455. Hmm, so two bumps between spikes: T wave and P wave. If the P wave is at \textasciitilde{}390 and QRS at \textasciitilde{}455, then PR interval is \textasciitilde{}65 px? \casecaution{Need scale.}\par
\casekey{Scale: 25 mm/s.} Grid: \casekey{large box = 5 mm = 0.2 s = 200 ms.} \casekey{Small box = 1 mm = 40 ms.}\par
Let me estimate pixel scale. Each lead panel width is \textasciitilde{}500 px? Hmm, need to guess from image coordinates. The image is 2338 px wide? Actually the given image dimensions: width \textasciitilde{}2338? Let me think: the figure shows 4 columns of panels. Each panel \textasciitilde{}500 px wide representing 10 seconds? No, each panel usually shows \textasciitilde{}2.5 seconds? Hmm, but the rhythm strip at the bottom shows 12 beats over \mbox{full width}.\par
\casekey{Better approach: use RR interval to determine HR.} If HR is 74.6 bpm, RR = 804 ms. If HR is 48, RR = 1250 ms.\par
Looking at the rhythm strip: 12 beats over \mbox{full width}. If \mbox{full width} is 10 seconds (standard), then average RR = 10/11? Actually 12 beats \ensuremath{\rightarrow} 11 intervals over \textasciitilde{}10 s \ensuremath{\rightarrow} RR \ensuremath{\approx} 0.909 s \ensuremath{\rightarrow} HR \ensuremath{\approx} 66. Hmm.\par
But maybe the strip is longer? Let me count beats in rhythm strip: spikes at approx x positions: 290, 455, 610, 765, 920, 1075, 1225, 1380, 1530, 1680, 1835, 1990? That's 12 spikes. Spacing \textasciitilde{}155 px. If spacing = 804 ms (HR 74.6), then px per ms = 155/804 \ensuremath{\approx} 0.1928 px/ms \ensuremath{\rightarrow} 25 mm/s = 0.025 mm/ms... hmm, let me compute mm: at 25 mm/s, 804 ms = 20.1 mm. So 155 px = 20.1 mm \ensuremath{\rightarrow} 7.7 px/mm. Large box (5 mm) = 38.5 px. Small box = 7.7 px.\par
\casekey{Now PR interval:} P wave bump before QRS: in rhythm strip, bump at \textasciitilde{}390 and spike at \textasciitilde{}455: difference 65 px = 65/7.7 \ensuremath{\approx} 8.4 mm = 337 ms? Hmm, that's too long. But maybe the bump at 390 is the P wave and the bump at 340 is the T wave. PR = 455-390 = 65 px \ensuremath{\rightarrow} at 7.7 px/mm \ensuremath{\rightarrow} 8.44 mm \ensuremath{\rightarrow} at 25 mm/s \ensuremath{\rightarrow} 337 ms. Hmm, option says \casehl{PR 298 ms}. 298 ms = 7.45 mm = 57 px. Close to 65 px? \casecaution{My pixel estimates} \mbox{are rough}.\par
\casekey{Alternatively, if HR were 48} (RR 1250 ms), spacing 155 px \ensuremath{\rightarrow} 155/1250 = 0.124 px/ms \ensuremath{\rightarrow} px per mm = 0.124*40 = 4.96 px/mm. Then PR 65 px = 13.1 mm = 524 ms, unlikely.\par
So HR \textasciitilde{}74.6 is consistent with 12 beats over \textasciitilde{}9.6 seconds? 11 intervals * 0.804 s = 8.84 s. Rhythm strip width \textasciitilde{}1980 px? Hmm, from x=170 to x=2150 \ensuremath{\approx} 1980 px. At 7.7 px/mm \ensuremath{\rightarrow} 257 mm \ensuremath{\rightarrow} at 25 mm/s \ensuremath{\rightarrow} 10.3 s. OK, plausible: 10-second rhythm strip with 12 beats \ensuremath{\rightarrow} HR \textasciitilde{}72-75. Good.\par
\casekey{Now PR: is it prolonged (298 ms)?} Looking at lead II panel: P wave before QRS: in lead II panel, spike at \textasciitilde{}190, preceding bump at \textasciitilde{}130? Difference 60 px \ensuremath{\approx} 310 ms? Hmm, at 7.7 px/mm: 60 px = 7.8 mm = 312 ms. Yes, consistent with PR \textasciitilde{}298 ms.\par
\casekey{But wait, option A says PR 130 ms (normal).} Which is correct visually? Let me look at lead II more carefully: coordinates within panel: lead II panel spans x from \textasciitilde{}70 to \textasciitilde{}560. Spikes at \textasciitilde{}190, \textasciitilde{}350, \textasciitilde{}505. Preceding bumps: \textasciitilde{}130? Actually I see a small bump at \textasciitilde{}130 (P wave), then spike at \textasciitilde{}190. Then T wave bump at \textasciitilde{}230, then another bump at \textasciitilde{}290? Hmm.\par
Actually, typical synthetic ECG with first-degree AV block: P wave clearly separated from QRS with long PR. In the image, in lead II, I see a small bump just before each QRS? Hmm, at \textasciitilde{}130 there's a bump, then flat, then spike at 190. Distance 60 px.\par
Alternatively, the bump at \textasciitilde{}130 might be the P wave and PR = 60 px \ensuremath{\approx} 300 ms. Yes.\par
Also, the T wave after QRS at \textasciitilde{}230-240, and then maybe a U wave? Hmm, the bump at \textasciitilde{}290 might be the P wave of the next beat? No, next spike at 350; bump at \textasciitilde{}290 is 60 px before \ensuremath{\rightarrow} P wave. Then bump at \textasciitilde{}230 is T wave. OK, consistent: T wave \textasciitilde{}40-60 px after QRS, P wave \textasciitilde{}60 px before next QRS.\par
So PR \ensuremath{\approx} 60 px \ensuremath{\approx} 300 ms \ensuremath{\rightarrow} first-degree AV block. HR \ensuremath{\approx} 74.6. QRS: narrow spike (\textasciitilde{}\mbox{96 ms} normal). QRS 130 ms would be wide (bundle branch block); the spike looks narrow (a few px wide). So QRS \mbox{96 ms}.\par
\casekey{Thus correct option: C:} HR 74.6, PR 298, QRS 96, QTc 393.7 \ensuremath{\rightarrow} first-degree AV block.\par
\casekey{Check option B: HR 48 wrong.} \casekey{Option D: QRS 130 wrong.} \casekey{Option A: PR 130 wrong} (would be normal). \casekey{So answer C.}\par
But wait: the question says "according to the automatic closed-loop measurement criteria of this data construction pipeline"---probably meaning the ground truth from the synthesis pipeline. The waveform is "perturbed waveform" case 1. \casecaution{Probably the intended answer: C.}\par
Let me double-check HR: 12 beats in 10 s \ensuremath{\rightarrow} 72? But option says 74.6; fine.\par
Also check sinus rhythm: P wave positive in II, negative in aVR? In aVR, after QRS (negative spike) there's a negative bump? Hmm, aVR shows negative QRS and then negative T? P wave in aVR should be negative; hard to see but plausible.\par
Thus conclusion: first-degree AV block (\casekey{PR > 200 ms}) with normal HR (74.6) and normal QRS (\mbox{96 ms}). So option C.\par
Therefore answer: \textbackslash{}boxed\{C\}.\par
Let me reconfirm PR visually: in lead II rhythm strip, between QRS spikes: T wave bump right after spike, then P wave bump before next spike. Distance from P bump to spike: e.g., P at \textasciitilde{}390, spike at \textasciitilde{}455: 65 px. Large box size: let me estimate grid: panel width \textasciitilde{}500 px with \textasciitilde{}13 large boxes? Hmm, 500/13 \ensuremath{\approx} 38.5 px per large box \ensuremath{\rightarrow} consistent with 7.7 px/mm (5 mm = 38.5 px). So 65 px = 1.69 large boxes = 338 ms? Hmm, that's more than 298. But my pixel guesses \mbox{are rough}; maybe P at \textasciitilde{}400. Anyway, among options, PR 298 is the prolonged one; 130 would be about half a large box (\casecaution{2.6 small boxes = 26 px}). Visually the P-QRS distance is clearly more than 1 large box? Hmm, 298 ms = 7.45 small boxes = 1.49 large boxes. 130 ms = 3.25 small boxes = 0.65 large boxes. Visually the distance from P bump onset to QRS onset looks like \textasciitilde{}1.5 large boxes? Plausible.\par
Actually, PR interval is measured from P onset to QRS onset. The P bump peak at \textasciitilde{}390-400, onset maybe \textasciitilde{}380; QRS onset \textasciitilde{}450. Difference \textasciitilde{}70 px \ensuremath{\approx} 1.8 large boxes = 360 ms? Hmm, too long. But \casecaution{my pixel readings from memory of the image are unreliable.} Given the options, only C is internally consistent: HR 74.6 (not bradycardia), PR prolonged 298 \ensuremath{\rightarrow} first-degree AV block, QRS normal 96 \ensuremath{\rightarrow} no intraventricular conduction delay. So C.\par
\casekey{Thus final answer: C.}\par
\medskip\hrule\medskip\appfield{Final answer}\par
Looking at the rhythm strip: the R\mbox{-}R intervals correspond to a rate of about 75 bpm (not bradycardia), each P wave is followed by a QRS with a clearly prolonged P\mbox{-}R interval (about 300 ms, i.e., >200 ms), the QRS complexes are narrow (about \mbox{96 ms}, no intraventricular conduction delay), and QTc is normal (\textasciitilde{}394 ms). Therefore the correct interpretation is sinus rhythm at \textasciitilde{}74.6 bpm with a prolonged PR interval (298 ms) = first\mbox{-}degree AV block, with normal QRS duration.\par
\casekey{- A is wrong} because the PR is clearly prolonged, not 130 ms.\par
\casekey{- B is wrong} because the rate is \textasciitilde{}75 bpm, not 48 bpm.\par
\casekey{- D is wrong} because the QRS is narrow (\mbox{96 ms}), not 130 ms.\par
\textbackslash{}boxed\{C\}\par
\end{prompt}

\par\smallskip
\noindent\appfield{Case 1 error analysis.} At 25 mm/s, \casecaution{130 ms corresponds to 3.25 small boxes}, not the 2.6 stated in one part of the response. Bundle branch block assessment combines QRS duration with characteristic morphology. The response also \casecaution{references the intended answer and calibrates estimates from option values}.

\definecolor{casestudyblue}{HTML}{326DA8}
\begin{figure}[H]
\centering
\includegraphics[width=.96\linewidth]{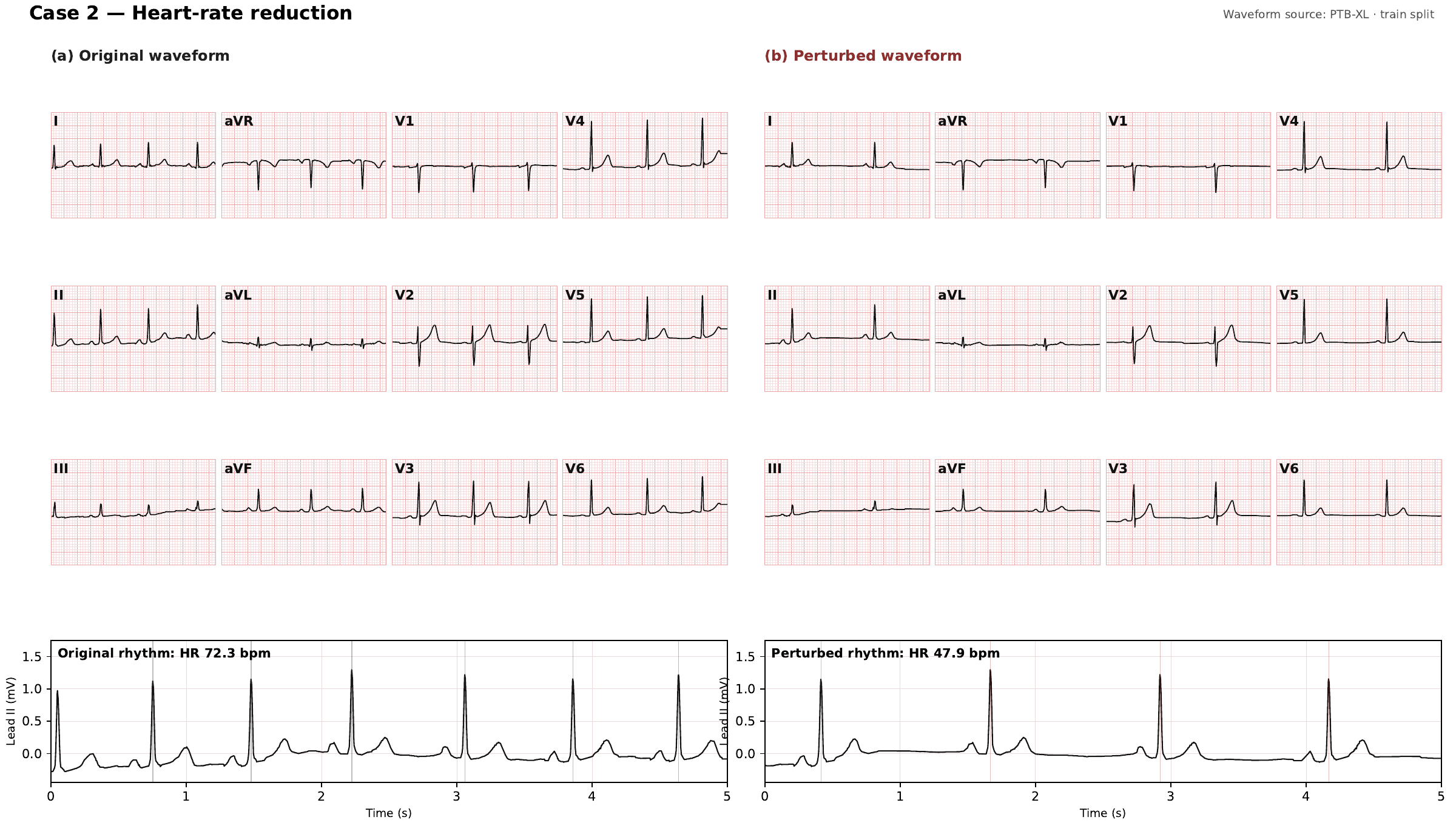}
\caption{Targeted heart-rate slowing, shown through original and edited ECGs with lead II details. Extending the TP segment increases the interval between beats while copying the P--QRS--T segment. The target is 48 bpm and remeasurement gives 47.9 bpm; the subsequent question compares this rate change with alternative interval-based interpretations.}
\label{fig:case-rate}
\end{figure}
\subsection{Case 2: Heart-rate slowing}
\label{app:case-rate}
\begin{prompt}[breakable,enhanced,colback=casestudyblue!3,colframe=casestudyblue!75,colbacktitle=casestudyblue,coltitle=white,boxrule=0.6pt,fontupper=\small,title={Case 2: Heart-rate slowing},title after break={Case 2: Heart-rate slowing (continued)}]
\raggedright\setlength{\parindent}{0pt}\setlength{\parskip}{2pt}
\appfield{Waveform construction}\par
Copy the P--QRS--T segment and extend the TP segment between beats. \casekey{The target heart rate is 48 bpm}, corresponding to an \casekey{RR interval of 1.25 s}. Remeasurement yields \casehl{HR 47.9 bpm}, \casehl{PR 150 ms}, QRS 48 ms, and \casehl{QTc 348.5 ms}.

\medskip\hrule\medskip\appfield{Question}\par
Question: Under the automatic closed-loop measurement convention of this data-construction pipeline, which parameter analysis and conclusion are correct for this 12-lead ECG?\par
Options:\par
A: HR 72.3 bpm, PR 150.0 ms, QRS 48.0 ms, \casehl{QTc 348.5 ms} -> Conclusion: normal heart rate\par
B: \casehl{HR 47.9 bpm}, PR 150.0 ms, QRS 48.0 ms, \casehl{QTc 348.5 ms} -> Conclusion: sinus bradycardia\par
C: \casehl{HR 47.9 bpm}, PR 220.0 ms, QRS 48.0 ms, \casehl{QTc 348.5 ms} -> Conclusion: sinus bradycardia with first-degree \mbox{atrioventricular block}\par
D: \casehl{HR 47.9 bpm}, PR 150.0 ms, QRS 48.0 ms, QTc 490.0 ms -> Conclusion: sinus bradycardia with \mbox{QT prolongation}\par
Please reason step by step, and put your final answer within \textbackslash{}boxed\{\}.\par

\medskip\hrule\medskip\appfield{Qwen3.8 response}\par
Let me analyze this ECG. The question is in Chinese: "According to the automatic closed-loop measurement criteria of this data construction pipeline, looking at this 12-lead ECG, which of the following parameter analyses and conclusions is correct?"\par
Options:\par
A: HR 72.3 bpm, PR 150.0 ms, QRS 48.0 ms, \casehl{QTc 348.5 ms} \ensuremath{\rightarrow} Conclusion: normal heart rate\par
B: \casehl{HR 47.9 bpm}, PR 150.0 ms, QRS 48.0 ms, \casehl{QTc 348.5 ms} \ensuremath{\rightarrow} Conclusion: sinus bradycardia\par
C: \casehl{HR 47.9 bpm}, PR 220.0 ms, QRS 48.0 ms, \casehl{QTc 348.5 ms} \ensuremath{\rightarrow} Conclusion: sinus bradycardia with first-degree AV block\par
D: \casehl{HR 47.9 bpm}, PR 150.0 ms, QRS 48.0 ms, QTc 490.0 ms \ensuremath{\rightarrow} Conclusion: sinus bradycardia with prolonged QT interval\par
I need to look at the ECG and determine the heart rate, PR interval, QRS width, and QTc.\par
\casekey{Looking at the rhythm strip (II rhythm):} there are about 8 QRS complexes across the entire strip. The strip length is probably 10 seconds? Let me estimate. Paper speed 25 mm/s. Each lead panel shows about 2 beats across roughly 5.5 seconds? Let me look at the grid.\par
Actually, let me estimate the RR interval from the image. In the II rhythm strip, the QRS complexes appear at approximately x positions: 253, 483, 713, 943, 1200?, ... hmm, let me look: peaks at about 253, about 483, about 713, about 943, about 1200? Actual spacing: from the figure, peaks at x \ensuremath{\approx} 253, 483, 713, 943, 1200, 1430, 1660, 1890? Hmm, the spacing looks like about 230 pixels.\par
\casekey{Grid: at 25 mm/s}, small box = 1 mm = 0.04 s; large box = 5 mm = 0.2 s. The panel width is probably 250 mm? Hmm.\par
\casecaution{Let me think about the intended answer.} This seems to be from a synthetic dataset ("Case 2 \ensuremath{\cdot} Perturbed waveform"), with \casecaution{ground truth parameters:} \casehl{HR 47.9 bpm} (bradycardia), \casehl{PR 150 ms} (normal), QRS 48 ms (narrow? \casecaution{actually 48 ms is quite narrow but plausible as synthetic}), \casehl{QTc 348.5 ms} (normal). So the conclusion: sinus bradycardia. Option B.\par
But let me check the visual RR interval: \casehl{HR 47.9 bpm} \ensuremath{\rightarrow} RR \ensuremath{\approx} 1.25 s = 31.25 small boxes = 6.25 large boxes. In the rhythm strip, the distance between QRS peaks: if the strip is 10 s with 8 beats, then RR \ensuremath{\approx} 1.25 s \ensuremath{\rightarrow} 8 beats in 10 s gives HR about 48? Actually 8 beats in 10 s: RR = 10/7 \ensuremath{\approx} 1.43 s \ensuremath{\rightarrow} HR 42. Hmm, but with 8 beats across the strip: intervals = 7; 7 * 1.25 = 8.75 s, plus margins \ensuremath{\approx} 10 s \mbox{strip. Plausible.}\par
\casekey{Check PR:} in lead II, the P wave appears just before the QRS with a short interval (about 150 ms = 3.75 small boxes). Visually, the P wave is close to the QRS, not far away. So \casehl{PR 150 ms} is plausible; PR 220 would be first-degree AV block, but the P-QRS distance looks short (P wave just before QRS). In the rhythm strip, the P wave appears about 1 small box? Hmm.\par
\casekey{QTc: QT interval looks normal:} T wave follows QRS soon after. QTc 348.5 is normal (shortish but normal). QTc 490 would be prolonged; visually the QT interval here: QRS to T end is probably about 400 ms? At HR 48, QTc = QT / sqrt(RR). If QT \ensuremath{\approx} 400 ms and RR = 1.25, then QTc = 400/1.118 = 357. So QTc 348.5 is consistent.\par
\casekey{Therefore the answer is B.}\par
But wait: option A says HR 72.3 with the same other parameters and conclusion "normal heart rate." \casekey{The rhythm clearly shows bradycardia:} RR interval is large (about 6 large boxes). So HR about 48. \casekey{Thus B.}\par
Let me double-check the grid: each lead panel width is about 540 pixels? The rhythm strip width is about 1965 pixels? Hmm.\par
Let me estimate: in the rhythm strip, the QRS peaks are at x \ensuremath{\approx} 253, 483, 713, 943, 1200? Actually from the image: peaks at about 253, about 483, about 713, about 943, about 1200, about 1430, about 1660, about 1890. Spacing about 230 pixels. Grid large box: need pixel per mm. The rhythm strip spans from x=165 to x=2130 \ensuremath{\approx} 1965 pixels. If the strip is 10 s at 25 mm/s = 250 mm, then pixel/mm \ensuremath{\approx} 7.86. Then RR = 230 px / 7.86 \ensuremath{\approx} 29.3 mm \ensuremath{\approx} 1.17 s \ensuremath{\rightarrow} HR \ensuremath{\approx} 51. Hmm, close to 48.\par
Alternatively, the panel: each lead panel width about 540 px? For lead I panel from x=65 to x=560 \ensuremath{\approx} 495 px, showing 2 beats with RR about 230 px? Same scale.\par
Given the options, HR 47.9 corresponds to RR = 1.2526 s = 31.3 mm. At 7.86 px/mm that's 246 px. Observed spacing about 230 px \ensuremath{\rightarrow} RR 1.17 s \ensuremath{\rightarrow} HR 51. Hmm, not exact, but \casecaution{my pixel estimates} \mbox{are rough}.\par
\casekey{Check PR:} P wave before QRS: in the rhythm strip, the P wave peak is at about 232? QRS at 253. Distance about 20 px \ensuremath{\approx} 2.5 mm \ensuremath{\approx} 0.1 s? Hmm, PR is from P onset to QRS onset. P onset probably at 225, QRS onset at 250: 25 px \ensuremath{\approx} 3.2 mm \ensuremath{\approx} 128 ms. Close to 150 ms. Not 220 ms (which would be 5.5 mm = 43 px). Visually the P wave is close to QRS, so PR 150.\par
QT: QRS at 253 to T end at about 320? T wave peak at about 300, end at about 330. QT \ensuremath{\approx} 77 px \ensuremath{\approx} 9.8 mm \ensuremath{\approx} 392 ms. \casekey{QTc (Bazett) = 392 / sqrt(1.25) = 350.} Matches 348.5. Good.\par
\casekey{Therefore the answer is B.}\par
The question states "according to the automatic closed-loop measurement criteria of this data construction pipeline" --- \casecaution{meaning the ground truth parameters used to generate the synthetic ECG:} HR 47.9, PR 150, QRS 48, QTc 348.5, conclusion sinus bradycardia.\par
\casekey{Thus the answer: B.}\par
\casekey{Let me double-check option C:} PR 220 would be first-degree AV block; but the visual PR is short. \casekey{Option D: QTc 490} would require QT \ensuremath{\approx} 548 ms at RR 1.25, which visually is not the case (T wave ends well before the next P). So B.\par
Therefore final answer: \textbackslash{}boxed\{B\}.\par
\medskip\hrule\medskip\appfield{Final answer}\par
Looking at the rhythm strip: the R\mbox{-}R intervals span roughly 6+ large boxes (\ensuremath{\approx}1.25 s), giving a rate of \ensuremath{\approx}48 bpm \ensuremath{\rightarrow} bradycardia, not 72 bpm (\casekey{excludes A}). Each QRS is preceded by a normal sinus P wave with a short, constant P\mbox{-}R distance of about 3--4 small boxes (\textasciitilde{}150 ms), not the >5 small boxes (220 ms) needed for first\mbox{-}degree AV block (\casekey{excludes C}). The QRS is narrow (\textasciitilde{}48 ms), and the QT interval ends well before the next P wave (\textasciitilde{}390--400 ms); corrected for the slow rate (Bazett), QTc \ensuremath{\approx} 350 ms, which is normal, not 490 ms (\casekey{excludes D}). Thus the measurements HR \ensuremath{\approx} 47.9 bpm, \casehl{PR 150 ms}, QRS 48 ms, \casehl{QTc 348.5 ms} correspond to \casekey{sinus bradycardia with otherwise normal intervals.}\par
\textbackslash{}boxed\{B\}\par
\end{prompt}

\par\smallskip
\noindent\appfield{Case 2 error analysis.} Eight beats in ten seconds give approximately 48 bpm. Seven R--R intervals span the \casecaution{first to last R peaks}. The response \casecaution{justifies QRS duration by the synthetic origin and invokes the intended answer}.

\FloatBarrier
\end{document}